\documentclass[a4paper]{report}
\usepackage[utf8]{inputenc}
\usepackage{amsmath,amssymb,amsfonts}
\usepackage{graphicx,caption,subcaption}
\usepackage{pdflscape}
\usepackage{algorithm,algpseudocode}
\usepackage{mathrsfs}
\usepackage{setspace}
\usepackage{titlepic}
\usepackage{booktabs, xurl, color, epstopdf, xcolor}
\usepackage[bottom]{footmisc}

\usepackage[margin=1.5in]{geometry}
\usepackage{pdfpages}
\usepackage{textcomp}
\usepackage[acronym,automake,symbols,nogroupskip]{glossaries-extra}
\usepackage[sort&compress]{natbib}
\usepackage{afterpage}
\usepackage{multirow}
\usepackage{tikz}
\usetikzlibrary{shadows,positioning}
\usepackage{mathtools}

\newcommand{\tab}{\hspace*{2em}}
\newcommand\blankpage{%
\null
\thispagestyle{empty}%
\addtocounter{page}{-1}%
\newpage}

\graphicspath{{images/}}

\makeglossaries
\newacronym{lsh}{LSH}{Locality Sensitive Hashing}
\newacronym{fire}{FiRE}{Finder of Rare Identities}
\newacronym{knn}{$k$NN}{$k^\text{th}$ Nearest Neighbor}
\newacronym{knnw}{$k$NNW}{$k$NN-weight}
\newacronym{lic}{LIC}{Local Isolation Coefficient}
\newacronym{lof}{LOF}{Local Outlier Factor}
\newacronym{ldof}{LDOF}{Local Distance-based Outlier Factor}
\newacronym{inflo}{INFLO}{INFLuenced Outlierness}
\newacronym{ldf}{LDF}{Local Density Factor}
\newacronym{lde}{LDE}{Local Density Estimate}
\newacronym{kdeos}{KDEOS}{Kernel Density Estimation Outlier Scores}
\newacronym{loop}{LoOP}{Local Outlier Probabilities}
\newacronym{cof}{COF}{Connectivity-based Outlier Factor}
\newacronym{odin}{ODIN}{Outlier detection using Indegree Number}
\newacronym{fastabod}{FastABOD}{Fast Angle-Based Outlier Detection}
\newacronym{abof}{ABOF}{Angle-based outlier factor}
\newacronym{sod}{SOD}{Subspace Outlier Degree}
\newacronym{hbos}{HBOS}{Histogram-based Outlier Score}
\newacronym{adwin}{ADWIN}{ADaptive WINdowing}
\newacronym{arf}{ARF}{Adaptive Random Forest}
\newacronym{scRNA-seq}{scRNA-seq}{single-cell RNA sequencing}
\newacronym{EPCs}{EPCs}{Endothelial Progenitor Cells}
\newacronym{CTCs}{CTCs}{Circulating Tumor Cells}
\newacronym{PBMCs}{PBMCs}{peripheral blood mononuclear cells}
\newacronym{PBMC}{PBMC}{peripheral blood mononuclear cell}
\newacronym{ESCs}{ESCs}{embryonic stem cells}
\newacronym{ESC}{ESC}{embryonic stem cell}
\newacronym{RMS}{RMS}{Root Mean Square}
\newacronym{DE}{DE}{differentially expressed}
\newacronym{ISH}{ISH}{\textit{In Situ} Hybridization}
\newacronym{SNV}{SNV}{Single Nucleotide Variant}
\newacronym{tsne}{t-SNE}{t-Distributed Stochastic Neighbor Embedding}
\newacronym{DCs}{DCs}{Dendritic Cells}
\newacronym{DC}{DC}{Dendritic Cell}
\newacronym{FACS}{FACS}{Fluorescence-activated cell sorting}
\newacronym{dbos}{DB-outlier score}{Distance-based outlier score}
\newacronym{pan}{$\textit{P@}n$}{\textit{Precision at} $n$}
\newacronym{AP}{\textit{AP}}{\textit{Average Precision}}
\newacronym{rocauc}{\textit{ROC AUC}}{\textit{Receiver Operator Characteristics Area Under the Curve}}
\newacronym{auc}{\textit{AUC}}{\textit{Area Under the Curve}}
\newacronym{roc}{\textit{ROC}}{\textit{Receiver Operator Characteristics}}
\newacronym{tpr}{\textit{TPR}}{\textit{True Positive Rate}}
\newacronym{fpr}{\textit{FPR}}{\textit{False Positive Rate}}
\newacronym{fp}{FP}{False Positives}
\newacronym{tn}{TN}{True Negatives}
\newacronym{awe}{AWE}{Accuracy-Weighted Ensemble}
\newacronym{aee}{AEE}{Additive Expert Ensemble}
\newacronym{ob}{OB}{Online Bagging-ADWIN}
\newacronym{osmoteb}{OSMOTEB}{Online SMOTE Bagging}
\newacronym{lb}{LB}{Leveraging Bagging}

\begin{document}

    \begin{titlepage}
    \begin{center}
        \vspace{3cm}
        \textbf{NEIGHBORHOOD DENSITY ESTIMATION USING SPACE-PARTITIONING BASED HASHING SCHEMES}

        \vspace{3cm}

        \textbf{AASHI JINDAL}\\
        \vspace{3cm}

        \includegraphics[width=4cm, height=4cm]{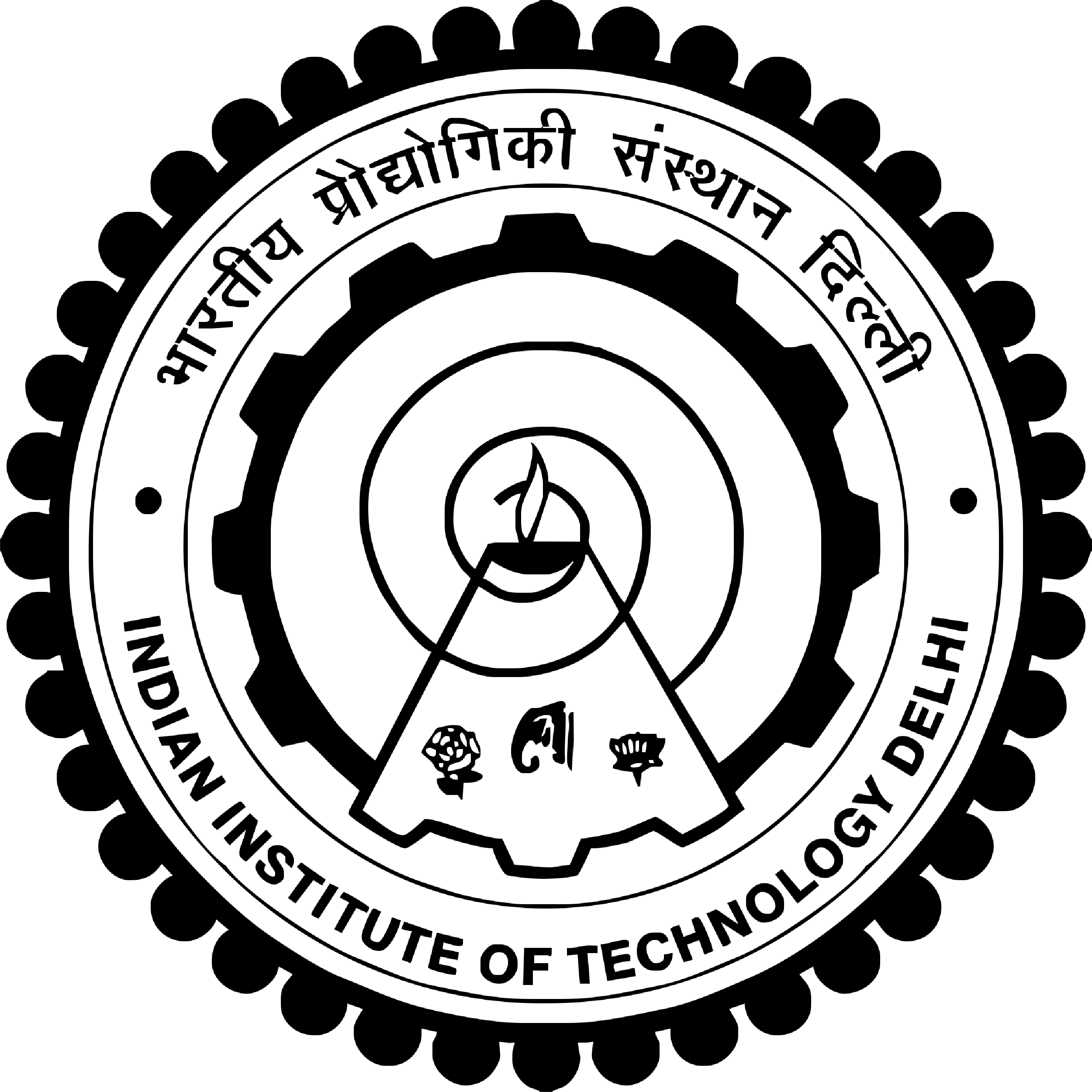}
        \\
        \vspace{5cm}
        \textbf{DEPARTMENT OF ELECTRICAL ENGINEERING}\\
        \textbf{INDIAN INSTITUTE OF TECHNOLOGY DELHI}\\
        \textbf{June 2023}\\
    \end{center}
    \afterpage{\blankpage}
\end{titlepage}
\afterpage{\blankpage}
\begin{titlepage}
    \topskip0pt
    \vspace*{\fill}
    \textbf{\textcopyright Indian Institute of Technology Delhi (IITD), New Delhi, 2023}
    \vspace*{\fill}
\end{titlepage}
\afterpage{\blankpage}
\begin{titlepage}
    \begin{center}
        \vspace{2cm}
        \textbf{NEIGHBORHOOD DENSITY ESTIMATION USING SPACE-PARTITIONING BASED HASHING SCHEMES}

        \vspace{2cm}
        by\\

        \vspace{1cm}
        \textbf{AASHI JINDAL}\\

        \vspace{1cm}
        DEPARTMENT OF ELECTRICAL ENGINEERING\\
        \vspace{0.5cm}
        Submitted\\
        \vspace{0.25cm}
        \textit{in fulfillment of the requirements of the degree of Doctor of Philosophy\\
            to the}\\

        \vspace{2cm}

        \includegraphics[width=4cm, height=4cm]{logo}
        \\
        \vspace{2cm}

        \textbf{INDIAN INSTITUTE OF TECHNOLOGY DELHI}\\
        \textbf{June 2023}\\
    \end{center}
\end{titlepage}


\setcounter{page}{1}


\clearpage
\renewcommand{\thepage}{\roman{page}}
\begin{center}
\LARGE{\textbf{Certificate}}\\
\end{center}
\vspace{0.3in}

\doublespacing{
\tab This is to certify that the thesis entitled \textbf{``Neighborhood Density Estimation Using Space-Partitioning Based Hashing Schemes''}, being submitted by \textbf{Aashi Jindal} for the award of the degree of \textbf{Doctor of Philosophy} to the Department of Electrical Engineering, Indian Institute of Technology Delhi, is a record of bonafide work done by her under my supervision and guidance. The matter embodied in this thesis has not been submitted to any other University or Institute for the award of any other degree or diploma.}\\

\vspace{0.6in}

\begin{tabular}{ll}
    \begin{minipage}[t]{.5\textwidth}
        \begin{flushleft}
            \large{\textbf{Dr. Jayadeva}}           \\
            \vspace{1mm}
            \normalsize{\textit{Professor}} \\
            \vspace{1mm}
            \text{Department of Electrical Engineering,} \\
            \vspace{1mm}
            \text{Indian Institute of Technology Delhi,} \\
            \vspace{1mm}
            \text{Hauz Khas, New Delhi - 110016,} \ \\
            \vspace{1mm}
            \text{INDIA.}
        \end{flushleft}
    \end{minipage} &
    
    \begin{minipage}[t]{.55\textwidth}
        \begin{flushleft}
            \large{\textbf{Dr. Shyam Prabhakar}} \\
            \vspace{1mm}
            \normalsize{\textit{Associate Director}}\\
            \vspace{1mm}
            \text{Laboratory of Systems Biology and Data Analytics,} \\
            \vspace{1mm}
            \text{Genome Institute of Singapore,}\\
            \vspace{1mm}
            \text{SINGAPORE}
        \end{flushleft}
    \end{minipage} \\ \\ \\
    
    \begin{minipage}[t]{.55\textwidth}
        \begin{flushleft}
            \large{\textbf{Dr. Debarka Sengupta}}           \\
            \vspace{1mm}
            \normalsize{\textit{Associate Professor}} \\
            \vspace{1mm}
            \text{Department of Computer Science \& Engineering,} \\
            \vspace{1mm}
            \text{Department of Computational Biology,}\\
            \vspace{1mm}
            \text{Centre for Artificial Intelligence,} \\
            \vspace{1mm}
            \text{Indraprashta Institute of Information Technology,} \\
            \vspace{1mm}
            \text{Okhla Phase III, Delhi - 110020, India} \\
            \vspace{1mm}
            \normalsize{\textit{(Adj.) Associate Professor}} \\        
            \vspace{1mm}    
            \text{Institute of Health \& Biomedical Innovation,}\\
            \vspace{1mm}
            \text{QUT, AUSTRALIA}\\
        \end{flushleft}
    \end{minipage} &\\
        
\end{tabular}




\setcounter{page}{1}
\addcontentsline{toc}{chapter}{Certificate}

\includepdf{blank.pdf}

\renewcommand{\thepage}{\roman{page}}
\begin{center}
\LARGE{\textbf{Acknowledgements}}\\
\end{center}

\vspace{0.3in}

I sincerely thank my supervisors Dr. Jayadeva, Dr. Debarka Sengupta and Dr. Shyam Prabhakar for their valuable feedback. I also thank other SRC members Dr. I.N. Kar, Dr. Shouri Chatterjee, and Dr. Vivekanandan Perumal. I thank Prof. Suresh Chandra for his encouragement. I thank my colleagues Dr. Sumit Soman and Dr. Udit Kumar for their moral support. I thank department staff members Rakesh, Yatindra, Satish, and Mukesh for always helping with the department work.

I thank my parents, Rakesh Jindal and Meena Jindal, for always believing in me and supporting me in my decisions. I thank my parents-in-law, Sanjiv Kumar Gupta and Snehlata Gupta for being always understanding. I thank my husband Prashant Gupta for always being there whenever I needed him. He supported me both professionally and personally. My brother, Sahil Jindal always holds a special place in my life. My core family members, Shruti Singla, Praveen Singla, Shaffi Bindal, Mohit Bindal, Khushboo Mehrotra, Yuvan Singla, Tejas Bindal, and Vanya Bindal are an important part of my life.

I also thank my current employers, Ashok Juneja and Shweta Singla for giving me the time to complete my pending research work when needed. I am thankful to my company, Applied Solar Technologies India Pvt. Ltd. and my colleagues, Subhdip Rakshit, Amod Kumar, Shubham Sinha, Pulkit Tyagi, Bikas Gupta, Sukrit Singh Negi, Abhishek Chaudhary and Anil Hansda.

The following have been the three main philosophies of my life, "There is an almighty somewhere looking after us and guiding us to the correct path." "Behind every successful woman is her parents, husband, and family that supported her in her unconventional choices." Last, but not the least, "...all's well, that ends well."

\vspace{1in}

\begin{flushright}
\textit{(Aashi Jindal)}
\end{flushright}

\addcontentsline{toc}{chapter}{Acknowledgements}%

\includepdf{blank.pdf}

\begin{abstract}
    \thispagestyle{plain}
    \renewcommand{\thepage}{\roman{page}}
    \addcontentsline{toc}{chapter}{Abstract}
    \setcounter{page}{5}
Single cell messenger RNA sequencing (scRNA-seq) offers a view into transcriptional landscapes in complex tissues. Recent developments in droplet based transcriptomics platforms have made it possible to simultaneously screen hundreds of thousands of cells. It is advantageous to use large-scale single cell transcriptomics since it could lead to the discovery of a number of rare cell sub-populations. When the sample size reaches the order of hundreds of thousands, existing techniques to discover rare cells either scale unbearably slow or terminate altogether. We suggest the Finder of Rare Entities (FiRE), an algorithm that quickly assigns a rareness score to every individual expression profile under consideration. We show how FiRE scores can assist bioinformaticians in limiting the downstream analyses to only on a subset of expression profiles within ultra-large scRNA-seq data. 

Anomaly detection methods differ in their time complexity, sensitivity to data dimensions, and their ability to detect local/global outliers. The proposed algorithm FiRE is a 'sketching' based linear-time algorithm for identifying global outliers. FiRE.1, an extended implementation of FiRE fares well on local outliers as well. We provide an extensive comparison with 18 state-of-the-art anomaly detection algorithms on a diverse collection of 1000 annotated datasets. Five different evaluation metrics have been employed. FiRE.1's performance was particularly remarkable on datasets featuring a large number of local outliers. In the sequel, we propose a new "outlierness" criterion to infer the local or global identity of outliers.

We propose Enhash, a fast ensemble learner that detects \textit{concept drift} in a data stream. A stream may consist of abrupt, gradual, virtual, or recurring events, or a mixture of various types of drift. Enhash employs projection hash to insert an incoming sample. We show empirically that the proposed method has competitive performance to existing ensemble learners in much lesser time. Also, Enhash has moderate resource requirements. Experiments relevant to performance comparison were performed on 6 artificial and 4 real datasets consisting of various types of drifts.
\end{abstract}

\includepdf[pages={1}]{blank.pdf}

\renewcommand{\thepage}{\roman{page}}
\setcounter{page}{7}
\addcontentsline{toc}{chapter}{Hindi Abstract}%
\includepdf[pagecommand={\thispagestyle{plain}}, pages={1}]{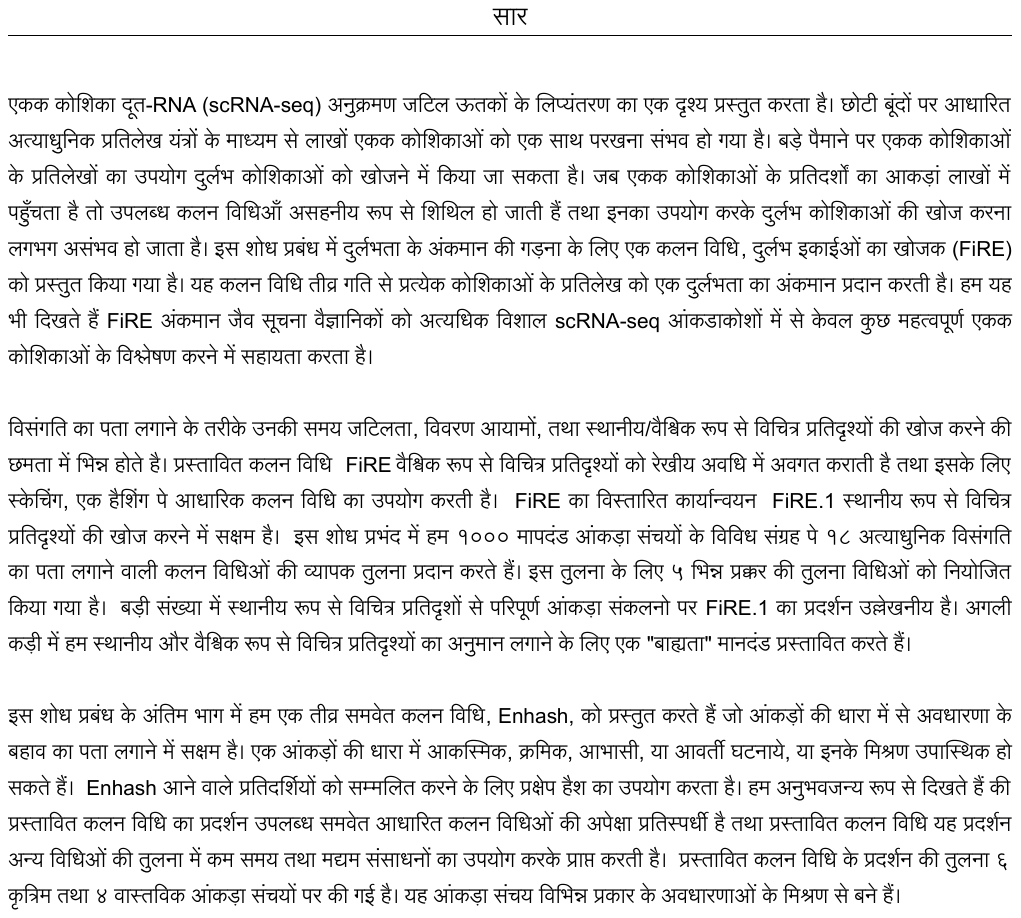}

\includepdf[pages={1}]{blank.pdf}


\setcounter{page}{9}

\tableofcontents
\includepdf[pages={1}]{blank.pdf}

\renewcommand{\thepage}{\roman{page}}
\setcounter{page}{13}
\listoffigures
\addtocontents{lof}{\protect\addcontentsline{toc}{chapter}{List of Figures}}

\includepdf[pages={1}]{blank.pdf}


\listoftables
\addtocontents{lot}{\protect\addcontentsline{toc}{chapter}{List of Tables}}

\printglossary[type=\acronymtype,nonumberlist,title=List of Abbreviations]
\afterpage{\blankpage}


\newpage
\renewcommand{\thepage}{\arabic{page}}
\setcounter{page}{1}

\chapter{Introduction}\label{chapter::1}

The process of identifying, and where appropriate, removing anomalous observations from data, is referred to as \textit{outlier detection}~\cite{hodge2004survey}. Most commonly, \textit{anomaly detection} is used interchangeably with outlier detection. The concept of outlier detection is widely used to identify faulty engines to avoid undesirable consequences in the future, outlying observations in patients' medical records to detect the possibility of a disease, anomalous trends in prices of a stock to book profits or avoid losses, unusual patterns in credit card transactions to identify possible theft, etc. In effect, outlier detection plays an important role in several analytical tasks. Outlier detection techniques may be supervised or unsupervised. We focus on unsupervised anomaly detection techniques. In scenarios where the complete dataset may be loaded into RAM for mining, \textit{offline algorithms} are sufficient. There may be scenarios where RAM is limited, e.g. IoT devices, or where the distribution of a dataset may evolve or change with time; in such cases, online algorithms are necessary. In non-stationary environments, algorithms must adapt to new distributions as well. The process of learning about changes in the data distribution in streaming environments is referred to as \textit{concept drift detection}~\cite{de2019overview}.

\section{Scope and objectives}
For both streaming and non-streaming environments, we identify rarity/outlierness as the flip side of density around the samples. Samples or points with sparser neighborhoods are more likely to be rare, as against those with dense neighborhoods. With increasing dimension, finding absolute nearest neighbors becomes prohibitive. Hence, to avoid the \textit{curse of dimensionality}~\cite{marimont1979nearest}, approximate nearest neighbors are computed. There have been some attempts in the past to use the local density around the sample as the measure of its rarity~\cite{LOF}. We expedite the process by finding the neighbors in constant time using the concept of hashing. Hashing methods are widely used for several practical applications such as data mining~\cite{srivastava2018cellatlassearch, Github:annoy}, audio-video fingerprinting~\cite{wang2003industrial, mao2011robust}, identification of copy videos~\cite{esmaeili2010robust}, clustering of large single-cell data~\cite{Sinha170308, sinha2020improved}. The primary reason being that hashing techniques identify the approximate nearest neighbor(s) of a query in $O(1)$ time. We exploit hashing to identify outliers quickly and efficiently amongst several thousands of samples in high-dimensional spaces. We briefly discuss about \textit{\glsxtrfull{lsh}} and its variants in the subsequent section. This is followed with a brief background and the importance of the problem (\textit{outlier detection} and \textit{concept drift detection}); the motivation to propose a new technique for the same.

\section{Locality Sensitive Hashing}
Indyk and Motwani~\cite{indyk1998approximate, gionis1999similarity} introduced \textit{\glsxtrfull{lsh}} to identify approximate nearest neighbors. The key idea behind the concept is to use hash functions such that similar objects/samples have a higher probability of collision than those samples which are far apart. Then, the points in the same bucket represent each other's neighborhood. The hash functions are chosen so that they reduce the complexity of objects owing to their dimensionality while preserving the inter-point distances within a relative error of $\epsilon$ (Johson-Lindenstrauss lemma)~\cite{lindenstrauss1984extensions}.
Formally, \glsxtrshort{lsh} is defined as follows:

A family $\mathcal{H}$ is defined for a metric space $\mathcal{M} = (M,d)$, a threshold $R > 0$ and an approximation factor $c > 1$. It is a family of functions $h:M\rightarrow S$ which maps elements from the metric space to a bucket $s\in S$. The family is called ($R$, $cR$, $P_1$, $P_2$)-sensitive if for any two points $p$, $q \in \mathbb{R}^d$
    \begin{itemize}
        \item if $\lVert p - q \rVert \leq R$ then Pr$_{\mathcal{H}}[h(p) = h(q)] \geq P_1$
        \item if $\lVert p - q \rVert \geq cR$ then Pr$_{\mathcal{H}}[h(p) = h(q)] \leq P_2$
    \end{itemize}

For \glsxtrshort{lsh} family $\mathcal{H}$ to be useful, it should satisfy $P_1 > P_2$. In effect, an \glsxtrshort{lsh} technique is a distribution on a family $\mathcal{H}$ of hash functions, such that for any pair of objects $p$, and $q$
\[Pr_{h\in \mathcal{H}}[h(p) = h(q)] = \textit{sim}(p,q),\]

where $\textit{sim}(p,q) \in [0,1]$ is some similarity function defined on the collection of objects~\cite{charikar2002similarity}. We briefly discuss some of the LSH-based techniques that find approximate nearest neighbors in high-dimensional spaces~\cite{datar2004locality, optimallsh}, even when the intrinsic dimensionality is high. These methods represent samples via compact representation such that inter-point distances in the original feature space are well-approximated.

\subsection{Sketching}
The sketch construction algorithm~\cite{wang2007sizing} involves randomly picking a threshold for a given feature and assigning it a bit. If the value of a feature is greater or smaller than a given threshold, then bit 1 or bit 0 is assigned, respectively. The sketches constructed using this technique result in bit vectors. Let $B$ be the sketch size in bits. The similarity between objects \textit{sim}(p,q) is then computed as the measure of similarity between their corresponding sketches. There exist several evidences that show that sketches constructed using random projections can be used to approximate $l_1$ distance on the feature vectors with the Hamming distance on their sketches~\cite{lv2006ferret}.

\subsection{Binary hash}
Binary hash~\cite{charikar2002similarity, wang2014hashing} involves random projection of a sample $x$ by picking $B$ random vectors $w_i,\text{ where }i \in \{1,...,B\}$ (from the Gaussian distribution). The bit vector is then assigned to $x$ on the basis of $sign(w_{i}^{T}x),\text{ where }i \in \{1,...,B\}$. The probability of two samples hashing into the same bucket is determined by the cosine similarity between them in the original feature space.

\subsection{Projection hash}
Projection hash~\cite{optimallsh} involves linear projection of a sample, followed by assignment to the bucket via quantization. For linear projection of a sample $x$, $B$ random vectors $w_i, \text{ where }i \in \{1,...,B\}$ are chosen from the Gaussian distribution and a bias $\textit{bias}$ from the uniform distribution over [-\textit{bin-width}, \textit{bin-width}], where \textit{bin-width} is the quantization width. The hash code is computed as $\lfloor (w_{i}^{T}x + \textit{bias})/\textit{bin-width} \rfloor$, where $i \in \{1,...,B\}$.

\section{Outlier detection}
\subsection{Outlier detection: Definition}
The commonly used notions of outliers are as follows:
\begin{itemize}
\item ``An outlying observation, or outlier, is one that appears to deviate markedly from other members of the sample in which it occurs~\cite{grubbs1969procedures}.''

\item According to Hawkins~\cite{hawkins1980identification}, an outlier is ``an observation which deviates so much from other observations as to arouse suspicions that it was generated by a different mechanism.''

\item ``An object $O$ in a dataset $T$ is a $DB(p, D)$-outlier if at least fraction $p$ of the objects in $T$ lies greater than distance $D$ from $O$~\cite{dboutlierscore}.''
\end{itemize}

\subsection{Outlier detection: Use cases}
Outlier detection is widely used across several applications to extract samples that distinct from the majority population. In specific scenarios, these samples act as noise, while in other cases, they may be used to extract some useful hidden phenomena. The most critical application of outlier detection is identifying anomalous patterns in medical diagnoses to identify brain tumor~\cite{prastawa2004brain}, cancerous masses in mammograms~\cite{tarassenko1995novelty}, etc. It can also be used to monitor industrial production and point out defective manufacturing products~\cite{unsupervised}.

\subsection{Outlier detection: Existing techniques}
Several algorithms have been proposed in the recent past to identify outliers. They may be broadly categorized as supervised and unsupervised algorithms. However, supervised algorithms are more expensive than unsupervised algorithms. Also, supervised algorithms are particularly restrictive in class-imbalance problems~\cite{japkowicz2002class}.

We discuss in detail some of the widely used unsupervised methods that provide outlier scores. A score enables a qualitative comparison of the concerned methods. The Distance-based outlier score (DB-outlier score)~\cite{dboutlierscore} modifies the conventional definition of distance-based outlier detection, to rank points based on their outlierness values. The number of points present within a given \textit{distance} from a given point determines its rank. A higher rank indicates a higher degree of sparsity around a data point. However, for high dimensional points, it is very difficult to choose an appropriate value of \textit{distance}.

\glsxtrfull{knn}~\cite{knn} identifies outliers using the distance of a point to its $k^\text{th}$ nearest neighbor. The longer the distance, the higher is the rank. \glsxtrfull{knnw}~\cite{knnw} is an extension of \glsxtrshort{knn} that is less sensitive to the choice of $k$. It also scales linearly for points in high dimensional spaces by using the Hilbert space-filling curve. \glsxtrshort{knnw} maps points to a $d$-dimensional hypercube $D = [0,1]^d$. A Hilbert space-filling curve maps this hypercube to the interval $I = [0,1]$. This facilitates a quick search of $k$ nearest neighbors of a point, by identifying its predecessors and successors on $I$. It considers the sum of distances to $k$ nearest neighbors as the weight of a point $p$. The points with high values of weights are the potential outliers.

\glsxtrfull{lic}~\cite{lic} is defined as the sum of the distance of the farthest point in the $k$-neighborhood of $p$ ($k$-\textit{dist}$(p)$), and the average distance of points in the $k$-neighborhood from $p$ ($\textit{LDS}_k(p)$). For points in a dense cluster, both $k-\textit{dist}(p)$ and $\textit{LDS}_k(p)$ are small and hence, \textit{\glsxtrshort{lic}} is also small.

\glsxtrfull{lof}~\cite{LOF} considers only a restricted neighborhood to identify outliers, and is, therefore, more suitable for local outlier detection. It assigns a degree of outlierness to every point, and hence, called as local outlier factor. For a given point $p$, it defines $\textit{dist}_k(p)$ as the distance between $p$ and its $k^\text{th}$ NN. The local neighborhood of $p$, $\textit{N}_k(p)$, consists of all points with distance less than or equal to $\textit{dist}_k(p)$. It calculates the \textit{reachability} to every point in its $k$-neighborhood $N_k(p)$, $p^{'}\in N_k(p)$, as follows.
\begin{align}\label{reachability}
\textit{reachability}_k(p \leftarrow p') = \max\{\textit{dist}_k(p), \textit{dist}(p,p')\}
\end{align}
In the above \eqref{reachability}, $\textit{dist}(p,p')$ is an $L_P$ distance. Local reachability density of point $p$, $\textit{lrd}(p)$, is defined as the ratio of cardinality of $N_k(p)$ and the sum of \textit{reachability} to points in $\textit{N}_k(p)$. For a given point $p$, $\textit{\glsxtrshort{lof}}(p)$ is calculated as follows.
\begin{align}
\textit{LOF}_k(p) = \frac{\sum_{o \in \textit{N}_k(p)}\frac{\textit{lrd}(o)}{\textit{lrd}(p)} }{|\textit{N}_k(p)|}
\end{align}
Intuitively, a point $p$ with lower value of $\textit{lrd}(p)$, and higher values of \textit{lrd} of points in $\textit{N}_k(p)$ is assigned higher values of \textit{\glsxtrshort{lof}}.
Simplified-LOF~\cite{simplifiedlof} simplifies the definition of local reachability density in \glsxtrshort{lof}. It redefines it as the inverse of $\textit{dist}_k(p)$ for a given point $p$.

\glsxtrfull{ldof}~\cite{ldof} is an extension of \glsxtrshort{lof} for identification of outliers in scattered datasets. \textit{\glsxtrshort{ldof}} is the ratio between $\textit{LDS}_k(p)$ and the average of pairwise distances between points in $k$-neighborhood of $p$. Points with $\textit{\glsxtrshort{ldof}} > 1$ are more likely to be outliers. \glsxtrfull{inflo}~\cite{inflo} considers both the $k$-neighborhood of $p$ ($\textit{N}_k(p)$), and the reverse neighborhood of $p$ ($\textit{RNN}_k(p)$) to estimate $\textit{lrd}(p)$. $\textit{RNN}_k(p)$ refers to the collection of points for which $p$ exists in their $k$-neighborhood. The combined space of $\textit{N}_k(p)$ and $\textit{RNN}_k(p)$ is collectively called as $k$-influenced space for $p$ ($\textit{IS}_k(p)$). For a given point $p$, $\textit{\glsxtrshort{inflo}}_k(p)$ is calculated as follows.
\begin{align}
\textit{INFLO}_k(p) = \frac{\textit{lrd}_{avg}(\textit{IS}_k(p))}{\textit{lrd}(p)}\text{, where }
\textit{lrd}_{avg}(\textit{IS}_k(p)) = \frac{\sum_{o \in \textit{IS}_k(p)}\textit{lrd}(o)}{|\textit{IS}_k(p)|}
\end{align}
\glsxtrshort{inflo} uses the same definition of $\textit{lrd}(p)$ as in Simplified-LOF.

\glsxtrfull{ldf}~\cite{ldf} incorporates \glsxtrfull{lde}, a kernel density estimate (KDE), to identify outliers. $\textit{LDE}(p)$ is computed as follows.
\begin{align} \label{lde}
\textit{LDE}(p) = \frac{\sum_{o \in \textit{N}_k(p)}\frac{1}{(2\pi)^{\frac{d}{2}}(h\times \textit{dist}_k(o))^d} \exp(-\frac{(\textit{reachability}_k(p \leftarrow o))^2}{2 \times (h \times \textit{dist}_k(o))^2})}{|\textit{N}_k(p)|}
\end{align}
In the above \eqref{lde}, $h$ represents a fixed bandwidth. $\textit{LDF}(p)$ is calculated as
\begin{align} \label{ldf}
\textit{LDF}(p) = \frac{\sum_{o \in \textit{N}_k(p)} \frac{\textit{LDE}(o)}{|\textit{N}_k(p)|}}{\textit{LDE}(p) + c\times \sum_{o \in \textit{N}_k(p)} \frac{\textit{LDE}(o)}{|\textit{N}_k(p)|}}
\end{align}
In \eqref{ldf}, $c$ is a scaling constant. The higher the value of $\textit{LDF}(p)$, the more likely it is to be an outlier. The kernel density estimator chosen by \glsxtrshort{ldf} does not integrate to 1, an essential property of any such estimator. To fix this, \glsxtrfull{kdeos}~\cite{kdeos} incorporates such estimators in \glsxtrshort{lof}, which preserves the said property for every point. $\textit{KDE}_k(p)$ for a given point $p$ is computed as follows.
\begin{align} \label{kde}
\textit{KDE}_k(p) = \frac{1}{N}\sum_{o \in \textit{N}_k(p)} \textit{K}_{\textit{h}(p)}(p - o)\text{, where }
\end{align}
$N$ is the dataset size, and $\textit{K}_{h(p)}(.)$ is the kernel function with bandwidth $\textit{h}(p)$, which is defined as follows.
\begin{align}
\textit{h}(p) = \min\{\textit{mean}_{o \in \textit{N}_k(p)}\textit{dist}(p,o), \epsilon\}
\end{align}

\glsxtrfull{loop}~\cite{loop} makes \glsxtrshort{lof} less amenable to the choice of $k$ for the estimation of local reachability density. To achieve this, \glsxtrshort{loop} conceptualizes a scoring method independent of data distribution and returns the probability of each point being an outlier.

\glsxtrfull{cof}~\cite{cof} redefines the notion of reachability as the average chaining distance (\textit{ac-dist}). It differentiates between points with low density and isolated points. Let $G=\langle p_{1}, p_{2},...,p_{N} \rangle$, which consists of all points. Assume two disjoint sets $A,B\subseteq G$ and $A\cap B=\phi$. \glsxtrshort{cof} defines the notion of distance between $A$ and $B$ as
\begin{align} \label{dist}
\textit{dist}(A,B)=\min\{\textit{dist}(x,y):x\in A\text{ }\&\text{ } y \in B\}
\end{align}
Let an ordered set of points $s=\langle p_{1}, p_{2},...,p_{r}\rangle$, termed as set based nearest path or an SBN-path on $G$ from $p_1$. An SBN-trail with respect to an SBN-path $s$ is a sequence $\langle e_{1}, e_{2}, ..., e_{r-1} \rangle$ such that for all $1\leq i \leq r - 1, e_i=(o_i, p_{i+1})$, where $o_i \in \langle p_1, p_2, ..., p_i\rangle$. So, \textit{ac-dist} from $p_1$ to $G-\{p_1\}$ is defined as
\begin{align}
\textit{ac-dist}_G(p_1) = \sum_{i=1}^{r-1}\frac{2(r-i)}{r(r-1)} \textit{dist}(e_i)
\end{align}
Equation \eqref{dist} is applicable to $e_i$ since $e_i$ is a sequence. Isolated points will have higher values of \textit{ac-dist} than points with low density. Now, \textit{\glsxtrshort{cof}} at $p$ with respect to its $k$-neighborhood is defined as
\begin{align}
\textit{COF}_k(p) = \frac{|\textit{N}_k(p)|\times\textit{ac-dist}_{\textit{N}_k(p)}(p)}{\sum_{o\in \textit{N}_k(p)}\textit{ac-dist}_{\textit{N}_k(o)}(o)}
\end{align}

\glsxtrfull{odin}~\cite{odin} utilizes directed \glsxtrshort{knn} graphs to identify outliers. Every point represents a vertex, and an edge connects a pair of nearest neighbors. For every point, there are exactly $k$ outgoing edges to its nearest neighbors. The resultant in-degree of the vertex is used to rank outliers. A point with a lower in-degree is more likely an outlier.

\glsxtrfull{fastabod}~\cite{fastabod} has been proposed as an outlier detection approach for points in high-dimensional spaces. It handles the curse of dimensionality by replacing the distance-based approach with an angle-based one. \textit{\glsxtrfull{abof}} is defined as the variance over the angles between the difference vectors of a point $p$ to the remaining points in a given dataset. \glsxtrshort{fastabod} approximates \textit{\glsxtrshort{abof}} and computes the variance in angles between the difference vectors of a point $p$ and $N_k(p)$ only. Points with lower values of \textit{\glsxtrshort{abof}} are potential outliers. \glsxtrfull{sod}~\cite{sod} solves the problem of identifying outliers in high-dimensional spaces by identifying relevant attributes. For every point $p$, it defines its reference set of points $\textit{R}(p)$. For $\textit{R}(p)$, it computes its variance along every attribute. The variance along an attribute $i$ is independent of other attributes. An attribute $i$ is relevant for $\textit{R}(p)$ if its variance is lower than the expected variance. The resultant relevant attributes define subspace hyperplane $\textit{H}(\textit{R}(p))$. The distance of $p$ from $\textit{H}(\textit{R}(p))$ is used to determine its degree of outlierness. A value near to $0$ indicates that point $p$ fits well into $\textit{H}(\textit{R}(p))$.

\glsxtrfull{hbos}~\cite{hbos} is an extremely fast unsupervised outlier detection approach. It assumes independence of features for faster computation. \glsxtrshort{hbos} constructs an individual histogram for every feature or dimension ($d$) in data. The height of each bin represents the density of points falling into that bin. These values are converted into probabilities by normalizing them using the dataset size. The outlier score for each data point is computed as
\begin{align}
\textit{HBOS}(p) = \sum_{i=1}^{i=d}\log\big(\frac{1}{\textit{hist}_i(p)}\big)\text{, where}
\end{align}
$\textit{hist}_i(p)$ is the density of point $p$ in dimension $i$, normalized by the number of samples.

The overall categorization of various methods can be summarized as follows~\ref{fig:chapter1::organization}:
\begin{figure}[!ht]
    \centering
\colorlet{ProcessBlue}{blue!50!cyan}
\begin{center}
    \begin{tikzpicture}[inner sep=2mm,
    ar/.style={->,>=stealth,thick},
    every node/.style={rectangle,rounded corners=2pt,
        drop shadow,draw=ProcessBlue,fill=ProcessBlue!35,thick,
        node distance=1.2cm 
    }]
    \node (n0)                                  {Unsupervised Outlier Detection Algorithms};
    \node (n1)      [below = of n0,xshift=-25ex,yshift=6ex]   {Statistical};
    \node (n2)      [below = of n0,xshift=15ex,yshift=0ex]   {Nearest-neighbor based};
    \node (n3)      [below = of n0,xshift=30ex,yshift=6ex]    {Sub-space based};
    \node (n4)      [below = of n1,xshift=10ex]     {HBOS~\cite{hbos}};
    \node (n5)      [below = of n2,xshift=-8ex,yshift=1ex]      {Global};
    \node (n6)      [below = of n2,xshift=8ex,yshift=1ex]      {Local};
    \node (n7)      [below = of n5,xshift=-8ex,yshift=6ex]      {$k$-NN~\cite{knn}};
    \node (n8)      [below = of n6,xshift=8ex,yshift=6ex]      {LOF~\cite{LOF}};
    \node (n9)      [below = of n5,xshift=-12ex,yshift=0ex]      {$k$-NN-\textit{weight}~\cite{knnw}};
    \node (n10)      [below = of n6,xshift=8ex,yshift=0ex]      {LIC~\cite{lic}};
    \node (n11)      [below = of n3,xshift=12ex]      {SOD~\cite{sod}};
    \foreach \ns/\ne in{0/1,0/2,0/3,1/4,2/5,2/6,5/7,6/8,5/9,6/10,3/11}
    \draw[ar] (n\ns) |- (n\ne);
    \end{tikzpicture}
\end{center}
    \caption{Overall categorization of well-known unsupervised anomaly detection algorithms. The three broad categories are statistical, sub-space based, and nearest-neighbor based.}
    \label{fig:chapter1::organization}
\end{figure}
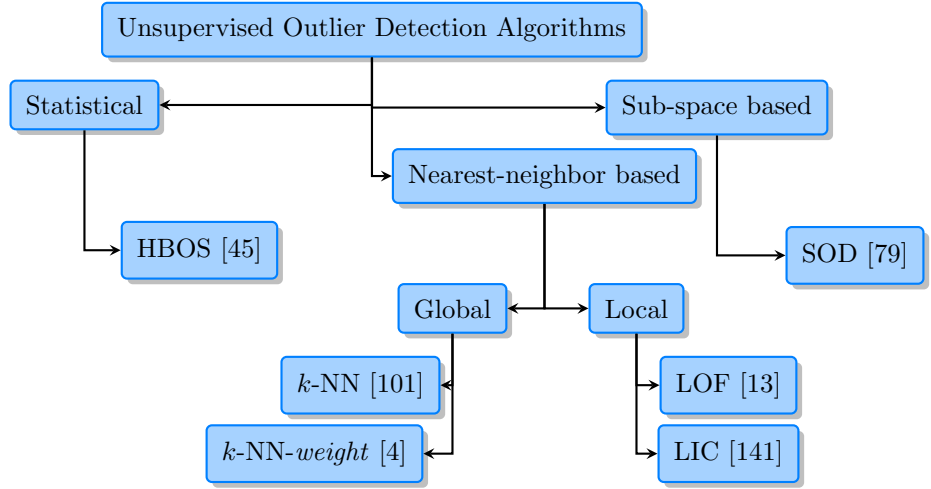

\subsection{Outlier detection: Motivation}
To date, several methods have been proposed that identify outliers. However, in general, methods do not scale with increasing number of samples $N$. Methods such as  \glsxtrshort{knn}~\cite{knn}, \glsxtrshort{odin}~\cite{odin}, \glsxtrshort{lof}~\cite{LOF}, Simplified-LOF~\cite{simplifiedlof}, \glsxtrshort{inflo}~\cite{inflo}, \glsxtrshort{loop}~\cite{loop}, \glsxtrshort{ldf}~\cite{ldf}, \glsxtrshort{lic}~\cite{lic}, DB-outlier score~\cite{dboutlierscore}, and \glsxtrshort{kdeos}~\cite{kdeos} require $O(N^2)$ to identify $k$ nearest neighbors (NNs). This may be reduced to $O(N\log N)$ if the dataset is indexed. \glsxtrshort{cof}~\cite{cof}, \glsxtrshort{ldof}~\cite{ldof}, and \glsxtrshort{fastabod}~\cite{fastabod} require an additional $O(k^2)$ computations per point. Therefore, they require an overall running time of $O(N\log N+N\times k^2)$. Very few methods scale well with increasing dimensions $d$. \glsxtrshort{sod}~\cite{sod} has an overall run-time complexity of $O(d \times N^2)$, where $d$ is the total number of features. This may be reduced to $O(d \times N\log N)$ if an indexed structure is used to find NNs. Also, there are too scarce methods with linear run-time complexity, such as Isolation Forest~\cite{iforest}, RS-Hash~\cite{rshash}, and \glsxtrshort{hbos}~\cite{hbos} (for fixed bin widths). For dynamic bin widths, \glsxtrshort{hbos} has $O(N\log N)$ complexity. \glsxtrshort{hbos} constructs a histogram for every feature independently and then combines them to determine the density of every point. However, it does not consider the possibility of irrelevant/redundant features.

The inability of the state-of-the-art algorithms to scale to large datasets poses a great challenge, especially for biological applications. One prevalent use-case is that of identification of minor cell types amongst thousands of single cells. These cell types have several profound applications, such as pathogenesis of cancer, angiogenesis in cancer and other disorders, etc. These rare cells correspond to the group of global outliers. 

There is another group of outliers called as local outliers, that are unusual only to a local section of the data~\cite{campello2015hierarchical}. A dataset may consist of local or global or a combination of outliers. Till date, there have been very limited efforts to quantify a given dataset on the basis of the presence of outlier sub-categories.       

\subsection{Outlier detection: Proposed technique}
We propose hashing based techniques to detect outliers in linear time. The proposed method uses an ensemble of different subspaces to determine outliers. \glsxtrfull{fire} is a \textit{sketching} based algorithm that detects global outliers. \glsxtrshort{fire} has been employed to detect rare cells in a single-cell study with approx. 68k cells expressed in approx. 32k space of genes. An extension of \glsxtrshort{fire}, FiRE.1, works well for the identification of local outliers as well. An extensive comparison of FiRE.1 has been shown with 18 state-of-the-art outlier detection algorithms on a diverse collection of 1000 datasets. We also propose a scoring criterion that assigns a value to a dataset based on it's outliers' local or global nature.

The notion of an outlier may also change with time, particularly in non-stationary environments. This is the motivation for the next chapter of the thesis, that focusses on detecting changes in data distributions over time.

\section{Concept drift detection}
\subsection{Concept drift detection: Definition and use cases}
During classification, a change in the distribution of data is referred to as the \textit{concept drift} problem~\cite{gama2014survey}. Let $p(c|x) = p(x|c)p(c)/p(x)$ be the Bayes posterior probability of a class that an instance $x$ belongs to. Then \textit{concept drift} can formally be defined as the one where $p_{t+1}(c|x) \neq p_{t}(c|x)$, i.e., the posterior probability changes over time. A shift in the likelihood of observing a data point $x$ within a particular class, and thus, alteration of class boundaries is referred to as \textit{real concept drift}. This involves \textit{replacement learning}. A concept drift without an overlap of true class boundaries is referred to as \textit{virtual concept drift}. This requires \textit{supplemental learning}. An example of \textit{real concept drift} involves a change in user's interests in news articles, while the underlying distribution of news articles remains the same. On the other hand, \textit{virtual concept drift} refers to the scenario when the user's interests remain the same, but the underlying distribution of new articles changes with time. A drift may be \textit{incremental}, \textit{abrupt} or \textit{gradual}.

\subsection{Concept drift detection: Existing techniques}
In this section, we discuss some of the widely used ensemble learners for drift detection.

\textit{Learn++}~\cite{learn++} is an \textit{Adaboost}~\cite{freund1997decision,schapire1998boosting,schapire1990strength} inspired algorithm, that generates an ensemble of weak classifiers, each trained on a different distribution of training samples. The multiple classifier outputs are combined using weighted majority voting. For incremental learning, Learn++ updates the distribution for subsequent classifiers such that instances from new classes are given more weights.

While Learn++ is suitable for incremental learning, $\textit{Learn}^{++}\textit{.NSE}$~\cite{learnNse} employs a passive drift detection mechanism to handle non-stationary environments. If the data from a new \textit{concept} arrives and is misclassified by the existing set of classifiers, then a new classifier is added to handle this misclassification. $\textit{Learn}^{++}\textit{.NSE}$ is an ensemble-based algorithm that uses weighted majority voting, where the weights are dynamically adjusted based on classifiers' performance.

\textit{Accuracy-Weighted Ensemble}~\cite{accuracyWeightedEnsemble} is an ensemble of weighted classifiers where the weight of a classifier is inversely proportional to its expected error.

\textit{Additive Expert Ensemble}~\cite{additiveExpertEnsemble} uses a weighted vote of experts to handle \textit{concept drift}. The weights of the experts that misclassify a sample are decreased by a multiplicative constant $\beta \in [0,1]$. In case the overall prediction is incorrect, new experts are added as well.

\textit{Dynamic Weighted Majority (DWM)}~\cite{dwm} dynamically adds and removes classifiers to cope with \textit{concept drift}.

Online bagging and boosting~\cite{oza2005online} ensemble classifiers are used in combination with different algorithms such as \textit{\glsxtrfull{adwin}}~\cite{adwin} for \textit{concept drift} detection. \glsxtrshort{adwin} dynamically updates the window by growing it when there is no apparent change, and shrinking it when the data evolves. In general, bagging is more robust to noise than boosting~\cite{dietterich2000experimental,oza2001experimental}. Henceforth, we discuss methods based on online bagging, that approximates batch-bagging by training every base model with $K$ copies of a training sample, where $K \sim \textit{Poisson(1)}$. We refer to the combination of online bagging classifiers with \glsxtrshort{adwin} as \textit{Online Bagging-ADWIN}.

\textit{Leveraging bagging}~\cite{leverageBagging} exploits the performance of bagging by increasing randomization. Resampling with replacement is employed in online bagging using \textit{Poisson(1)}. Leveraging bagging increases the weights of this resampling by using a larger value of $\lambda$ to compute the \textit{Poisson} distribution. It also increases randomization at the output by using output codes.

\textit{Online SMOTE Bagging}~\cite{onlineSmotebagging} oversamples a minor class by using SMOTE~\cite{chawla2002smote} at each bagging iteration to handle class imbalance. SMOTE generates synthetic examples by interpolating minor class examples.

\textit{\glsxtrfull{arf}}~\cite{arf} is an adaptation of Random Forest~\cite{breiman2001random} for evolving data streams. It trains a background tree when there is a warning and replaces the primary model if drift occurs.

\subsection{Concept drift detection: Motivation and proposed technique}
For drift detection in non-stationary environments, there are mainly two constraints: memory and time. For specific applications such as those related to IoT devices, there is only a limited main memory. A device can accommodate a limited amount of data, and the model size must be reasonable as well. To accommodate drift, the model must be able to update its parameters, and hence, predict (accordingly) almost instantaneously. For any model to be acceptable, it must have a reasonable performance within constraints (as mentioned above). The most popular methods for the detection of drifts in streaming environments have been ensemble learners~\cite{learnNse, learn++, arf, dwm, onlineSmotebagging, leverageBagging, accuracyWeightedEnsemble, additiveExpertEnsemble}. An ensemble method selectively retains few learners to maintain prior knowledge (since new information may be noise), discards, and adds new learners to accommodate new information.

We propose a fast, ensemble learner \textit{Enhash}, based on the concept of hashing that inserts an incoming sample, updates the model parameters, and predicts the class in $O(1)$ time. We show empirically on 6 artificial and 4 real datasets, the superiority of the proposed method Enhash in terms of both speed and performance. The datasets consist of various kinds of drift. The closest competitors of Enhash in terms of performance had significantly large requirements for RAM-hours. Thus, Enhash offers a perfect mix of good performance and superior run times.

\section{Organization of the thesis}
In this thesis, we begin with the problem of identifying rare samples in a dataset. These may be outliers in some contexts and ignored, but in many scenarios, their identification is of critical use. One high impact application is identifying rare cells in sc-RNA data corresponding to thousands of cells. Circulating tumour cells, cancer stem cells, endothelial progenitor cells, antigen-specific T cells, invariant natural killer T cells, etc. are examples of rare cells, whose identification is challenging yet potentially of great significance. For example, Circulating Tumor Cells (CTCs) offer unprecedented insights into the metastatic process with real-time leads for clinical management ~\cite{Krebs2010351}. Chapter \ref{chapter::2} discusses our algorithm for the identification of rare samples, with a focus on sc-RNA data by way of illustration. We propose a high-speed algorithm named as \glsxtrshort{fire} in chapter~\ref{chapter::2}. \glsxtrshort{fire} identifies dendritic cell types amongst several thousands of human blood cells. In a single cell sequencing dataset, local outliers act as noise. We propose an extension of \glsxtrshort{fire}, FiRE.1, in chapter~\ref{chapter::3} that identifies both local and global outliers. Chapter~\ref{chapter::3} assesses the relative performance of several well-known outlier detection algorithms. The methods are compared from the perspective of their ability to detect local and global outliers as well. For the identification of outliers in non-stationary environments, we propose Enhash in chapter~\ref{chapter::4}. Enhash accommodates the concept drift and identifies outliers when samples arrive in an online fashion. The thesis concludes with a chapter containing concluding remarks and scope for future work.


\chapter{Identification of Rare Events\protect\footnote{The work presented in this chapter has been published as a research paper titled ``\textit{Discovery of rare cells from voluminous single cell expression data}'' in Nature Communications (2018).}}\label{chapter::2}

\section{Introduction}
Rare events are outliers, that are not noises or erroneous readings but are generated from co-incidental or occasional processes that are valid in the problem/domain contexts. Events generated via such phenomenons tend to cluster themselves. The clusters of such events are fewer and warrant different treatment from a generic outlier. Rare event identification finds widespread application, e.g. detection of credit card fraud or credit-to-GDP gap~\cite{coffinet2019detection} in the financial domain; or identification of malignant cells~\cite{kraeft2000detection}, and identification of pandemic-causing virus strains, in the medical domain. Sociology and astronomy have benefited from rare event detection techniques as well~\cite{johnson2010future, zhang2004outlier}. 

In biology, one important use case of rare event detection is minor cell type identification. These minor cell types form small clusters and do not merely exist as singletons. These minor/ rare cell types have profound significance in biological processes. In conventional bioinformatics analysis, the abundant cell types overshadow the presence of minor cell types, and they are not included in the downstream analysis. Thus, depriving the studies of benefiting from the information provided by rare cells. We, therefore, propose an algorithm that overcomes the shortcomings of the current state-of-the-art technologies and detects rare cell types. 

Transcriptome analysis of individual cells is now possible thanks to the unrelenting advancement of technology over the past few years~\cite{Wagner20161145}. Cells, the fundamental building blocks of complex tissue, are formed in the presence of a variety of stimuli that influence their identity. In contrast to bulk RNA-sequencing, \glsxtrfull{scRNA-seq} examines the average expression-signature of genes in a population of heterogeneous cells at the level of individual cells. 

Processing hundreds of thousands of single cells is necessary for thorough characterization of all main and minor cell types in a complex tissue~\cite{Shapiro2013618}. In other words, the likelihood of detecting small cell subpopulations in a tissue is improved by using bigger sample numbers. It is mostly due to failure at the synthesis step, which prevents a significant percentage of transcripts that are cell-type specific from being discovered during sequencing. As a result, cell types represented by a small number of cells frequently fall short of having a significant impact on the cell type detection regime. The recent development of droplet-based transcriptomic systems has made it possible to profile tens of thousands of individual cells simultaneously for a substantially lower cost per cell. Numerous research has been published to date, with reported transcriptomes with ranging from thousands to hundreds of thousands of cells~\cite{Zheng,dropSeq,campbell2017molecular,srivastava2018cellatlassearch,lspca}. 

Rare cell discovery has been a commonplace element in the pipeline for downstream analysis since the development of single-cell transcriptomics.
Minor cell types in an organism are represented by rare cells. Even a singularity (outlier cell) may warrant attention when there are hundreds of profiled cells. However, the emphasis now turns to the finding of minor cell-types rather than merely singletons due to the rise in throughput capacity. Circulating tumour cells, cancer stem cells, endothelial progenitor cells, antigen-specific T cells, invariant natural killer T cells, etc. are examples of rare cell types. Rare cell populations play a significant role in the pathogenesis of cancer, modulating immunological responses, angiogenesis in cancer and other disorders, etc. despite their low abundance. Antigen-specific T cells are crucial to the formation of immunological memory~\cite{Slansky2003,Altman199694,manzo2015antigen}. \glsxtrfull{EPCs}, which originate from the bone marrow, have proven to be reliable biomarkers of tumor angiogenesis~\cite{kuo2012dynamics, cima2016tumor}. Stem cells have an ability to replace damaged cells, and to treat diseases like Parkinson's, diabetes, heart diseases, etc.~\cite{Jang200545}. \glsxtrfull{CTCs} offer unprecedented insights into the metastatic process with real-time leads for clinical management~\cite{Krebs2010351}.

Algorithms for detecting rare cell transcriptomes are scarce. Prominent among these are RaceID~\cite{raceId} and GiniClust~\cite{giniClust}. For the purpose of identifying outlier expression patterns, RaceID employs a parametric statistical technique that is computationally intensive. It uses unsupervised clustering as an intermediate step to define populous cell types, which in turn are used to determine outlier events (cells). GiniClust, on the other hand, employs a rather simple two-step algorithm. It starts by choosing informative genes based on the Gini index~\cite{gini1912variabilita}. It then applies a density-based clustering method, DBSCAN~\cite{ester1996density}, to discover outlier cells. Notably, to discriminate between major and minor cell types, both RaceID and GiniClust employ clustering. In fact, these algorithms involve the computation of pair-wise distances between cells. Both of these algorithms are memory inefficient and sluggish for huge \glsxtrshort{scRNA-seq} data due to a number of important design decisions. 

We propose \glsxtrshort{fire}, a conspicuously fast algorithm, to calculate the density around each subjected multi-dimensional data point. The workhorse algorithm used to accomplish this is Sketching ~\cite{rankLsh,lv2006ferret}. \glsxtrshort{fire}, which, in contrast to other methodologies, assigns a rareness score to each sample. In the context of rare cells, this is a score based on each cell's individual expression profile, thus giving the user a choice for focusing downstream analyses only on a small set of potentially rare cells. However, it must be noted that we use clustering to group these rare cells. We used the dropClust~\cite{Sinha170308} technique to cluster these potential rare cells detected by FiRE. Outlier cells, if any, get submerged into the minor cell clusters since dropClust does not give any special treatment to outliers. Although, dropClust itself removes poor quality cells and genes.

On a variety of real and simulated datasets, we assessed \glsxtrshort{fire}. We also demonstrated the efficiency of \glsxtrshort{fire} in delineating human blood dendritic cell subtypes using a large \glsxtrshort{scRNA-seq} data of human blood cells. 

\section{Overview of FiRE}

As a preliminary step for identifying unusual cells, both RaceID and GiniClust make use of clustering in some way. By its very nature, clustering frequently depends on a variety of delicate characteristics and performs poorly when data point densities vary. Another major problem is to decide the resolution of group identities. Multi-level clustering is frequently necessary since tiny clusters are frequently missed during the initial pass~\cite{campbell2017molecular}. This occurs because other significant cell types affect a dataset's expression variance. We questioned whether it was possible to create a novel, unified approach that directly estimates a cell's rarity without using clustering (multidimensional data point).

\begin{figure}[ht]
\centering
\includegraphics[width=\linewidth]{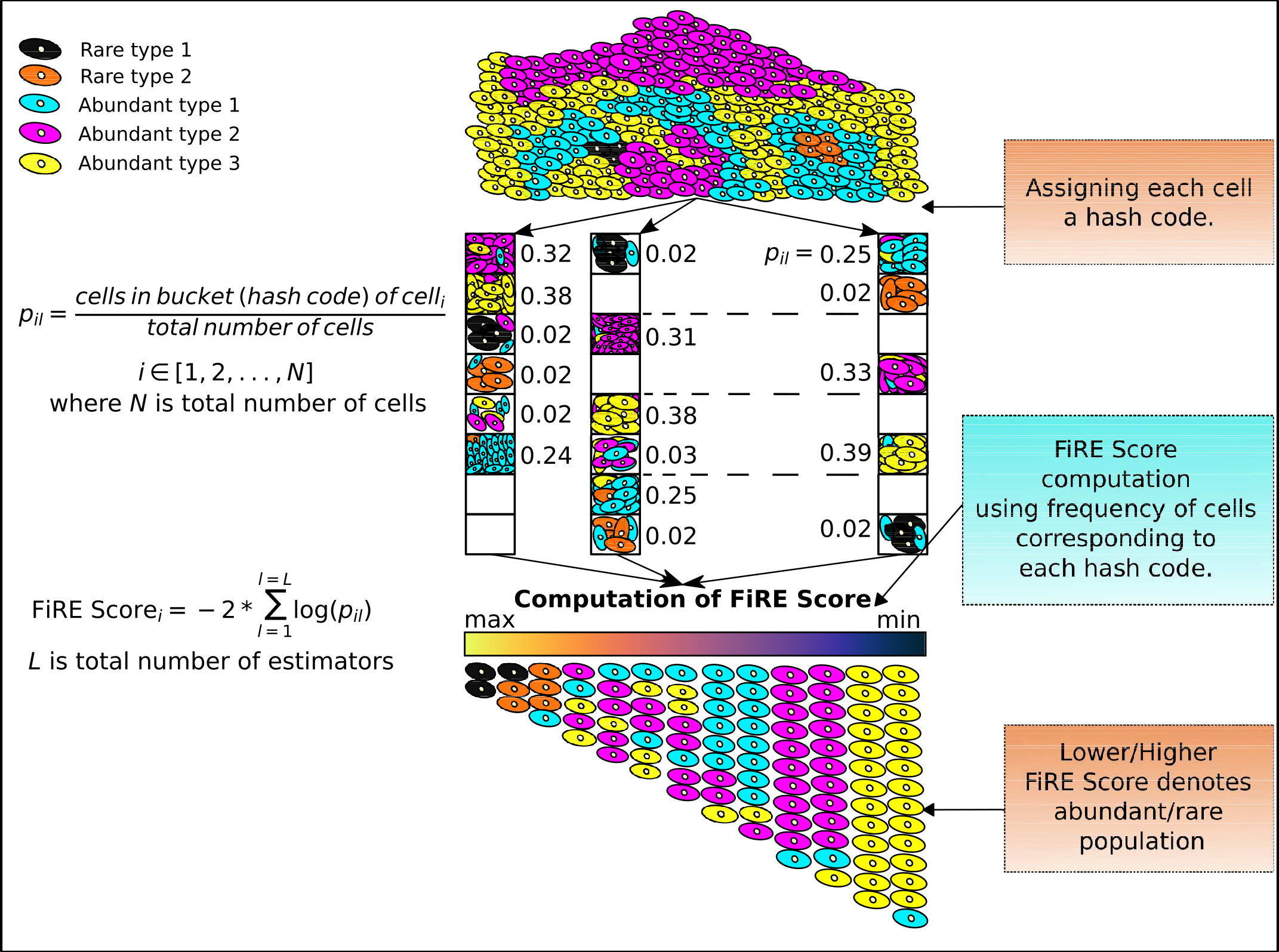}
\caption{Overview of \glsxtrshort{fire}. The first step is to assign each cell to a hash-code. As numerous similar cells can share the same hash-code, it is possible to think of a hash-code as an imagined bucket. The phase of creating the hash-code is repeated $L$ times to test the reliability of rarity estimates. The chance that any point will fall into the bucket of a given cell $i$ and estimator $l$ is calculated as $p_{il}$. These probabilities are combined in the algorithm's second phase to get an estimate of how rare each cell is.}
\label{fig:2_overview}
\end{figure}

To circumvent the above issues, we propose \glsxtrshort{fire} to identify rare cell types. Design of \glsxtrshort{fire} is inspired by the observation, that rareness estimation of a particular data point is the flip side of measuring the density around it. The algorithm capitalizes on the Sketching technique~\cite{rankLsh,lv2006ferret}, a powerful technique for low dimensional encoding of a large volume of data points. It works by randomly projecting points to low dimensional bit signatures (hash code), such that the weighted $L_1$-distance between each pair of points is approximately preserved. The computation involved in the creation of hash codes is linear w.r.t. the number of individual transcriptomes. A hash code can be imagined as a bucket, that tends to contain data points which are close by in the concerned hyper-dimensional space. The cell originating from a large cluster shares its bucket with many other cells, whereas a rare cell shares its bucket only with a few. To this end, \glsxtrshort{fire} uses the populousness of a bucket as an indicator of the rareness of its resident data points. To ward off biases, \glsxtrshort{fire} uses several such rareness estimates, to arrive at a consensus rareness score for each of the studied cells. This score is termed as the FiRE-score. The section~\ref{sec::stepsFire} contains an elaborate explanation of the various steps involved in \glsxtrshort{fire}. Figure~\ref{fig:2_overview} depicts a visual interpretation of \glsxtrshort{fire}.


\subsection{Steps involved in FiRE}\label{sec::stepsFire}


Let $X\in\mathbb{R}^d$ represent an input space. Let $S$ contains $N$ data points drawn independently and identically from the distribution $D$ over $X$. Assume, $U[a, b]=\frac{1}{b-a}$ represents a uniform distribution. Let $S_{j}^{(i)}$ represents $j$-th feature of $i$-th sample. Let, $\forall j\in\{1,...,d\}$, $mi_j=\min_{i\in\{1,...,N\}}S_{j}^{(i)}$ and $ma_j=\max_{i\in\{1,...,N\}}S_{j}^{(i)}$. Let, $L$, $M$, and $H$ are integers and $H$ is a prime number as well.

\glsxtrshort{fire} is a two-step algorithm. In the first step, it follows the Sketching~\cite{sizingsketches} process. \glsxtrshort{fire} randomly selects $M$ values with replacement from $\{1,...,d\}$. Say $\_ind^{(l)}$, where $1 \leq l \leq L$, represents this set. \glsxtrshort{fire} projects $S$ onto $\_ind^{(l)}$ for every $l$. Say projected data is $P_l$. For a $P_l$, \glsxtrshort{fire} generates a random threshold vector $\_th^{(l)}$ where $\_th^{(l)}_{j} \sim U[mi_{\_ind^{(l)}_{j}}, ma_{\_ind^{(l)}_{j}}]$, where $1\leq j \leq M$. Every data point in $P_l$ is then converted into a bit stream by thresholding it with $\_th^{(l)}$. Let the thresholded data be denoted by $T_{l}$; then, bit $j$ of $i$-th sample is generated as follows.
\begin{align}
(T_{l})^{(i)}_{j} =
\begin{cases}
1, (P_{l})^{(i)}_{j} \geq \_th^{(l)}_{j}\\
0, \text{otherwise}
\end{cases}
\end{align}

Further, \glsxtrshort{fire} samples $M$ elements with replacement from $\mathbb{I}^{+} \bigcup \{0\}$. Say, $\_pr^{(l)}$ represents this set. Hash index $\_index^{(l)}_{i}$ for each data point is computed by taking the dot product of their bit vector $T_{l}^{(i)}$ with $\_pr^{(l)}$, and modulo it with $H$.
\begin{align}
\_index^{(l)}_{i} = \sum_{j=1}^{j=M}\Big((T_{l})^{(i)}_{j} * \_pr^{(l)}_{j}\Big)\%H
\end{align}

In the second step, the neighborhood information is collected for every data point. All data points which map to the same hash index as that of element $i$ construct its neighborhood. Probabilistic neighborhood $\textit{neighborhood}^{(l)}_{i}$ is defined as
\begin{align}\label{eq::neighborhood}
\textit{neighborhood}^{(l)}_{i} = \frac{\text{Total data points with hash index }\_index^{(l)}_{i}}{N}
\end{align}
For every data point, this information is combined across $L$ and \textit{FiRE-score} of each data point, \textit{FiRE-score}$_i$, is computed as follows.
\begin{align}\label{eq::score}
\textit{FiRE-score}_i = -2\sum_{l=1}^{L}\log \textit{neighborhood}^{(l)}_{i}, \forall i \in \{1,...,N\}
\end{align}
The sparser the neighborhood of a point, the higher the value of \textit{FiRE-score}. Hence, points with higher values of \textit{FiRE-score} represent potential outliers. Algorithm~\ref{train} gives the pseudo-code of \glsxtrshort{fire}.

\begin{algorithm}[!ht]
    \caption{FiRE}\label{train}
    \begin{algorithmic}[1]
        \Function{FiRE}{\textbf{Input}: dataset} \# nSamples x nFeatures
            \State $\textit{nSamples} \gets \text{Number of data samples} (cells)$
            \State $\textit{nFeatures} \gets \text{Number of features} (genes)$
            \State $\textit{X[nSamples][nFeatures]} \gets \text{dataset}$
            \State $\textit{L} \gets \text{Number of runs}$
            \State $\textit{M} \gets \text{Number of dimensions to be sampled}$
            \State $\textit{H} \gets \text{Number of bins}$
            \State $\textit{bins[L][H][.]} \gets \text{Keeps the bin details across runs}$
            \State $\textit{neighborhood[nSamples][L]} \gets \text{Stores the size of the neighborhood of each sample}$
            \State $\textit{scores[nSamples]} \gets \text{Stores the score for each sample}$
            \State $\textit{mi} \gets \text{min(X)}$
            \State $\textit{ma} \gets \text{max(X)}$
            \For{$i=1$ to $L$}
                \State $\textit{\_index[nSamples]} \gets [0]\textit{ \# Keeps the bin index for all the sample points for the current run.}$
                \For{$j=1$ to $M$}

                    \State $\textit{\_ind} \gets \textbf{\textit{randomInt}}(low=1, high=nFeatures, dist="Uniform")$
                    \State $\textit{\_th} \gets \textbf{\textit{randomFloat}}(low=mi, high=ma, dist="Uniform")$
                    \State $\textit{\_pr} \gets \textbf{\textit{randomInt}}(low=1, high=maxInt, dist="Uniform")$

                    \For{$k=1$ to $nSamples$}
                        \State $\_v \gets 0$
                        \If {$X[k][\_ind] >= \_th$}
                            \State $\_v \gets 1$
                        \EndIf
                        \State $\_index[k] \gets \_index[k] + \_pr * \_v$
                    \EndFor
                \EndFor
                \For{$k=1$ to $nSamples$}
                    \State $\_index[k] \gets \_index[k] \% H$
                    \State $bins[i][\_index[k]].append(k)$
                \EndFor
            \EndFor

            \For{$i=1$ to $L$}
                \For{$j=1$ to $H$}
                    \State $\_b \gets bins[i][j]$
                    \State $\_l \gets \textbf{\textit{length}}(\_b)$
                    \For{$k=1$ to $\_l$}
                        \State $neighborhood[\_b[k]][i] \gets \log(\_l / nSamples)$
                    \EndFor
                \EndFor
            \EndFor
            \For{$k=1$ to $nSamples$}
                \State $\_t \gets 0$
                \For{$i=1$ to $L$}
                    \State $\_t \gets \_t + neighborhood[k][i]$
                \EndFor
            \State $scores[k] \gets -2*\_t$
            \EndFor
        \State \Return \textit{scores}
        \EndFunction
    \end{algorithmic}
\end{algorithm}

\section{Results}

\subsection{Experimental setup}\label{sec::setupFire}
\subsubsection{Description of datasets}\label{sec::dataFire}
For the various analyses, we used four publicly available \glsxtrshort{scRNA-seq} datasets.

For a simulation experiment of artificially planted rare cells, we used 293T and Jurkat cells data containing a total of $\sim$3200 cells, with an almost equal number of representative transcriptomes of each type. The cells were mixed \textit{in vitro} at equal proportions. Authors of the study resolved the cell types bioinformatically exploiting their \glsxtrshort{SNV} profiles~\cite{Zheng}.


We used a large scale \glsxtrshort{scRNA-seq} data containing expression profiles of $\sim$68k \glsxtrfull{PBMCs}, collected from a healthy donor~\cite{Zheng}. Single cell expression profiles of 11 purified subpopulations of \glsxtrshort{PBMCs} were used as a reference for cell type annotation. Both 293T-Jurkat cells and \glsxtrshort{PBMC} datasets are available for download from 10xgenomics.com (\url{https://support.10xgenomics.com/single-cell-gene-expression/datasets}).

We applied \glsxtrshort{fire} on a publicly available $\sim$2.5k mouse \glsxtrfull{ESCs} data~\cite{giniESC}(GSE65525). Mouse embryonic cells were sequenced at different points after the removal of leukemia inhibitory factor (LIF). Similar to Jiang \textit{et al.}~\cite{giniClust}, we used Day 0 data where stem cells were undifferentiated. Data contained a total of 2509 cells.

Our fourth \glsxtrshort{scRNA-seq} data contained single cell expression profiles of mouse intestinal organoids~\cite{raceId} (GSE62270). A set of 288 organoid cells were randomly selected and sequenced using a modified version of the cell expressions by linear amplification and sequencing (CEL-seq) method. UMI identifiers were used to count transcripts.

\subsubsection{Data preprocessing}

Mouse \glsxtrshort{ESCs} and mouse small intestine datasets were screened for low quality cells. For mouse \glsxtrshort{ESC} data, cells having more than 1800 detected genes were selected for analysis. For the intestine dataset, the cutoff for the number of detected genes was set at 1200. The remaining data were already filtered.

For each dataset, genes which had a read count exceeding 2 in at least three cells, were retained for downstream analysis. Each \glsxtrshort{scRNA-seq} data was normalized using median normalization. The thousand most variable genes were selected, based on their relative dispersion (variance/mean) with respect to the expected dispersion across genes with similar average expression~\cite{Zheng}. The normalized matrix was then $\log_2$ transformed after the addition of 1 as a pseudo count.


\subsubsection{IQR-thresholding-criteria for selection of cells for further downstream analysis}\label{sec::iqr}

\glsxtrshort{fire} marks a cell as rare if its FiRE-score is $\ge \text{q}3 + 1.5 \times \text{IQR}$, where q3 and IQR denote the third quartile and the
inter-quartile range (75th percentile - 25th percentile), respectively, of the number of \glsxtrshort{fire} scores across all cells.

\subsubsection{F$_1$ score computation for the simulation study}\label{sec::f1Score}
Both RaceID~\cite{raceId} and GiniClust~\cite{giniClust} provide a binary prediction for rare cells. The contamination parameter in \texttt{scikit-learn} package implementation of \glsxtrshort{lof} gives a threshold for the identification of outliers. In a two-class experiment (293T and Jurkat cells), it is straightforward to construct a confusion matrix. The F$_1$ score on a confusion matrix can easily be computed as follows.
\[ \text{F}_1\text{ score} = 2\frac{\text{precision} \times \text{recall}}{\text{precision} + \text{recall}}\] For the simulation experiment, rare cells were considered ones whose \glsxtrshort{fire} scores satisfied the IQR-thresholding-criterion.

For all algorithms, the F$_1$ score has been calculated with respect to the minor population of the Jurkat cells.

\subsubsection{Parameter value selection for FiRE}\label{sec::paramFire}
The process of hashing cells to buckets is repeated $L$ times. For obvious reasons, a large value of $L$ ensures rareness estimates with low variance.

For every estimator, the Sketching technique randomly sub-samples a fixed set of $M$ features. While a very small choice of $M$ requires a commensurately large number of estimators, a very large $M$ might make the \glsxtrshort{fire} scores sensitive to noisy expression readings.

For all experiments, the hash table size, i.e., $H$ was set 1017881. It should be a prime number large enough to avoid unwanted collisions between dissimilar cells. In practice, $H$ is chosen to be a prime number greater than ten times of the number of items to be hashed.

\begin{figure}[!ht]
\centering
\includegraphics[width=\linewidth, keepaspectratio]{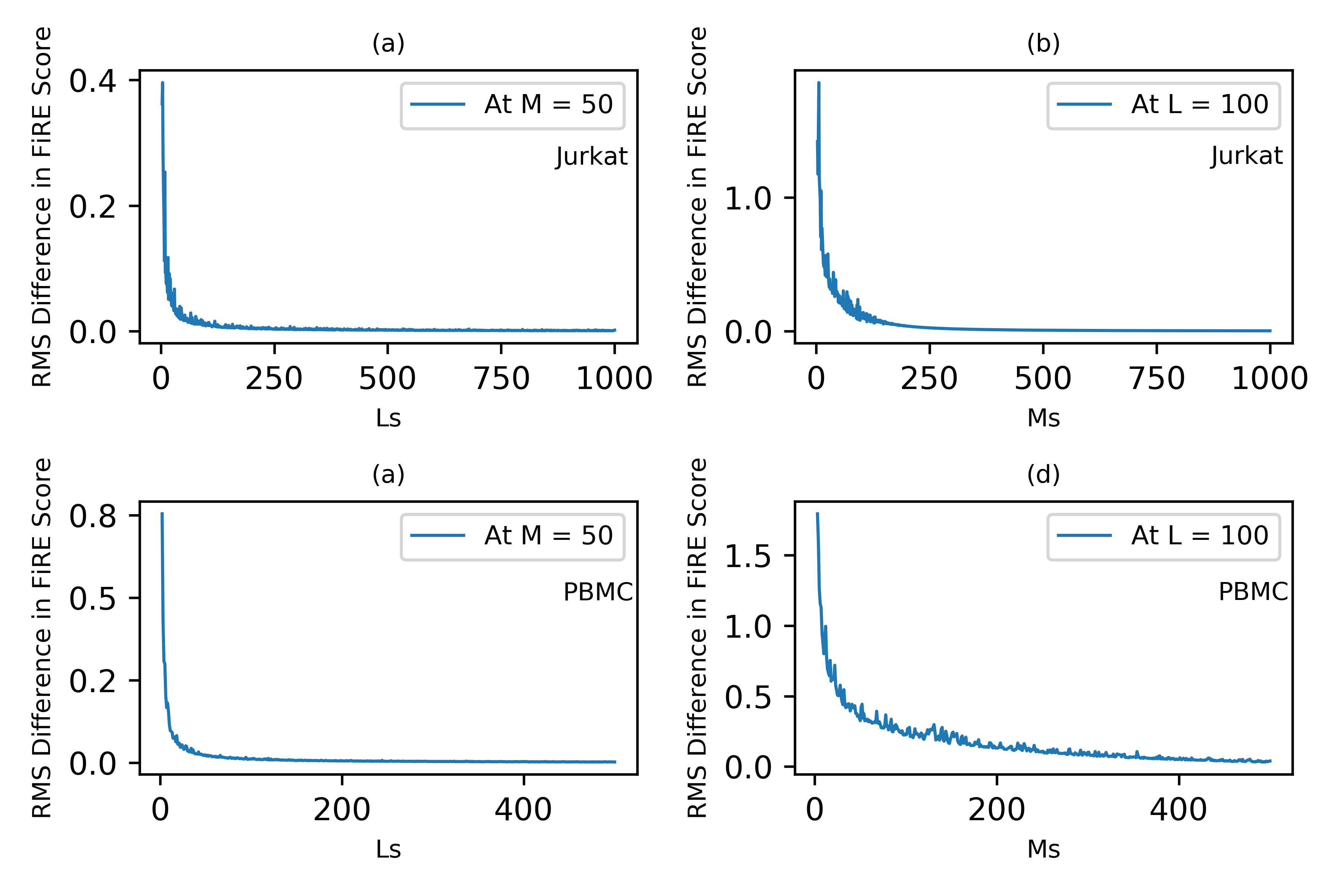}
\caption{Stability of \glsxtrshort{fire}. (a),(c) \glsxtrfull{RMS} difference in values of FiRE-score of every cell between two successive estimators. For calculation of \glsxtrshort{RMS}, FiRE-score is averaged across multiple seeds and normalized by the value of L. (b),(d) \glsxtrshort{RMS} difference in values of FiRE-score between two successive values of M. For calculation of \glsxtrshort{RMS}, FiRE-score is averaged across multiple seeds and normalized by the value of M. (a)-(b) \glsxtrshort{RMS} has been shown on a simulated dataset consisting of a mixture of Jurkat and 293T cells~\cite{Zheng}. (c)-(d) \glsxtrshort{RMS} has been shown on $\sim$68k Peripheral Blood Mononuclear Cells (PBMCs)~\cite{Zheng}.}
\label{fig:lrmse}
\end{figure}

On two independent datasets, we experimented with different values of $L$ and $M$. We found $L=100$ and $M=50$ were a reasonably good choice to be considered as default values of $L$ and $M$, respectively (Figure~\ref{fig:lrmse}).

\subsubsection{Identification of Differential Genes}\label{sec::DEgenes}
A traditional Wilcoxon's rank sum test was used to identify differentially expressed (\glsxtrshort{DE}) genes with an FDR cutoff of 0.05 and an inter-group absolute fold-change cutoff of 1.5. Fold-change values were measured between group-wise mean expression values of a given gene. We qualified a gene to be a cell-type specific one if it was found differentially up-regulated in a particular cluster, as compared to each of the remaining clusters.


\subsubsection{Simulation experiment to assess FiRE's sensitivity to DE genes}\label{sec::DEexp}

To analyze the sensitivity of \glsxtrshort{fire} to cell type identity, we generated an artificial \glsxtrshort{scRNA-seq} data using the \texttt{splatter} R package~\cite{splatter}. The following command was used to generate this data:

\texttt{splatSimulate(group.prob = c(\textcolor{green}{0.95}, \textcolor{green}{0.05}), method = \textcolor{red}{'groups'}, verbose = F, batchCells = \textcolor{green}{500}, de.prob = c(\textcolor{green}{0.4}, \textcolor{green}{0.4}), out.prob = \textcolor{green}{0}, de.facLoc=\textcolor{green}{ 0.4},\\ de.facScale = \textcolor{green}{0.8}, nGenes = \textcolor{green}{5000})}

The generated dataset had 500 cells and 5000 genes per cell. Out of the 500 cells, 472 cells represented the major cell-type, whereas 28 cells defined the minor one.

Genes for which expression counts exceeded 2 in at least three cells were considered for analysis. The filtered data was $\log_2$ transformed after adding 1 as a pseudo count. On the transformed data, differential genes were detected using Wilcoxon's rank sum test with an FDR cutoff of 0.05 and as an inter-group absolute value of fold change cut-off of 2.32 ($\log_2$(5)). The differentially expressed genes, which were 180 in number, were removed from the data and kept as a separate set. Genes with a p-value of more than 0.05, 2387 in number, were kept as a separate set of non-differential genes.



\subsection{FiRE discovers cells with varying degrees of rareness}

\begin{figure}[!ht]
\centering
\includegraphics[width=\linewidth]{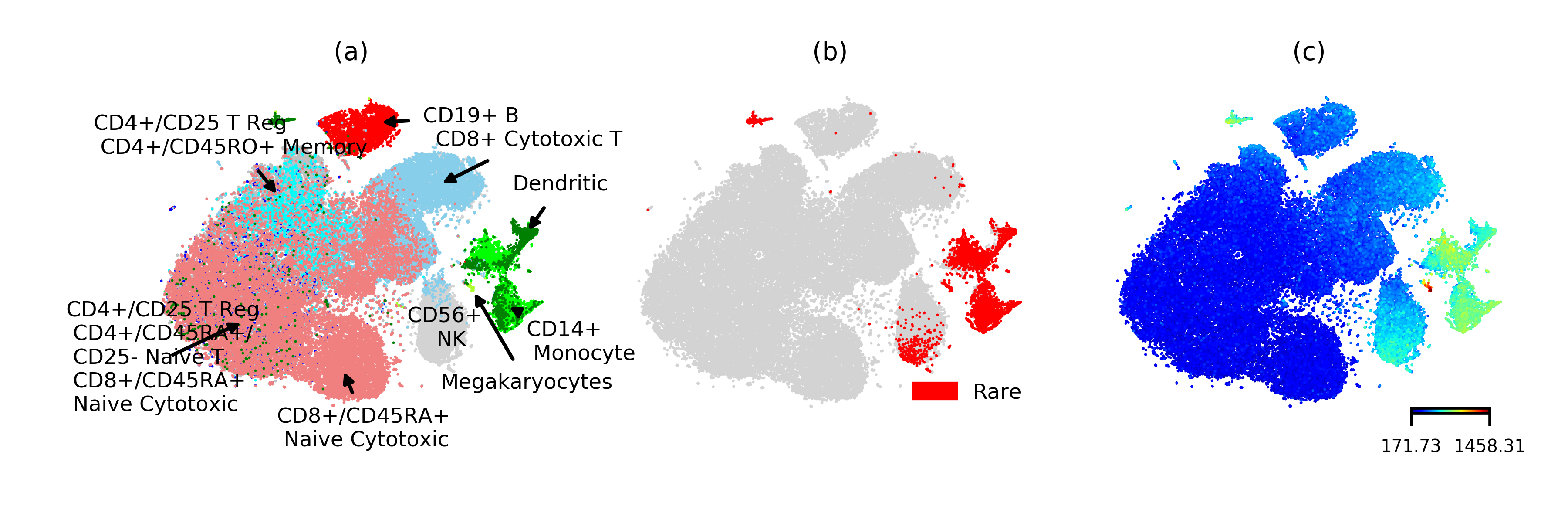}
\caption{Performance evaluation of \glsxtrshort{fire} on Peripheral Blood Mononuclear Cells (\glsxtrshort{PBMCs}). (a) \glsxtrshort{tsne} based 2D embedding of the data with color coded cluster identities as reported by Zheng and colleagues~\cite{Zheng}. (b) Rare population identified by \glsxtrshort{fire} using IQR-thresholding-criteria. (c) Heat map of FiRE scores for the individual \glsxtrshort{PBMCs}. The cluster of megakaryocytes (0.3\%), the rarest of all the cell types are assigned the highest FiRE scores.}
\label{fig:suppPBMC}
\end{figure}

\begin{figure}[ht]
\centering
\includegraphics[width=\linewidth]{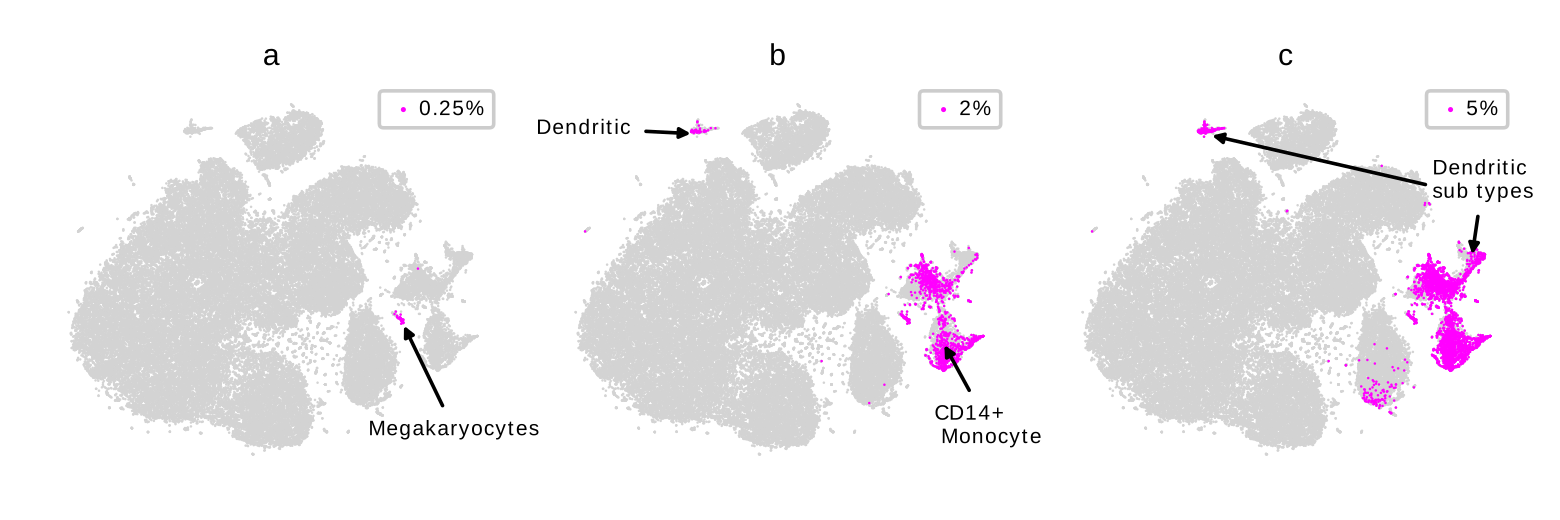}
\caption{In the $\sim$68k \glsxtrshort{PBMC} data~\cite{Zheng}, the appearance of minor cell populations with varying degrees of rarity is accompanied by a rise in the number of chosen rare cells. Figures~(a)-(c) demonstrate, respectively, the top 0.25\%, 2\%, and 5\% cells chosen based on FiRE scores.}
\label{fig:pbmcScore}
\end{figure}

\glsxtrshort{fire} assigns a continuous score to each cell, such that outlier cells and cells originating from the minor cell populations are assigned higher values in comparison to cells representing major sub-populations. A continuous score gives users the freedom to decide the degree of the rareness of the cells, to be further investigated. To illustrate this, we applied \glsxtrshort{fire} on a \glsxtrshort{scRNA-seq} data containing $\sim$68k Peripheral Blood Mononuclear Cells (\glsxtrshort{PBMCs}), annotated based on similarity with purified, well known immune cell sub-types~\cite{Zheng} (Section~\ref{sec::dataFire}). The authors of the study performed unsupervised clustering of the cells and annotated the clusters based on previously known markers (Figure~\ref{fig:suppPBMC}a). We overlaid FiRE scores on the 2D map reported as part of the study (Figure~\ref{fig:suppPBMC}c). The top 0.25\% highest FiRE scores exclusively corresponded to the smallest, unambiguously annotated cluster harboring megakaryocytes (Figure~\ref{fig:pbmcScore}a). Of note, megakaryocytes represent only 0.3\% of the entire set of the profiled cells. As we increased the proportion from 0.25\% to 2.0\% and subsequently 5.0\%, the next batches of minor cell sub-types made their way into the extended set of rare cells. These include sub-classes of monocytes and dendritic cell sub-types (Figures~\ref{fig:pbmcScore}b and~\ref{fig:pbmcScore}c). The case study highlights the utility of \glsxtrshort{fire} in discovering cells with varying degrees of rareness.

While a continuous score is helpful, sometimes a binary annotation about cell rarity eases out the analysis. To this end, we introduced a thresholding scheme using the properties of the score distribution (Section~\ref{sec::iqr}). Figure~\ref{fig:suppPBMC}b highlights cells in the $\sim$68k \glsxtrshort{PBMC} data, which are detected as rare going by the threshold based dichotomization. As expected, a majority of the detected rare cells originated from known minor cell types such as megakaryocytes, dendritic cells, and monocytes.

It should be noted that unlike GiniClust and RaceID, \glsxtrshort{fire} refrains from using clustering as an intermediate step to pinpoint the rare cells. Clustering is done in a later phase for delineating minor cell types from the detected rare cells.

\subsection{FiRE detects artificially planted rare cells with high accuracy}

\begin{figure}[ht]
\centering
\includegraphics[width=\linewidth]{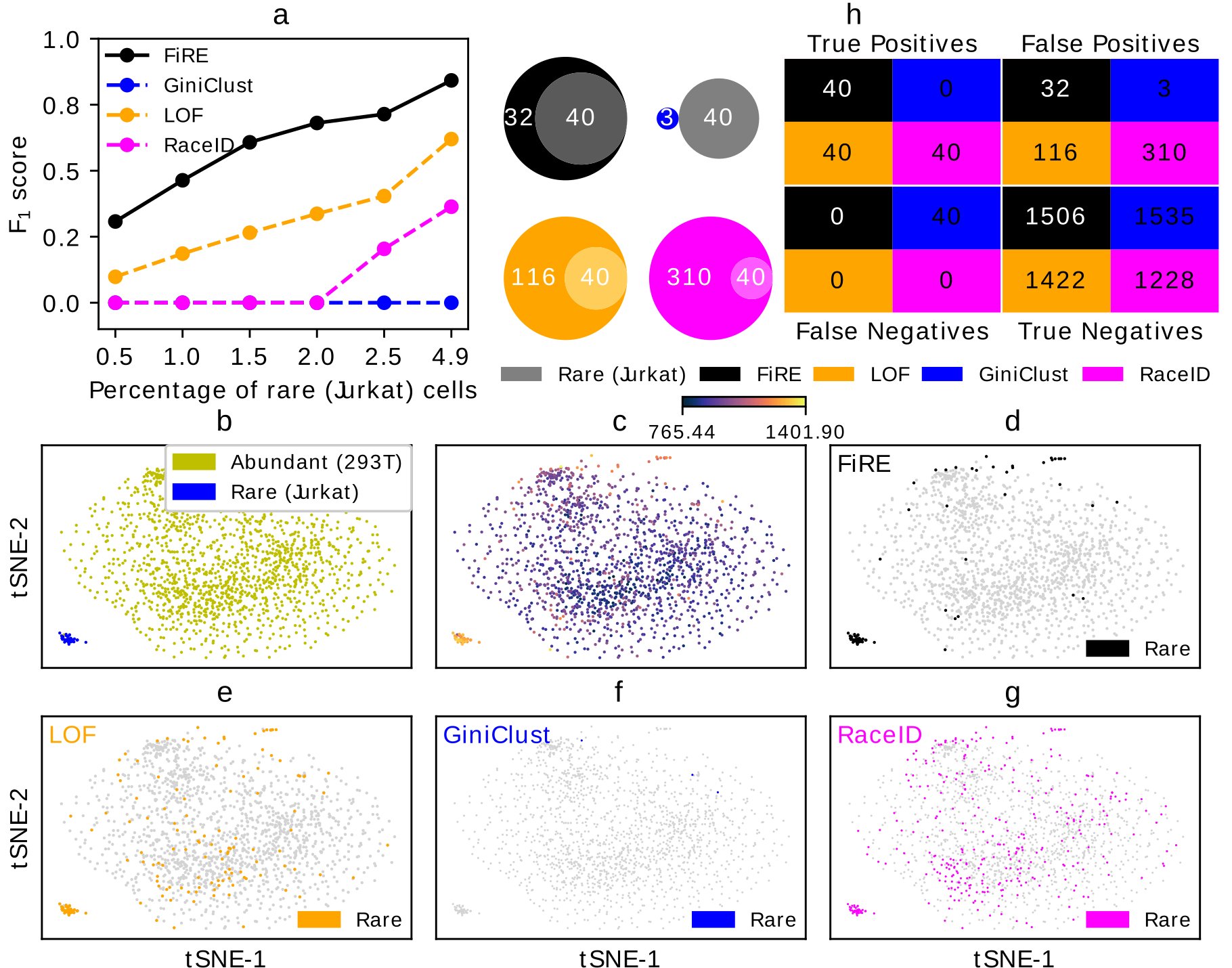}
\caption{Minor cell types' detectability in a simulated dataset with a mixture of Jurkat and 293T cells (known annotations)~\cite{Zheng}. (a) F$_1$ scores were determined relative to the rare (Jurkat) population, while bioinformatically altering the percentage of artificially planted rare cells. It is noteworthy that both \glsxtrshort{fire} and \glsxtrshort{lof}~\cite{LOF} use a threshold to their continuous scores for zeroing on the rare cells. On the other hand, GiniClust~\cite{giniClust} and RaceID~\cite{raceId} offer binary annotations for cell-rarity. (b) \glsxtrshort{tsne} based 2D embedding of the cells with color-coded identities. (c) FiRE-score intensities were displayed on the \glsxtrshort{tsne} based 2D map. Figures~(d)-(g) demonstrate the rare cells detected by various algorithms. (h) Congruence of methods with known annotations. Note: Results shown in Figures~(b)-(h) correspond to a rare cell concentration of 2.5\%.}
\label{fig:jurkat}
\end{figure}

We designed a simulation experiment to evaluate the performance of \glsxtrshort{fire} in the presence of ground truth information pertaining to cell-type identity. For this, we used a \glsxtrshort{scRNA-seq} data comprising 293T and Jurkat cells combined \textit{in vitro} in equal ratio (Section~\ref{sec::dataFire})~\cite{Zheng}. The authors exploited the \glsxtrfull{SNV} profile of each cell to determine its lineage. We considered this genotype based annotation scheme to be near confirmatory. With this data, we mimicked the rare cell phenomenon by bioinformatically diluting Jurkat cell proportion in the data. We varied the proportion of Jurkat cells between 0.5\% and 5\%. Besides GiniClust and RaceID, we compare \glsxtrshort{fire} with a rare event detection algorithm called Local Outlier Factor or LOF. \glsxtrshort{lof} is a widely used algorithm in the field of data mining. The performance of various methods was measured using F$_1$ score (Section~\ref{sec::f1Score}) with respect to the minor population of the Jurkat cells. F$_1$ score reflects the balance between precision and sensitivity. \glsxtrshort{fire} clearly outperformed \glsxtrshort{lof}~\cite{LOF}, RaceID~\cite{raceId} and GiniClust~\cite{giniClust} on each of the test-cases (Figure~\ref{fig:jurkat}a). Notably, RaceID and GiniClust report dichotomized predictions for rare cells, whereas \glsxtrshort{fire} and \glsxtrshort{lof} offer both continuous scores and binary prediction. \glsxtrshort{fire} implements an IQR-based-thresholding technique for the dichotomization (Section~\ref{sec::iqr}).

\begin{figure}[!ht]
\centering
\includegraphics[width=0.8\linewidth, keepaspectratio]{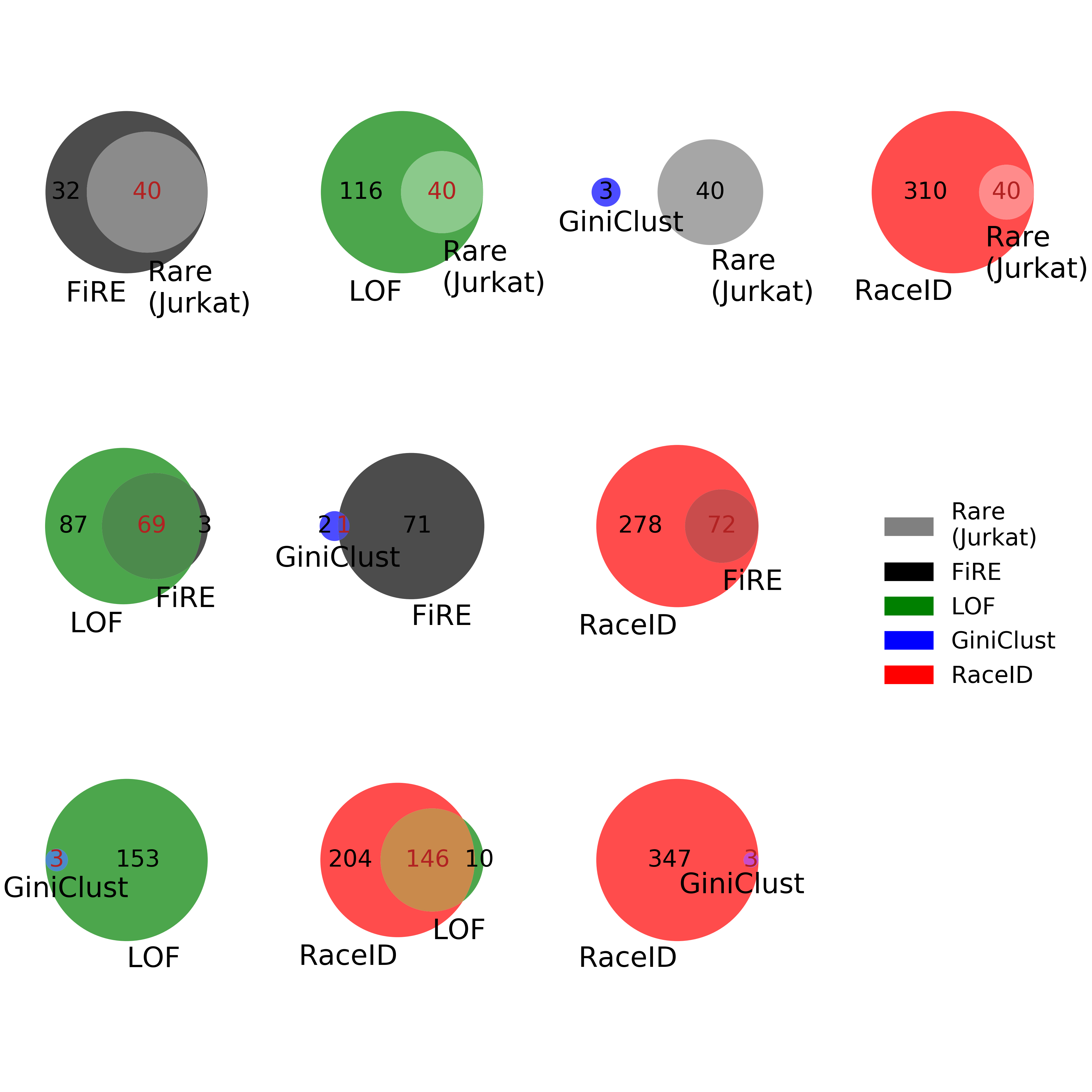}
\caption{Congruence of methods with known annotations and congruence between pairs of methods on a simulated, \glsxtrshort{scRNA-seq} data consisting of 293T and Jurkat cells mixed \textit{in vitro} in equal proportion~\cite{Zheng}.
}
\label{fig:jurkatVenn}
\end{figure}

We took a closer look at working of the methods at a rare cell concentration of 2.5\%. We found FiRE scores of the rare cells to be unambiguously higher compared to the abundant cell type (Figure~\ref{fig:jurkat}c). Figures~\ref{fig:jurkat}d-~\ref{fig:jurkat}g mark the rare cells detected by each of the algorithms. Among all algorithms, \glsxtrshort{fire} displayed the highest level of congruence with the known annotations (Figure~\ref{fig:jurkat}h). Further, congruence between methods has also been depicted in Figure~\ref{fig:jurkatVenn}.

\begin{figure}[!ht]
\centering
\includegraphics[width=0.8\linewidth, keepaspectratio]{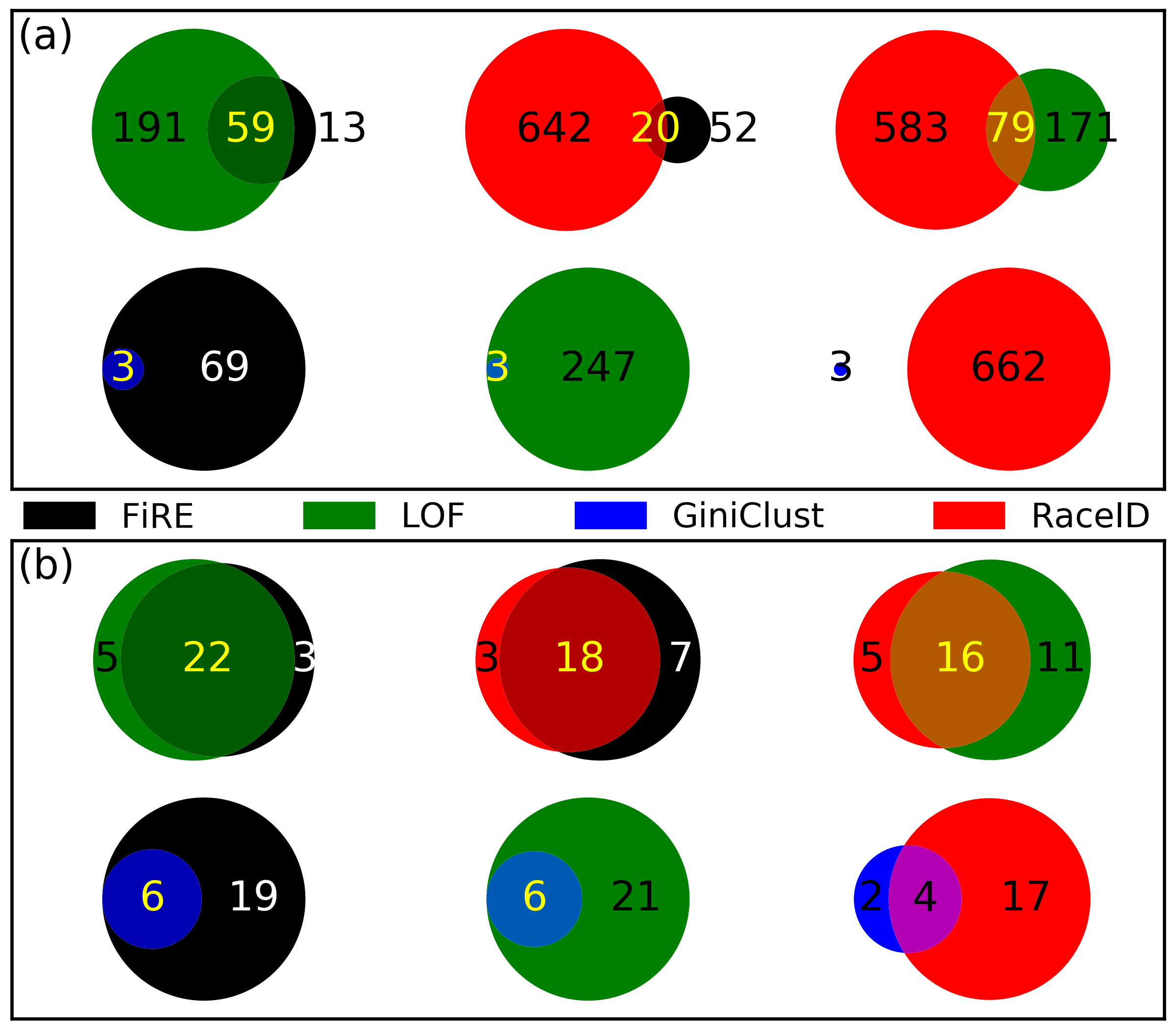}
\caption{Congruence of methods using Venn diagrams. (a),(b) Performance comparison of \glsxtrshort{fire}, GiniClust~\cite{giniClust}, RaceID~\cite{raceId} and \glsxtrshort{lof}~\cite{LOF} as per the rare cells identified by them. (a) Performance comparison of methods above on Embryonic Stem Cells (ESCs) data~\cite{giniESC}. \glsxtrshort{fire} could easily identify the Zscan-4 enriched, 2C-like cell cluster as reported by Jiang \textit{et al.}~\cite{giniClust}. Also, the \glsxtrshort{fire} predicted rare cells had the least overlap with the ones predicted by RaceID, which could not identify those 2C-like cells. (b) Performance on mouse small intestine cells~\cite{raceId}. \glsxtrshort{fire} could identify the rare cell types in the secretory lineage, which consisted of goblet, enteroendocrine, paneth and tuft cells (as discussed in Grun \textit{et al.}~\cite{raceId}).}
\label{fig:suppRaceIDdata}
\end{figure}

To evaluate the performance of the techniques, we used two additional datasets: $\sim$2.5k Embryonic Stem Cells (\glsxtrshort{ESCs})~\cite{giniESC}, and 288 mouse intestinal organoids cell data~\cite{raceId} (Section~\ref{sec::dataFire}). \glsxtrshort{fire} and \glsxtrshort{lof} could identify Zscan-4 enriched, 2C-like cells, as stated by the authors of the GiniClust algorithm~\cite{giniClust} (Figure~\ref{fig:suppRaceIDdata}a). In addition, \glsxtrshort{fire} had the least overlap with RaceID, which could not identify the 2C-like cell type. Figure~\ref{fig:suppRaceIDdata}b depicts the performance of the various methods in identifying rare cell types in the secretory lineage of mouse small intestine, as reported by the authors of the RaceID algorithm. Both \glsxtrshort{fire} and \glsxtrshort{lof} could detect almost all of the designated rare cell types, including the goblet, tuft, paneth, and enteroendocrine cells. GiniClust could detect only a fraction of these cells.

\subsection{FiRE is sensitive to cell type identity}
A simulation study was designed to analyze the robustness and sensitivity of FiRE-score with respect to the number of differentially expressed genes. We first generated an artificial \glsxtrshort{scRNA-seq} data consisting of 500 cells and two cell types. The minor cell type represented about 5\% of the total population (Section~\ref{sec::DEexp}). We kept aside the \glsxtrshort{DE} genes selected through a stringent criterion. For every iteration of the experiments, we replaced a fixed number of non-DE genes by pre-identified \glsxtrshort{DE} genes. We varied the count of differentially expressed genes between 1 to 150 to track the sensitivity of \glsxtrshort{fire} in detecting the minor population.

\begin{figure}[!ht]
\centering
\includegraphics[width=\linewidth]{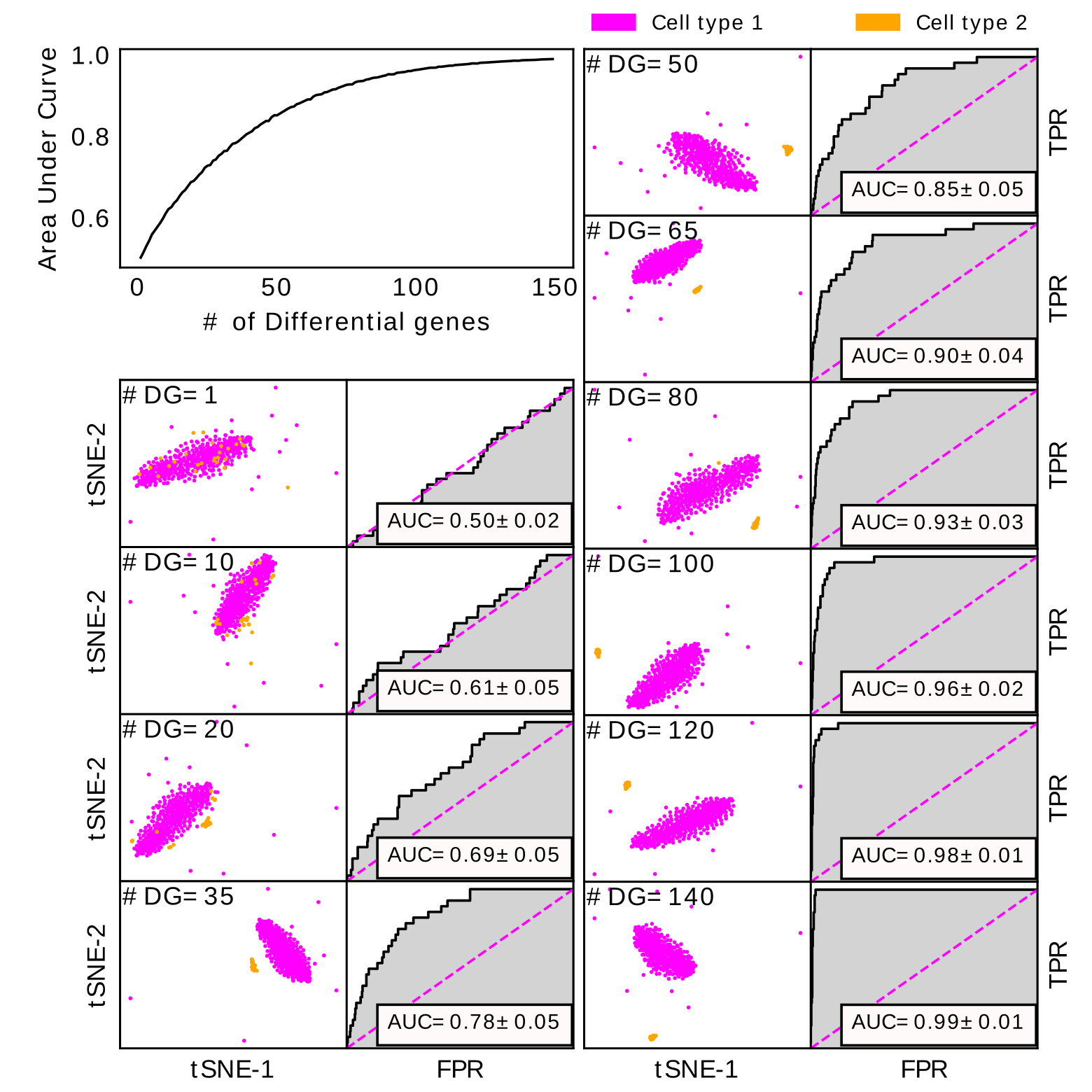}
\caption{Sensitivity of FiRE to cell type. As soon as there are enough differentially expressed genes to create a small cluster that represents the minor cell subpopulation, fire begins properly identifying the minor cell type on \glsxtrshort{scRNA-seq} data generated using the R tool \texttt{splatter}~\cite{splatter}. The succeeding ROC-AUC plots use the figure in the upper-left corner as their legend. Each \glsxtrshort{tsne} and ROC figure pair represents one of the 1000 times the experiment was run with respect to a particular set of differentially expressed genes. Cell-group annotations were used in the ROC-AUC study, and individual cells were given FiRE ratings.}
\label{fig:desimulation}
\end{figure}

With the given set of differentially expressed genes, FiRE scores were obtained and used for computing the ROC-AUC with respect to the minor population. For every count of differentially expressed genes, the aforementioned process was repeated 1000 times to report an average ROC-AUC (Figure~\ref{fig:desimulation}).

With a small number of \glsxtrshort{DE} genes, \glsxtrshort{fire} struggled to detect the minor cell population. However, \glsxtrshort{fire} predictions improved sharply when 20 or more \glsxtrshort{DE} genes were introduced. It reflects the robustness of \glsxtrshort{fire} against noise. A plausible explanation for the same could be that a small number of differential genes fail to stand out in the presence of cell-type specific expression noise (biological plus technical).

\subsection{FiRE is scalable and fast}\label{sec:fire:timecomplexity}
Both RaceID and GiniClust are slow and incur significant memory footprints. For both these methods, clustering takes $O(N^2)$ time, where $N$ is the number of cells. RaceID additionally spends enormous time in fitting parametric distributions for each cell-gene combinations. On the other hand, \glsxtrshort{lof} requires a large number of k-nearest neighbor queries to assign an outlierness score to every cell. \glsxtrshort{fire}, on the other hand, uses Sketching~\cite{rankLsh}, a randomized algorithm for converting expression profiles into bit strings while preserving the weighted $L_1$ distance between data points. The main advantage of randomized algorithms is, that they usually save a lot of computational time. \glsxtrshort{fire} generates a rareness estimation of $N$ cells in linear, i.e., $O(N)$ time, where the constant is $L\times M$.

\begin{figure}[!ht]
\centering
\includegraphics[width=0.6\linewidth]{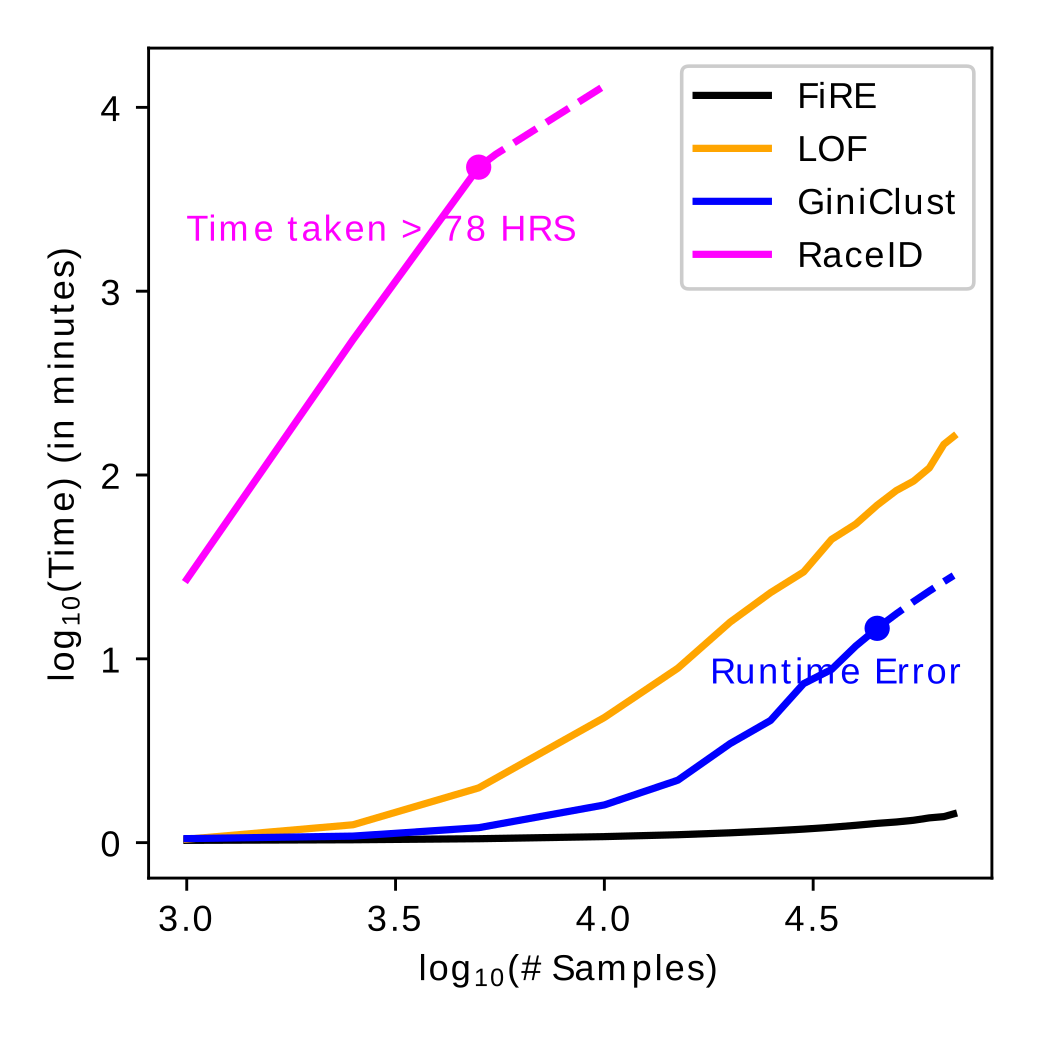}
\caption{\glsxtrshort{fire} is fast. Execution time collected for the four methods with cell counts varying from 1k to $\sim$68k.}
\label{fig:time}
\end{figure}

\glsxtrshort{fire} does two passes over the dataset. To quantify the time complexity of the first pass of \glsxtrshort{fire}, let us assume a dataset of $N$ cells represented in $d$ dimensions. In sketching technique, hash value computation requires two steps 1. thresholding of sub-spaced feature vector and 2. linear combination of thresholded vectors. Assume that $L$ such hash values are generated. Thus, to perform sketching, $L$ random estimators of $M$ dimensions would be generated. Both of the aforementioned steps require $O(M)$ computations for an estimator. Expanding this estimate for each estimator and every sample, the total time required to compute the hash value is $O(N \times L \times M + N \times L \times M) \approx O(N \times L \times M)$. Now, to limit the range of hash value, a modular operation is also performed that can be completed in constant time for every sample. Thus, the total time to hash the dataset is given by $O(N \times L \times M + N) \approx O(N \times L \times M)$. Assuming that total number of possible hash values are $H$. Further assume that these hash values are represented in a form of an array of size $H$ and every element in that array can be accessed in constant time. Interleaving the counting of number of samples sharing same hash value will require additional $O(N \times L)$ time. Now, in the second pass, the final rarity score is computed. This computation of final score will require consulting the hash values of one sample across $L$ estimators that can be completed in $O(L)$ time, and for $N$ samples this time would be $O(N \times L)$. Thus, total time to compute the rarity score (combining both first pass and second pass) can be given by $O(N \times L \times M + N \times L + N \times L) \approx O(N \times L \times M)$. It can be argued that the $L$ and $M$ are the constants for an algorithm run. Thus, the effective complexity of score computation is $O(N)$.

We tracked the time taken by \glsxtrshort{lof}, RaceID, GiniClust, and \glsxtrshort{fire}, while varying the input data size (Figure~\ref{fig:time}, Table~\ref{table:time},~\ref{table:fire:time}) on a single core of a machine with a clock speed of 1.9 GHz, and 1024 GB DDR4-1866/2133 ECC RAM. \glsxtrshort{fire} turned out to be remarkably faster as compared to \glsxtrshort{lof}, RaceID, and GiniClust. For \glsxtrshort{fire}, we recorded $\sim$26 seconds on the $\sim$68k \glsxtrshort{PBMC} data. GiniClust reported a runtime error, when the input expression profiles increased beyond $\sim$45k, while RaceID took $\sim$79 hours for just 5k cells.

\begin{table}[!ht]
\caption{Run-time complexities of algorithms is compared. These complexities are presented with respect to number of samples only.}
\label{table:fire:time}
\centering
\begin{tabular}{| c | c | p{6cm} |}
        \toprule
        Algorithm & Complexity & Remarks \\
        \hline
        FiRE~\cite{fire} & O(N) & \\
        \hline
        LOF~\cite{LOF} & O$(N^{2})$, O$(N\log N)$& Tree based approaches can be used to speed-up the nearest neighbor searches. \\
        \hline
        GiniClust~\cite{giniClust} & O$(N^{2})$ & \\
        \hline
        RaceID~\cite{raceId} & O$(N^{2})$ & Although the complexity is same as GiniClust, it has a very large constant associated with the big O notation, because it also learns multiple parametric distributions. \\
        \bottomrule
        \end{tabular}
\end{table}

\begin{table}[!ht]
\caption{\glsxtrshort{fire} is fast. Execution time (in minutes) collected for the four methods with cell counts varying from 1k to $\sim$68k.}
\label{table:time}
\makebox[1 \textwidth][c]{
\resizebox{1.1 \linewidth}{!}{ %
\begin{tabular}{| c | c | c | c | c |}
\toprule
Samples & FiRE time (min.)~\cite{fire} & LOF time (min.)~\cite{LOF} & GiniClust time (min.)~\cite{giniClust} & RaceID time (min.)~\cite{raceId}\\
\midrule
1000 & 0.0289066871 & 0.0442489147 & 0.0517063022 & 26.14682858 \\
\hline
2500 & 0.0365521709 & 0.250727284 & 0.085117805 & 544.0205631 \\
\hline
5000 &  0.0502097885 & 0.985178566 & 0.2058585723 & 4729.290497 \\
\hline
10000 & 0.0783540368 & 3.79507608 & 0.6036896904 & -\\
\hline
15000 & 0.1040728211 & 7.898171449 & 1.187051054 & -\\
\hline
20000 & 0.1308689475 & 14.83272862 & 2.456504417 & -\\
\hline
25000 & 0.1575588862 & 21.9108673 & 3.621491746 & -\\
\hline
30000 & 0.1842079679 & 28.68896627 & 6.329408483 & -\\
\hline
35000 & 0.2123669028 & 43.65191623 & 7.754190397 & -\\
\hline
40000 & 0.2424690485 & 53.09929506 & 10.74303849 & -\\
\hline
45000 & 0.2728209337 & 67.55955516 & 13.65257459 & -\\
\hline
50000 & 0.2950131019 & 81.28970995 & - & -\\
\hline
55000 & 0.3231681029 & 91.62621465 & - & - \\
\hline
60000 & 0.3635717352 & 108.2481951 & - & -\\
\hline
65000 & 0.3853084048 & 145.867392 & - & -\\
\hline
68579 & 0.4325375557 & 160.8379446 & - & -  \\
\bottomrule
\end{tabular}
}}
\end{table}

\subsection{FiRE resolves heterogeneity among dendritic cells }

\begin{figure}[!ht]
\centering
\includegraphics[width=0.95\linewidth]{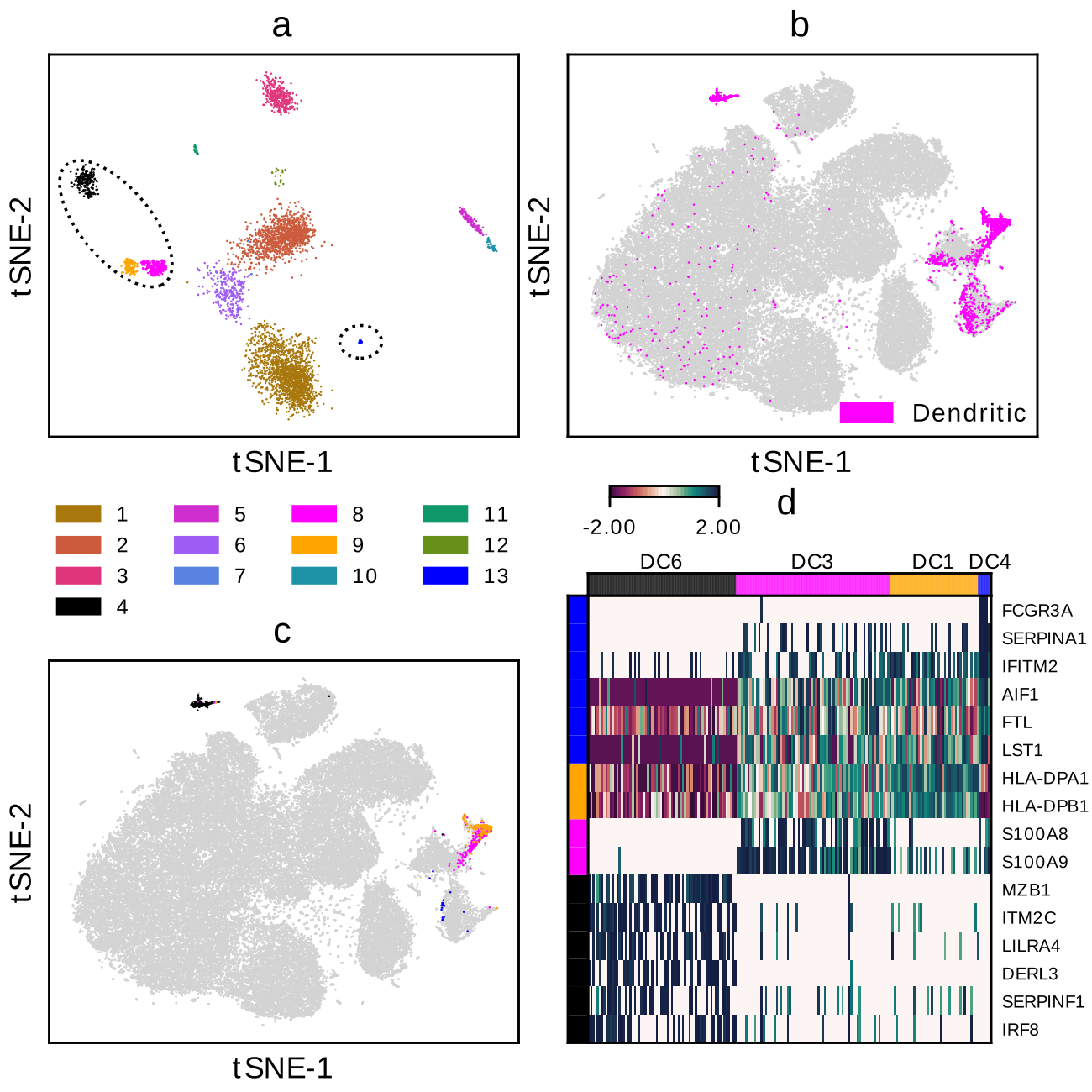}
\caption{\glsxtrshort{fire}-defined dendritic cell heterogeneity in human blood. (a) \glsxtrshort{tsne} based 2D plot of rare cells detected by \glsxtrshort{fire}. Cells are color coded based on their cluster identity as determined by dropClust. (b) Dendritic cells, annotated by the authors, are highlighted in the 2D map adopted from Zheng \textit{et al}~\cite{Zheng}. (c) Dendritic cell sub-types detected from FiRE-reported rare cells are color-coded as per Figure~(a). (d) Characterization of dendritic cell sub-types using markers reported by Villani and colleagues~\cite{Villanieaah4573}.}
\label{fig:pbmcdc}
\end{figure}

Dendritic Cells (\glsxtrshort{DCs}) play a central role in antigen surveillance. \glsxtrshort{DCs} are among the rarest immune cell types, constituting about 0.5\% of the \glsxtrshort{PBMCs}~\cite{fearnley1999monitoring}. A recent study by Villani and colleagues delineated six different sub-types of dendritic cells, by analyzing the expression profiles of a Fluorescence-activated cell sorting (\glsxtrshort{FACS}) sorted population of dendritic cells and monocytes. The several DC sub-types reported by the authors are as follows : CD141$^{+}$ DCs (DC1), CD1C$^{+}\_A$ cDCs (DC2), CD1C$^{+}\_B$ cDCs (DC3), CD1C$^{-}$CD141$^{-}$ (DC4), DC5, and plasmacytoid DCs (DC6, pDCs)~\cite{Villanieaah4573}.

\begin{figure}[!ht]
\centering
\includegraphics[width=\linewidth, keepaspectratio]{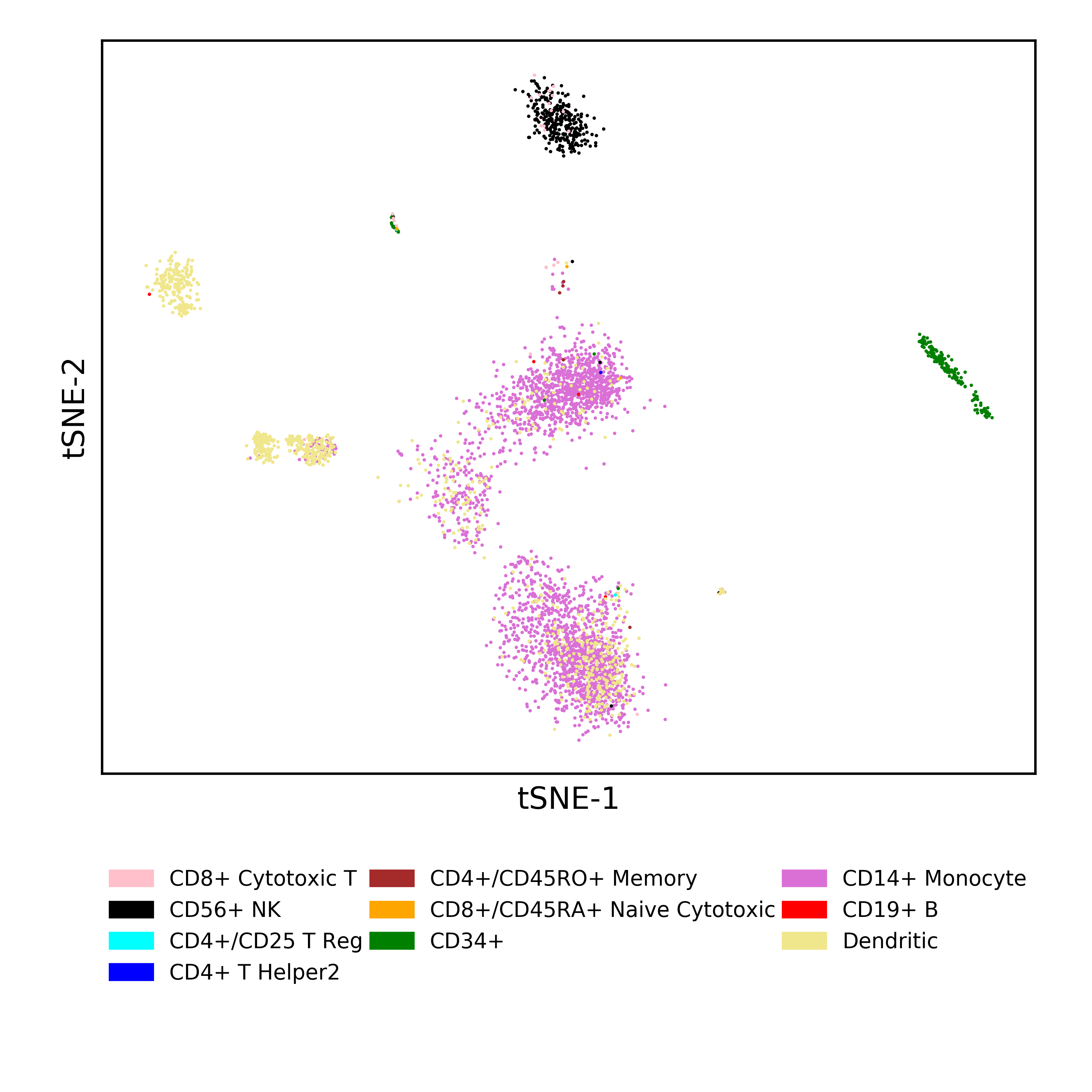}
\caption{2D embedding of rare cells detected by \glsxtrshort{fire} on $\sim$68k \glsxtrshort{PBMCs}. Cells are color coded based on the cell type annotations reported by Zheng~\cite{Zheng}.}
\label{fig:annot}
\end{figure}

We asked if some of the dendritic cell sub-types could be identified in unfractionated \glsxtrshort{PBMC} data. To this end, we applied \glsxtrshort{fire} on $\sim$68k \glsxtrshort{PBMC} data. \glsxtrshort{fire} reported a total of 4238 rare cells, which we then clustered using dropClust~\cite{Sinha170308}. Out of the 13 clearly distinguishable clusters R1-R13, R4, R8, R9, and R13, exclusively consisted of dendritic cells as per the annotations provided by Zheng and colleagues~\cite{Zheng} (Figures~\ref{fig:pbmcdc}a,~\ref{fig:pbmcdc}c, Figure~\ref{fig:annot}). For these 4 \glsxtrshort{DC} clusters, we conducted differential expression analysis to find the cell types specific genes (Section~\ref{sec::DEgenes}). Upon overlaying our differential genes with the ones reported by Villani and colleagues, we could confidently resolve four (DC1, DC3, DC4, DC6) out of the six sub-types reported by Villani \textit{et al.} (Figure~\ref{fig:pbmcdc}d).

To summarize, when applied to the $\sim$68k \glsxtrshort{PBMC} data, \glsxtrshort{fire} helped in delineating four distinct \glsxtrshort{DC} subtypes, of which DC1, DC3, and DC4, were unresolved by the unsupervised clustering used in the original study~\cite{Zheng}. Notably, in the \glsxtrshort{tsne} based 2D embedding, these cell types were visually co-clustered within themselves or the monocytes.

\section{Conclusions}
Of late, single cell transcriptomics has considerably refined our understanding about the true nature of cellular phenotype. It has also accelerated the discovery of new cell types. Most of these new cell types are rare since it's quite improbable for an abundant cell type to remain unobserved for a very long time. A truly rare cell type can only be found by profiling several thousands of cells~\cite{Shapiro2013618}. While technological advances over the past years have enabled us to perform ultra high-throughput single cell experiments, scalable methods for rare cell detection are nearly non-existent. \glsxtrshort{fire} attempts to fill that gap, with a number of pragmatic design considerations. Most notable among these is its ability to avoid clustering as an intermediate step. A typical clustering technique is not only time consuming, but also incapable of comprehensively charting the minor cell types in a complex tissue on a single go~\cite{campbell2017molecular}.

While RaceID~\cite{raceId} and GiniClust~\cite{giniClust} offer binary predictions, \glsxtrshort{fire} gives a rareness score to every individual expression profile. We demonstrated how these scores might help the user focus their downstream analyses on a small fraction of the input \glsxtrshort{scRNA-seq} profiles. A score is particularly helpful since a number complex techniques such as pseudo-temporal analysis~\cite{trapnell2014dynamics}, shared the nearest neighborhood based topological clustering~\cite{xu2015identification} etc. are applicable only on a few hundreds of cells.

\glsxtrshort{fire} makes multiple estimations of the proximity between a pair of cells, in low dimensional spaces, as determined by the parameter $M$. The notion of similarity for \glsxtrshort{lof}~\cite{LOF}, on the other hand, is confounded by the arbitrary scales of the input dimensions. As a result, even though \glsxtrshort{lof} consistently outperforms RaceID and GiniClust, it struggles to match the performance of \glsxtrshort{fire}.

\glsxtrshort{fire}, in principle, does not discriminate between an outlier and cells representing minor cell-types. We adhered to dropClust~\cite{Sinha170308} for clustering of rare cells detected by \glsxtrshort{fire}. dropClust~\cite{Sinha170308} does not administer any special treatment to outliers. As a result, outlier cells, if any, get submerged into the minor cell clusters. However, one may wish to use hierarchical or density based clustering techniques to flag outlier cells. A new version of FiRE will be compared against other state-of-the-art techniques such as scanpy~\cite{wolf2018scanpy}, Seurat~\cite{satija2015spatial}, scAIDE~\cite{scaide}, MicroCellClust~\cite{microclust}, etc.

\glsxtrshort{fire} took $\sim26$ seconds to analyze a \glsxtrshort{scRNA-seq} dataset containing $\sim68$k expression profiles. Such unrivaled speed, combined with the ability to pinpoint the truly rare expression profiles, makes the algorithm future proof.
 
Outliers may also signal anomalous samples, or samples pointing to anomalies. Identifying anomalies requires identifying both local and global outliers. In the next chapter, we discuss an extension of \glsxtrshort{fire} and its application to anomaly detection.

\chapter{Linear Time Identification of Local and Global Outliers\protect\footnote{The work presented in this chapter has been published as a research paper titled ``\textit{Linear time identification of local and global outliers}'' in Neurocomputing (2021).}}\label{chapter::3}

\section{Introduction}

Anomaly detection refers to the problem of finding patterns in data that do not conform to expected behavior~\cite{chandola2009anomaly, knorr1997unified}. Anomaly detection has innumerable practical applications, including credit card fraud detection~\cite{creditcardfraud}, fault detection~\cite{faultdetection}, rare cell type detection in large scale gene expression data, and system health monitoring~\cite{systemhealthmonitoring}. Outlier detection methods may be supervised~\cite{supervised}, semi-supervised~\cite{semisupervised, semisupervised1}, or unsupervised~\cite{campos2016evaluation, unsupervised}. Unsupervised methods are the most popular, because of the challenges in gathering annotated anomaly examples. Outliers may be treated as global, that are eccentric and irregular relative to the entire set of data points; or local, that are unusual only to a local section of the data~\cite{campello2015hierarchical}.

To date, several studies have reviewed anomaly detection algorithms~\cite{zimek2012survey,hodge2004survey}. However, little has been reported in the direction of comprehensive performance evaluation of widely used methods~\cite{campos2016evaluation,goldstein2016comparative} on several annotated datasets. The most comprehensive work in this context is by Campos and colleagues~\cite{campos2016evaluation}, who reported a comparison of 12 well known methods~\cite{knn, knnw, odin, LOF, simplifiedlof, cof, inflo, loop, ldof, ldf, kdeos, fastabod} on about 1000 datasets.

The existence of local and global outliers is well recognized, but there has been little visible effort to assess the relative performance of existing methods for their ability to detect outlier sub-categories. We propose a scoring criterion that assigns a value to a dataset based on it's outliers' local or global nature. This measure is used to classify datasets based on their global and local outlier compositions. The analysis reveals varied performance with different algorithms.

In this study, we incorporate the Finder of Rare Entities (\glsxtrshort{fire}) (discussed in the previous chapter). \glsxtrshort{fire} was proposed as a linear time hashing-based algorithm first envisaged for identifying rare cells. In molecular biology, rare cells are analogous to global outliers, whereas sequencing artifacts and biological noise dominate the emergence of local outliers. In this study, we propose an extension of \glsxtrshort{fire}, referred to as \textit{FiRE.1} in the sequel, that also takes into account local outliers.

Subspace outlier degree (\glsxtrshort{sod})~\cite{sod} was proposed to tackle the problem of outlier detection in high dimensional data. We also considered the histogram-based outlier score (\glsxtrshort{hbos})~\cite{hbos}, which is a linear time outlier detection algorithm. Two more algorithms based on $k$ nearest neighbors namely local isolation coefficient (\glsxtrshort{lic})~\cite{lic} and distance-based outlier score (\glsxtrshort{dbos})~\cite{dboutlierscore} are also included for comparisons. These 6 algorithms, including \glsxtrshort{fire} and FiRE.1, are compared with 12 algorithms evaluated by Campos~\cite{campos2016evaluation} for comparing different algorithms.

\section{FiRE.1: FiRE for local outliers}

\begin{figure}
    \centering
    \includegraphics[width=\textwidth, keepaspectratio]{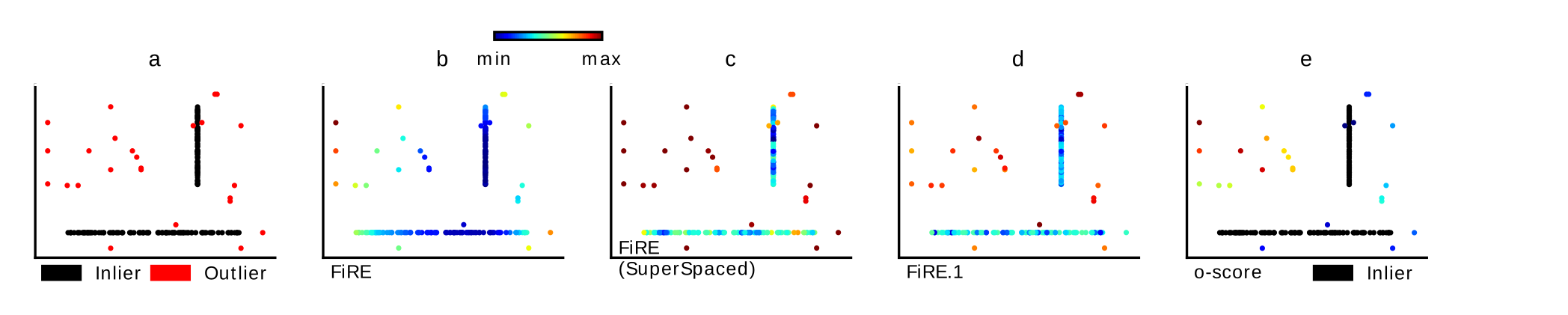}
    \caption{Performance of \glsxtrshort{fire} and FiRE.1 on a simulated dataset with local and global outliers (the illustration is inspired by~\cite{sod}). (a) A 2-dimensional simulated dataset containing both local and global outliers. (b) Distribution of \textit{FiRE-score} when $M=2$. Global outliers are well captured and have high values of \textit{FiRE-score}. On the other hand, local outliers deviating marginally from a local population are not captured. (c) Distribution of \textit{FiRE-score} when $M > d$ ($M=100$). In addition to global outliers, local ones are also identified since sufficient hash indexes account for minor differences between local outliers and their local population. (d) FiRE.1 identifies both local and global outliers. (e) Variation of outlierness criterion \textit{o-score} on different types of outliers. For global outliers, the values are high and vice-versa. }
    \label{fig:fireSuperspaced}
\end{figure}

In \glsxtrshort{fire}, a threshold divides the given dimension into two parts. Thus, the resultant bit vector of length $M$ is composed of 0s and 1s. As a result, the maximum possible unique hash indices for a given value of $M$ are $2^M$ in number. For low-dimensional data, if $M\leq d$, we will end up with a small number of hash indexes. Hence, small variations in data are not adequately captured, as shown in Figure~\ref{fig:fireSuperspaced}. Figure~\ref{fig:fireSuperspaced}(a) contains both local and global outliers. For $M=2$, maximum possible hash indexes is $4$. As shown in Figure~\ref{fig:fireSuperspaced}(b), high values of \textit{FiRE-score} are assigned to global outliers. However, local outliers are missed. In contrast, when $M>d$ in Figure~\ref{fig:fireSuperspaced}(c), local outliers are also captured.

\begin{figure}
    \centering
    \includegraphics[width=\textwidth, keepaspectratio]{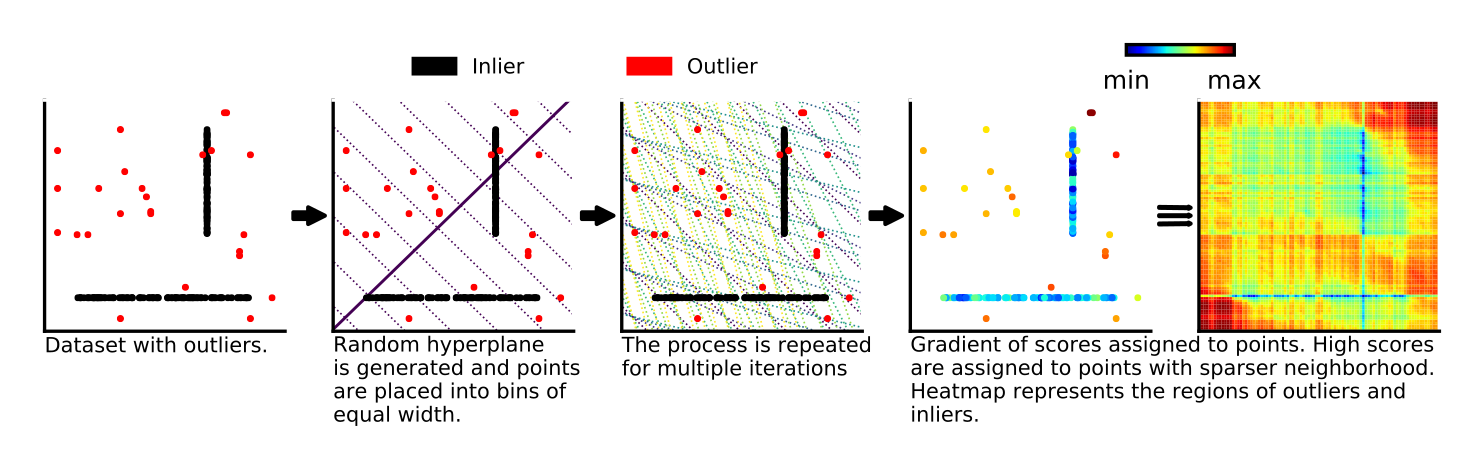}
    \caption{An overview of FiRE.1 on a simulated dataset. The heatmap depicts FiRE.1 approximated regional densities.}
    \label{fig:overview}
\end{figure}

Thus, one limitation of \glsxtrshort{fire} is it's dependence on $M$, to decide the number of bins in which data points need to be segregated. As a result, very few bins are created for low-dimensional data, leading to insufficient granularity in the bins needed to identify local outliers. Hence, FiRE's performance falls on datasets with local outliers. To overcome this, we propose FiRE.1, that replaces the sketching process with a projection hash. The hash index $h^{(l)}_{i}$ for sample $i$ is computed as follows.

\begin{align}\label{proj}
h^{(l)}_{i} = \lfloor\frac{1}{\text{\textit{bin-width}}}\sum_{j=1}^{j=M}(P_{l})^{(i)}_{j}*t^{(l)}_{j} + bias_{l}\rfloor
\end{align}
In \eqref{proj}, $bias_{l} \sim U[-\textit{bin-width},\textit{bin-width}]$~\cite{optimallsh}. Figure~\ref{fig:overview} represents an overview of FiRE.1 steps. For a given value of $M$, sub-spaced points are projected onto a line by using a dot product with a randomly chosen vector $t^{(l)}$. The projected values are divided into equally spaced bins, determined by \textit{bin-width}, that controls the bin size. The total number of bins in FiRE.1 is notionally infinite. The smaller value of \textit{bin-width} allows the points to get distributed across a large number of bins. Consequently, a smaller bin-width is preferable for datasets with more local outliers. In contrast to \glsxtrshort{fire}, FiRE.1 provides explicit control over bin width (Figure~\ref{fig:fireSuperspaced}(d)).

After dividing points into bins, \eqref{eq::neighborhood} is used to compute the neighborhood of every point probabilistically. Similar to \glsxtrshort{fire}, \eqref{eq::score} is used to compute the rareness score of every point. Points with high \textit{FiRE-score}/\textit{FiRE.1-score} lie in sparser neighborhoods and are more likely to be outliers. Algorithm~\ref{algo::firescore} describes the working of FiRE.1.

The overall space requirement of FiRE.1 is $L \times (2M + N)$. FiRE.1 needs $L \times M$ space to store the weight matrix and $L \times M$ space to store the sub-spacing matrix. It needs $L \times N$ space to store hash indices of every point.

\begin{algorithm}
    \caption{FiRE.1}\label{algo::firescore}
    \begin{algorithmic}
        \State \textbf{Input:} Data points ($x^{(1)}, x^{(2)}, ..., x^{(N)}$), $x^{(i)} \in S$, $S \subseteq \mathbb{R}^d$
        \State \hspace{10mm} $L \gets \text{Number of projectors}$
        \State \hspace{10mm} $M \gets \text{Size of subspace}$
        \State \hspace{10mm} $\textit{bin-width} \gets \text{Width of each bin}$
        \State \textbf{Initialize:} For every projector $l \in \{1,...,L\}$
        \State \hspace{20mm} $h^{(l)} \gets$ $N$ size array for hash indexes.
        \State \hspace{20mm} $m^{(l)} \gets$ $M$ size array for selected dimensions.
        \State \hspace{20mm} $t^{(l)} \gets$ $M$ size array for weights.
        \State \hspace{20mm} $bias_{l} \gets$ Scalar for generated bias.
        \State \hspace{15mm} $mi_{j} \gets \min_{i\in\{1,...,N\}}S^{(i)}_{j}$   $\forall j\in\{1,...,d\}$
        \State \hspace{15mm} $ma_{j} \gets \max_{i\in\{1,...,N\}}S^{(i)}_{j}$   $\forall j\in\{1,...,d\}$
        \State \textbf{Run:}
        \State \textbf{for} $l \in \{1,...,L\}$
        \State \hspace{10mm} $m^{(l)} \gets$ Randomly sampled $M$ dimensions from $d$ with replacement.
        \State \hspace{10mm} $P_{l} \gets$ Projected data from $S$ onto $m^{(l)}$.
        \State \hspace{10mm} $t^{(l)}_{j} \sim U[mi_{m^{(l)}_{j}}, ma_{m^{(l)}_{j}}]$  $\forall j \in \{1,...,M\}$
        \State \hspace{10mm} $bias_{l} \sim U[-\textit{bin-width}, \textit{bin-width}]$
        \State \hspace{10mm} $h^{(l)}_{i} = \lfloor\frac{1}{\text{\textit{bin-width}}}\sum_{j=1}^{j=M}(P_{l})^{(i)}_{j}*t^{(l)}_j + bias_{l}\rfloor$
        \State \hspace{10mm} $\textit{N}^{(l)}_i = \frac{\text{Total data points with hash index }h^{(l)}_{i} }{N}$  $\forall i \in \{1,...,N\}$
        \State $\textit{score}_i = -2\sum_{l=1}^{L}\log \textit{N}^{(l)}_i$   $\forall i \in \{1,...,N\}$
    \end{algorithmic}
\end{algorithm}

\section{Outlier score on unseen data}
The \glsxtrshort{fire} family of algorithms assigns scores to every data sample or point, by dividing the whole space into bins. \glsxtrshort{fire} algorithms approximate the data distribution based on each bin's population density. This information can be used to assign scores to unseen points as well.

Let $S^{'}$ contains $N^{'}$ points drawn independently and identically from a distribution $D$ over $X$. Points $x_i \in S^{'}\:\:\forall i \in \{1,...,N^{'}\}$ are then projected onto $m^{(l)}\:\:\forall l \in \{1,...,L\}$. Say, $P_l$ denotes this projected set. Then, \eqref{proj} is used to place all unseen data points into bins, say $hu^{(l)}_i$ for point $i$. Probabilistic neighborhood $Nu^{(l)}_i$ in this scenario is defined as
\begin{align}\label{eq::unseenNeighborhood}
Nu^{(l)}_i = \frac{1 + \text{Total data points with hash index }hu^{(l)}_{i}\text{ in }h^{(l)}}{1+N}
\end{align}

In \eqref{eq::unseenNeighborhood} $N$ denotes the total number of training instances. When a point falls into an empty bin, it's probabilistic neighborhood is $1/(N+1)$. FiRE.1 uses \eqref{eq::score} to assign scores based on these neighborhood values to every point. Algorithm~\ref{algo::unseendata} describes FiRE.1's procedure for assigning scores to unseen samples. The heatmap in figure~\ref{fig:overview} shows a heatmap depicting scores for regions that have no training samples.

\begin{algorithm}
    \caption{Outlier Score on unseen data.} \label{algo::unseendata}
    \begin{algorithmic}
        \State \textbf{Input:} Unseen data points ($x^{(1)}, x^{(2)}, ..., x^{(N^{'})}$), $x^{(i)}\in S^{'}, S^{'}\subseteq\mathbb{R}^d$
        \State \hspace{10mm} $L \gets \text{Number of projectors}$
        \State \hspace{10mm} $N \gets \text{Total number of training instances}$
        \State \hspace{10mm} $\textit{bin-width} \gets \text{Width of each bin}$
        \State \hspace{10mm} $h \gets$ $L\times N$ size array for seen hash indexes.
        \State \hspace{10mm} $m \gets$ $L\times M$ size array for dimensions.
        \State \hspace{10mm} $t \gets$ $L \times M$ size array for weights.
        \State \hspace{10mm} $bias \gets$ $L$ size array for bias.
        \State \textbf{Run:}
        \State \textbf{for} $l \in \{1,...,L\}$
        \State \hspace{10mm} $P_l \gets$ Project data from $S^{'}$ onto $m^{(l)}$.
        \State \hspace{10mm} $hu^{(l)}_{i} = \lfloor\frac{1}{\text{\textit{bin-width}}}\sum_{j=1}^{j=M}(P_l)^{(i)}_j*t^{(l)}_j + bias_l\rfloor$
        \State \hspace{10mm} $\textit{Nu}^{(l)}_{i}=\frac{1 + \text{Total data points with hash index }hu^{(l)}_{i} \text{ in } h^{(l)} }{1+N}$  $\forall i \in \{1,...,N^{'}\}$
        \State $\textit{score}_i = -2\sum_{l=1}^{L}\log \textit{Nu}^{(l)}_i$   $\forall i \in \{1,...,N^{'}\}$
    \end{algorithmic}
\end{algorithm}

\subsection{Local vs Global outliers}\label{outliernessMetric}

We propose an "outlierness" criterion \textit{o-score} to determine the type of outliers present in a dataset. Let $S$ denote a given set of data points, where $O$ denotes points marked as outliers and $I$ denotes points marked as inliers. For every $o \in O$, \textit{o-score} is defined as.
\begin{align}
\textit{o-score}(o) = \frac{\min_{i \in I}(\textit{dist}(o,i)) }{\max_{i \in I}(\textit{dist}(o,i))}
\end{align}
For global outliers, both minimum and maximum distances to inliers are comparable. However, for local outliers, the minimum value is small, because the context is local. Hence, global outliers have higher values of \textit{o-score} when compared with local outliers, as shown in Figure~\ref{fig:fireSuperspaced}(e). The existence of noise may lead to a spurious minimum and maximum distance values. Therefore, we average $\phi$ smallest (largest) distances between $o$ and inliers for the minimum (maximum) distance.

\section{Run time complexity}\label{sec::complexity}
Methods such as \glsxtrshort{knn}~\cite{knn}, \glsxtrshort{odin}~\cite{odin}, \glsxtrshort{lof}~\cite{LOF}, Simplified-LOF~\cite{simplifiedlof}, \glsxtrshort{inflo}~\cite{inflo}, \glsxtrshort{loop}~\cite{loop}, \glsxtrshort{ldf}~\cite{ldf}, \glsxtrshort{lic}~\cite{lic}, \glsxtrshort{dbos}~\cite{dboutlierscore}, and \glsxtrshort{kdeos}~\cite{kdeos} require $O(N^2)$ running time to identify $k$ nearest neighbors. This may be further reduced to $O(N\log N)$ if the dataset is indexed. \glsxtrshort{knnw}~\cite{knnw} has an overall complexity of $O(N\log N)$. On the other hand, \glsxtrshort{cof}~\cite{cof}, \glsxtrshort{ldof}~\cite{ldof}, and \glsxtrshort{fastabod}~\cite{fastabod} require an additional $O(k^2)$ computations per point and therefore incur an overall cost in $O(N\log N + N\times k^2)$. \glsxtrshort{sod} has an overall complexity of $O(d\times N^2)$. This can also be reduced to $O(d\times N\log N)$ if an index structure is used to find NNs. \glsxtrshort{hbos}~\cite{hbos} has $O(N)$ complexity for fixed bin width and $O(N\log N)$ for dynamic bin widths. However, PyOD~\cite{zhao2019pyod} supports the implementation of \glsxtrshort{hbos} with fixed bin width and hence, it has $O(N)$ time complexity.

\begin{table}[th!]
\caption{Run-time complexities of algorithms is compared. These complexities are presented with respect to the number of samples only.}
\label{table:fire1:time}
\centering
\begin{tabular}{| c | c | p{5.5cm} |}
        \toprule
        Algorithm & Complexity & Remarks \\
        \hline
        & & \\
        FiRE~\cite{fire} & \multirow{2}{*}{O(N)} & \\
        FiRE.1~\cite{cattral2002evolutionary} & & \\
        & & \\
        \hline
        & & \\
        kNN~\cite{knn} & \multirow{10}{*}{O$(N^{2})$, O$(N\log N)$} & \multirow{10}{5.5cm}{Tree/index based approaches can be used to speed-up the nearest neighbor searches.} \\
        ODIN~\cite{odin} & & \\
        LOF~\cite{LOF} & & \\
        Simplified-LOF~\cite{simplifiedlof} & & \\
        INFLO~\cite{inflo} & & \\
        LoOP~\cite{loop} & & \\
        LDF~\cite{ldf} & & \\
        LIC~\cite{lic} & & \\
        DB-outlier score~\cite{dboutlierscore} & & \\
        KDEOS~\cite{kdeos} & & \\
        & & \\
        \hline
        & & \\
        kNNW~\cite{knnw} & O$(N \log N)$ & \\
        & & \\
        \hline
        & & \\
        COF~\cite{cof} & \multirow{3}{*}{O$(N \log N + N \times k^{2})$} & \multirow{3}{5.5cm}{For every sample an extra computation of O$(k^{2})$ is performed, where $k$ is the number of nearest neighbors needs to be evaluated.} \\
        LDOF~\cite{ldof} & & \\
        Fast-ABOD~\cite{ldof} & & \\
        & & \\
        & & \\
        \hline
        & & \\
        SOD~\cite{sod} & O$(d \times N^{2})$ & \multirow{3}{5.5cm}{Here $d$ is data dimension used for subspace creation. Reduction in time can be achieved by indexing.} \\
        & O$(d \times N \log N)$ & \\
        & & \\
        & & \\
        \hline
        & & \\
        \multirow{4}{*}{HBOS~\cite{hbos}} & O$(N)$ & With fixed bin width (PyOD~\cite{zhao2019pyod} implements this version)\\
        & & \\
        \cline{2-3}
        & & \\
        & O$(N \log N)$ & With dynamic bin width. \\
        & & \\
        \bottomrule
        \end{tabular}
\end{table}

\glsxtrshort{fire} and FiRE.1 require only $O(N)$ running time. Both \glsxtrshort{fire} and FiRE.1 perform only two passes over the entire dataset. In the first pass, all data points are hashed individually to their respective bins in $O(N)$ operations. The overall time for initializing a weight matrix for hashing is determined by $L\times M$. The hashing of a point to a bin for a given projector involves dot product with the weight vector, and subsequent quantization into one of the bins. The overall time for hashing a point is $L \times M$. Thus, the overall complexity of the first pass is $O(CN)$, where the constant $C$ is $L \times M$. In the second pass, the neighborhood information is extracted for each entity depending on the accumulation of data points in bins. Therefore, this step is also linear in the sample size. Thus, the overall complexities of \glsxtrshort{fire} and FiRE.1 are $O(N)$. Table~\ref{table:fire1:time} summarizes the run-time complexities of different algorithms.

\section{Experimental setup}
\subsection{Description of datasets}
The effectiveness of outlier detection approaches needs to be evaluated across benchmark datasets with known nature of outlierness. Campos and colleagues~\cite{campos2016evaluation} created a publicly available database of approximately 1000 datasets (including all variants), derived from 23 datasets from UCI repository~\cite{Dua:2019}. For every dataset, two different versions: unnormalized, and attribute-wise normalization, have been considered. Further, for a dataset consisting of duplicates, the corresponding version without duplicates is also generated. Also, different sub-sampled versions are generated from a given dataset, with different percentages of outliers. They argue that random subsampling may have a different impact on the resultant outliers, and hence, 10 different versions were generated. The database consists of diverse datasets ranging from 5 - 1,555, dataset size varying from 80 to $\sim$60,000, and outlier percentage from 2\% to $\sim$40\%.

\subsection{Metrics for comparing outlier detection methods}
The literature is replete with several well-known performance metrics: $\textit{Precision at }n$ (\glsxtrshort{pan}) measures how many points are marked as top $n$ outliers are true outliers. The highest value is 1 when all top $n$ points are true outliers. \glsxtrshort{pan} is sensitive to the value of $n$. For all experiments, total outliers in a dataset are used as the value of $n$.
\begin{align}
\textit{P@}n = \frac{|\{o \in O \:|\: \textit{rank}(o) \leq n\}|}{n}
\end{align}
It must be noted that for datasets with few outliers, \glsxtrshort{pan} may be low. Hence, when methods are compared across datasets with different proportions of outliers, \glsxtrshort{pan} is adjusted for chance and is referred to as $\textit{Adjusted P@}n$~\cite{campos2016evaluation}.
\begin{align}
\textit{Adjusted P@}n = \frac{\textit{P@}n - |O|/N}{1 - |O|/N}
\end{align}
\textit{Average Precision} (\glsxtrshort{AP}) captures the values of \textit{Precision} with increasing values of \textit{Recall}. In other words, the average of values of \textit{Precision} when \textit{Recall} varies between $1/|O|$ and 1. \glsxtrshort{AP} is equivalent to area under the \textit{Precision-Recall} curve.

\begin{align}
\textit{AP} = \frac{1}{|O|}\sum_{o \in O} \textit{P@rank}(o)
\end{align}
Similarly \glsxtrshort{AP} must be adjusted for chance when methods are compared across datasets with differing outlier proportions~\cite{campos2016evaluation}.
\begin{align}
\textit{Adjusted AP} = \frac{\textit{AP} - |O|/N}{1 - |O|/N}
\end{align}
\textit{Receiver Operator Characteristics Area Under the Curve} (\glsxtrshort{rocauc}) measures how well a method distinguishes between inliers and outliers at different thresholds. The higher the value of \glsxtrshort{auc}, the better the method's performance. \glsxtrshort{rocauc} is insensitive to outlier percentage.
\begin{align}
\textit{ROC AUC} = \textit{mean}_{o \in O, i \in I}
\begin{cases}
1 \text{ if } \textit{score}(o) > \textit{score}(i)\\
1/2 \text{ if } \textit{score}(o) = \textit{score}(i)\\
0 \text{ if } \textit{score}(o) < \textit{score}(i)
\end{cases}
\end{align}
$\textit{Adjusted P@}n$, \textit{Adjusted AP}, and \glsxtrshort{rocauc} are the most suitable metrics for comparing methods across datasets with different outlier proportions. For skewed datasets, \glsxtrshort{pan}, $\textit{Adjusted P@}n$, \glsxtrshort{AP}, or \textit{Adjusted AP} are more suitable than \glsxtrshort{rocauc}~\cite{aucvspr}. \glsxtrshort{AP} approximates the area under the \textit{Precision-Recall} curve, while \glsxtrshort{rocauc} measures the area under the \textit{True Positive Rate} (\glsxtrshort{tpr})-\textit{False Positive Rate} (\glsxtrshort{fpr}) curve. \textit{Recall} (\ref{recall}) is same as \glsxtrshort{tpr} (\ref{tpr}). If the size of the negative class significantly exceeds the positive one, then even large changes in false positives (\glsxtrshort{fp}) are not captured by \textit{FPR,} which is used in \glsxtrshort{roc}. This is because \textit{FPR,} compares false positives with true negatives (\glsxtrshort{tn}) (\ref{fpr}). On the other side, \textit{Precision} compares false positives with true positives (\ref{precision}). Since true positives are comparable to false positives, changes in false positives are easily captured by \textit{Precision}.
\begin{align}\label{tpr}
\textit{True Positive Rate }(\textit{TPR}) = \frac{TP}{TP+FN}
\end{align}
\begin{align}\label{fpr}
\textit{False Positive Rate } (\textit{FPR}) = \frac{FP}{FP+TN}
\end{align}
\begin{align}\label{precision}
\textit{Precision} = \frac{TP}{FP+TP}
\end{align}
\begin{align}\label{recall}
\textit{Recall} = \frac{TP}{TP+FN}
\end{align}

\glsxtrshort{pan} also differs from \glsxtrshort{AP}. As \textit{Recall} varies, \textit{Precision} does not change linearly. Hence, if a high \textit{Recall} is desired for an outlier class, then \textit{Precision} may drop. In contrast, \glsxtrshort{pan} finds \textit{Precision} only for a given threshold $n$. A high value of \glsxtrshort{pan} indicates, that at least some of the highest scores are assigned to true outliers. In other words, every evaluation measure serves a different purpose, and hence, a comparison between methods using a single metric may not always be appropriate~\cite{aucvspr}.

\section{Performance comparison of methods on a repository of almost 1000 datasets}

\subsection{Tuning of hyperparameters}\label{tuning_params}
Eighteen methods were compared in our study. Campos and colleagues~\cite{campos2016evaluation} reported the performance of 12 methods using the ELKI framework~\cite{DBLP:journals/pvldb/SchubertKEZSZ15}. The methods used in this study are \glsxtrshort{knn}, \glsxtrshort{knnw}, \glsxtrshort{lof}, Simplified-LOF, \glsxtrshort{kdeos}, \glsxtrshort{cof}, \glsxtrshort{ldof}, \glsxtrshort{inflo}, \glsxtrshort{loop}, \glsxtrshort{ldf}, \glsxtrshort{fastabod}, and \glsxtrshort{odin}. For all these methods, the value of $k$ was tuned in the range 1 to 100. \glsxtrshort{fastabod}, \glsxtrshort{ldf}, and \glsxtrshort{kdeos} have additional hyperparameters. For \glsxtrshort{fastabod}, a polynomial kernel of degree 2 was used. A constant value of 1 was used for kernel bandwidth multiplier $h$ for both \glsxtrshort{kdeos} and \glsxtrshort{ldf}. In addition to this, \glsxtrshort{ldf} also needs a constant $c$ ($c=0.1$). These previously reported results were compared with additional methods. Predictions of \glsxtrshort{dbos}, \glsxtrshort{lic}, and \glsxtrshort{sod} were also computed using the ELKI framework.\glsxtrshort{lic} was tuned by trying different values of $k$ ranging between 1 and 100. However, for \glsxtrshort{sod}, $k$ was varied between 2 and 100 (so that there are at least 2 NNs to choose from to define the reference set). In addition to this, for \glsxtrshort{sod}, we adhered to $\alpha=0.8$, as suggested in Kriegel \textit{et al.}~\cite{sod}. For the \glsxtrshort{dbos}, the $distance$ was varied between 0.1 and 0.8, with a step size of 0.01. For \glsxtrshort{hbos}, the PyOD toolkit~\cite{zhao2019pyod} was used. PyOD implementation supports the construction of histograms with fixed bins. For \glsxtrshort{hbos}, the numbers of bins varied from 1 to 100. For FiRE.1, the value of $L$ was fixed at 100, while in \glsxtrshort{fire}, the value of $L$ was selected from \{50, 100, ...,500\}. For both \glsxtrshort{fire} and FiRE.1 the value of $M$ was selected from $\{\lceil\log d\rceil,\lceil\sqrt{d}\rceil, \lceil\frac{d}{2}\rceil, d, \lceil1.5 \times d\rceil\}$. In addition to this, FiRE.1 also needs to be tuned for \textit{bin-width}. The value of \textit{bin-width} was tuned over \{10, 9, 8, 7, 6, 5, 4, 3, 2, 1, 0.1, $1\mathrm{e}{-2}$, $1\mathrm{e}{-3}$, $1\mathrm{e}{-4}$, $1\mathrm{e}{-5}$, $1\mathrm{e}{-6}$\}.

\subsection{Comparison of methods using Friedman ranking}
\begin{figure}
    \centering
    \includegraphics[width=\linewidth, keepaspectratio]{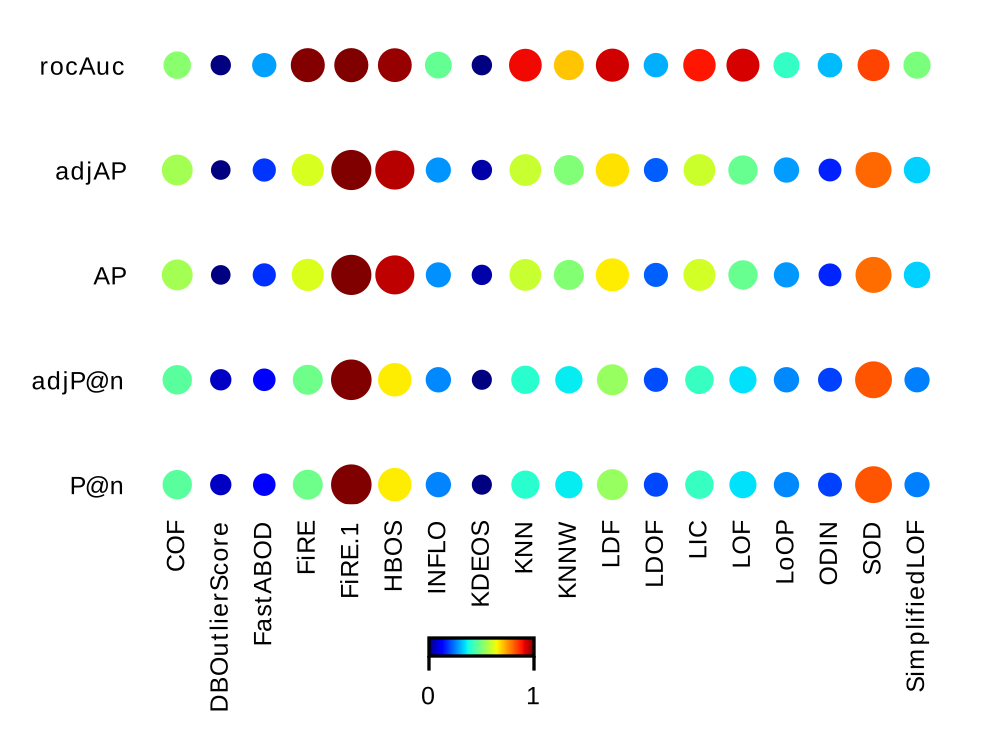}
    \caption{Performance comparison of 18 outlier detection methods on $\sim$1000 datasets. The performance of all methods was evaluated on 5 evaluation metrics. Friedman ranking determined for every method for each metric. The method with the smallest Friedman rank performs the best on a given measure. The color intensity in a heatmap depicts the inverse of the Friedman ranking.}
    \label{fig:comparison}
\end{figure}

Friedman ranking is a non-parametric statistical technique that measures the performance of a set of methods on multiple datasets. Friedman ranking was computed for methods on all evaluation measures. For a given evaluation measure, methods were ranked from 1 to $\Phi$ for a given dataset, where $\Phi$ is the total number of methods that could scale for the given dataset. The increasing values of ranks were assigned to methods with decreasing values of measure. Thus, the method with the highest value of the measure (the best performance) has the lowest rank. The ties between methods with the same values of a certain measure were broken by replacing them with an average value of successive ranks. For a given evaluation measure, we reported the average rank across all datasets (Figure~\ref{fig:comparison}). Among 18 methods, FiRE.1, \glsxtrshort{fire}, \glsxtrshort{hbos}, and \glsxtrshort{sod} have outstanding performances when compared with others. However, \glsxtrshort{sod} could not scale for all datasets due to higher running time complexity. \glsxtrshort{sod} also requires $O(N^2)$ memory for every dimension. For every sample $p$, it calculates distances between points in $R(p)$ and it's mean. FiRE.1 consistently yielded the best performance on all considered evaluation criteria. FiRE.1 exhibited a significant margin w.r.t. the remaining methods in terms of the calculated \glsxtrshort{pan} and \textit{Adjusted} $\textit{P@}n$, which indicates that FiRE.1 reports less false positive than the rest of the methods.

\subsection{Comparison of linear complexity methods from the perspective of outlierness type}\label{sec::outliernessType}
\begin{figure}
    \centering
    \includegraphics[width=\linewidth, keepaspectratio]{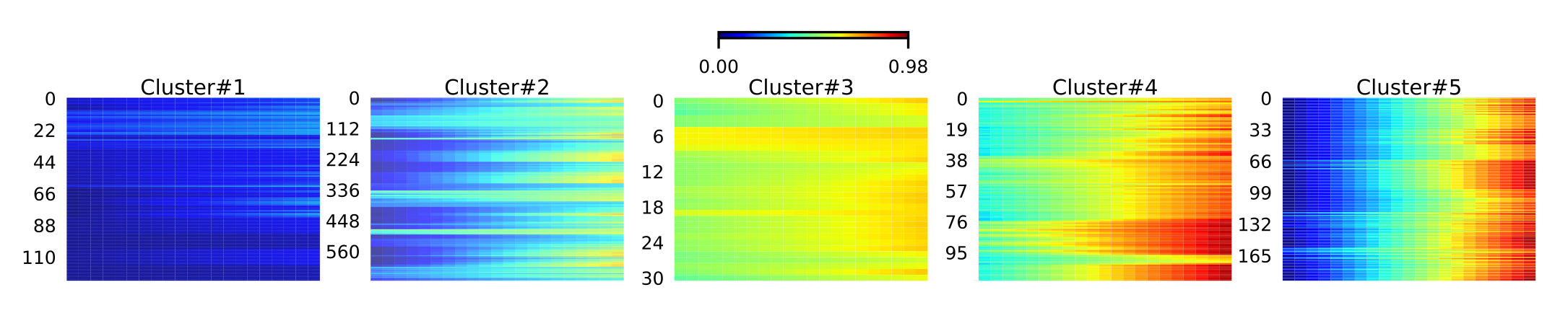}
    \caption{Performance comparison of linear time complexity outlier detection methods - FiRE.1, \glsxtrshort{fire}, and \glsxtrshort{hbos} based on \textit{o-score}. The datasets are grouped into 5 different clusters using hierarchical clustering. Cluster\#1 consists of 127 (11.2\% of the entire collection of datasets), cluster\#2 has 667 (59.1\%), cluster\#3 has 31 (2.7\%), cluster\#4 has 113 (10\%), and cluster\#5 has 191 (16.9\%) datasets. Every row in the heatmap represents a dataset. The \textit{o-score} of a given dataset is distributed in 20 bins of a histogram. The bin edges are arranged in columns along the corresponding row of the dataset. The distribution of \textit{o-score} varies across clusters. }
    \label{fig:comparison_clusters}
\end{figure}

The fitness of the linear time algorithms viz. \glsxtrshort{fire}, FiRE.1 and \glsxtrshort{hbos} was evaluated on varied data types, clustered based on the abundance of local and global outliers. For data grouping, we used the \textit{o-score} with $\phi=10$ (section~\ref{outliernessMetric}). For a given dataset, an \textit{o-score} is assigned to every point marked as an outlier. A dataset with large number of local (global) outliers is assigned a low (high) \textit{o-score}. For each dataset, a univariate histogram with 20 bins is constructed using the outliers' \textit{o-score}. The resultant 21 bin edges are used as features for every dataset. The matrix of size $\textit{total\_datasets} \times 21$ is then clustered into 5 different groups using hierarchical clustering (\textit{SciPy}~\cite{scipy} function scipy.cluster.hierarchy.cut\_tree) as shown in Figure~\ref{fig:comparison_clusters}. A cluster was composed of datasets with similar \textit{o-score}. The count of datasets varied across clusters. Cluster\#2 was the largest with $\sim$59\% of the total datasets. Figure~\ref{fig:comparison_clusters} shows varying \textit{o-score} across cluster\#1 to cluster\#4 in an ascending order. Cluster\#5 consists of datasets with wide range range of \textit{o-score}.

For each cluster, the methods were compared against each other and ranked for every evaluation measure. The performance of a method varies with clusters, as shown in Table~\ref{outlier_table}. For clusters \#1 and \#2, FiRE.1 has the smallest Friedman rank for nearly all evaluation measures. Clusters \#1 and \#2 comprise of datasets with the smallest \textit{o-score}, which are rich with local outliers. On the other end, cluster\#4 is composed of datasets with the largest \textit{o-scores}. \glsxtrshort{fire} has the smallest Friedman rank. \glsxtrshort{hbos} has the smallest Friedman ranking for cluster\#3, which consists of 31 datasets; this cluster mainly consists of non-duplicate and normalized versions of the Hepatitis dataset. In cluster\#5, FiRE.1 outperforms others when compared using \glsxtrshort{pan}, $\textit{Adjusted P@}n$, and \glsxtrshort{rocauc}. However, the performance of \glsxtrshort{hbos} is better for the \glsxtrshort{AP} and \textit{Adjusted AP} metrics. Clearly, it is imperative to evaluate the performance of a method on a given dataset from the context of different evaluation measures. The overall performance of FiRE.1 is superior when averaged across all clusters, closely followed by \glsxtrshort{hbos} in terms of \glsxtrshort{AP} and \textit{Adjusted AP}.

\begin{table}
    \caption{The performance of FiRE.1~\cite{gupta2021linear}, \glsxtrshort{fire}~\cite{fire}, and \glsxtrshort{hbos}~\cite{hbos} is evaluated on 5 different clusters. A cluster consists of datasets with similar distributions of \textit{o-score}. Different clusters consist of a different count of datasets. For every cluster, the performances have been compared from the context of all evaluation measures and graded using Friedman ranking. The lowest value of Friedman ranking across methods for a given measure is boldfaced. }
    \begin{center}
        \begin{tabular}{|c|c|c|c|c|c|c|c|}
            \toprule
            Algorithm & Cluster & Cluster & \textit{Adjusted} & \textit{Adjusted} &  &  & \textit{ROC-} \\ 
                      & ID & size & \textit{AP} & \textit{P@n} & \textit{AP} & \textit{P@n} & \textit{AUC} \\
            \midrule
            FiRE.1&\multirow{3}{*}{1}&\multirow{3}{*}{127}&\textbf{0.546}&\textbf{0.506}&\textbf{0.552}&\textbf{0.504}&\textbf{0.625}\\
            FiRE&&&0.888&0.882&0.880&0.877&0.733\\
            HBOS&&&0.708&0.764&0.711&0.758&0.683\\
            \hline
            FiRE.1&\multirow{3}{*}{2}&\multirow{3}{*}{667}&\textbf{0.659}&\textbf{0.621}&\textbf{0.655}&\textbf{0.623}&0.726\\
            FiRE&&&0.682&0.740&0.683&0.740&\textbf{0.649}\\
            HBOS&&&0.694&0.741&0.696&0.742&0.709\\
            \hline
            FiRE.1&\multirow{3}{*}{3}&\multirow{3}{*}{31}&0.699&0.830&0.699&0.830&\textbf{0.610}\\
            FiRE&&&0.911&0.823&0.905&0.823&0.876\\
            HBOS&&&\textbf{0.526}&\textbf{0.754}&\textbf{0.518}&\textbf{0.754}&0.624\\
            \hline
            FiRE.1&\multirow{3}{*}{4}&\multirow{3}{*}{113}&0.796&0.761&0.799&0.768&0.694\\
            FiRE&&&\textbf{0.613}&\textbf{0.687}&\textbf{0.617}&\textbf{0.692}&\textbf{0.612}\\
            HBOS&&&0.640&0.812&0.647&0.812&0.757\\
            \hline
            FiRE.1&\multirow{3}{*}{5}&\multirow{3}{*}{191}&0.636&\textbf{0.571}&0.641&\textbf{0.574}&\textbf{0.595}\\
            FiRE&&&0.826&0.811&0.828&0.811&0.803\\
            HBOS&&&\textbf{0.556}&0.638&\textbf{0.557}&0.637&0.639\\
            \hline
            FiRE.1&\multirow{3}{*}{overall}&\multirow{3}{*}{1129}&\textbf{0.657}&\textbf{0.619}&\textbf{0.656}&\textbf{0.621}&\textbf{0.686}\\
            FiRE&&&0.729&0.765&0.729&0.765&0.687\\
            HBOS&&&0.662&0.734&0.664&0.733&0.697\\
            \bottomrule
        \end{tabular}
        \label{outlier_table}
    \end{center}
\end{table}

\subsection{Performance comparison of linear-time methods on large datasets}~\label{sec::otherDatasets}
The performances of methods have been compared on additional datasets to illustrate their wide applicability on extremely large database sizes. \textit{Cod-RNA}~\cite{codrna} is a genome sequenced data that consists of non-coding RNAs (ncRNAs) as an outlier class. \textit{Protein-homology}~\cite{2004kdd} was used for the KDD-Cup 2004. It was used to identify proteins that are homologous to a native sequence. Non-homologous sequences were labeled as outliers. \textit{Poker}~\cite{cattral2002evolutionary} dataset from UCI repository~\cite{Dua:2019} categorizes a set of 5 cards in hands from 4 different suits into one of the 10 categories. Out of those categories, 8 are rare and represent a minor fraction of the data. \textit{RCV1}~\cite{rcv1} consists of newswire stories produced by Reuters and is used mainly for text categorization. It is a multi-label dataset and consists of 103 different labels. The top 10 labels with the highest number of sample count represent the inlier category, and samples corresponding to bottom 30 labels are outliers (similar to processing steps in Huang \textit{et al}~\cite{streamingvldb}). An inlier category may cover the entire dataset. Thus, when the samples are categorized, an outlier category takes precedence over an inlier one. Table~\ref{other_datasets_stats} presents some additional statistics about the datasets such as sample size, dimensions, and outlier percentage.

\begin{figure}
    \centering
    \includegraphics[width=\linewidth, keepaspectratio]{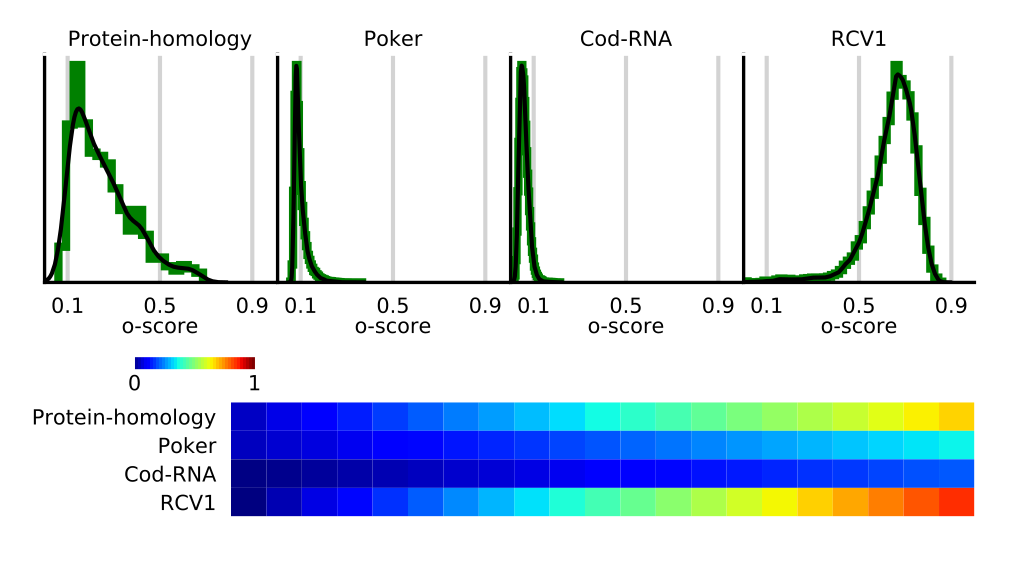}
    \caption{Density plot and heatmap illustrating the distribution of \textit{o-score} for 4 different large size datasets. The density plot shows the frequency distribution of \textit{o-score} for outliers. Every row in the heatmap represents a dataset. The \textit{o-score} of a given dataset is distributed in 20 bins of a histogram. The bin edges are arranged in columns along the corresponding row of the dataset.}
    \label{fig:bigdata_heatmap}
\end{figure}

For large datasets, we could only evaluate the performance of the linear time algorithms. Specifics about the parameter tuning is furnished in section~\ref{tuning_params}. Figure~\ref{fig:bigdata_heatmap} shows the distribution of outliers' \textit{o-score} in a dataset. \textit{Poker}, and \textit{Cod-RNA} mainly consist of local outliers, while \textit{Protein-homology}, and \textit{RCV1} consist of a mixture of local and global outliers.

Table~\ref{more_datasets} captures metrics depicting the performances of \glsxtrshort{fire}, FiRE.1, and \glsxtrshort{hbos} on these datasets for all evaluation measures. FiRE.1 offers the best performance on \textit{Cod-RNA}, and \textit{Poker} datasets, that mainly consist of local outliers (as shown in Figure~\ref{fig:bigdata_heatmap}). On \textit{RCV1}, \glsxtrshort{fire} and FiRE.1 performed competitively. \glsxtrshort{hbos} yielded the best performance on \textit{Protein-homology} dataset, followed closely by FiRE.1. Both \textit{RCV1} and \textit{Protein-homology} consist of a mixture of local and global outliers.

\begin{table}[htbp]
    \caption{Summary of the large datasets}
    \begin{center}
        \begin{tabular}{||c|c|c|c||}
            \hline
            Dataset & Samples & Features & Outliers(\%) \\ [0.5ex]
            \hline\hline
            \textit{Cod-RNA}~\cite{codrna} & 488565 & 8 & 33.33\%\\
            \textit{Protein-homology}~\cite{2004kdd} & 145751 & 74 & 0.89\%\\
            \textit{Poker}~\cite{cattral2002evolutionary} & 1025010 & 10 & 7.6\%\\
            \textit{RCV1}~\cite{rcv1} & 804414 & 1000 & 4.3\%\\
            \hline
        \end{tabular}
        \label{other_datasets_stats}
    \end{center}
\end{table}

\begin{table}[htbp]
    \caption{The performances of FiRE.1~\cite{gupta2021linear}, \glsxtrshort{fire}~\cite{fire}, and \glsxtrshort{hbos}~\cite{hbos} are compared on 4 large datasets. The values of the following evaluation measures are reported: \textit{Adjusted AP}, \textit{Adjusted P@$n$}, \glsxtrshort{AP}, \glsxtrshort{pan}, and \glsxtrshort{rocauc}. A method with the highest value of evaluation measure for a given dataset is boldfaced.}
    \begin{center}
        \begin{tabular}{|c|c|c|c|c|c|c|}
            \toprule
            & &\textit{Adjusted} & \textit{Adjusted} & & & \textit{ROC-} \\ 
            Dataset & Algorithm &\textit{AP} & \textit{P@n} & \textit{AP} & \textit{P@n} & \textit{AUC} \\ 
            \midrule
            \multirow{3}{*}{Cod-RNA~\cite{codrna}}&FiRE.1&\textbf{0.35}&\textbf{0.3701}&\textbf{0.5667}&\textbf{0.58}&\textbf{0.684}\\
            &FiRE&0.0994&0.2089&0.3996&0.4726&0.58499\\
            &HBOS&0.0955&0.1144&0.397&0.4096&0.5548\\
            \hline
            \multirow{3}{*}{Protein-homology~\cite{2004kdd}}&FiRE.1&0.6787&0.659&0.6816&0.662&0.8823\\
            &FiRE&0.013&0.0112&0.0219&0.02&0.753\\
            &HBOS&\textbf{0.7147}&\textbf{0.6862}&\textbf{0.7173}&\textbf{0.689}&\textbf{0.9384}\\
            \hline
            \multirow{3}{*}{Poker~\cite{cattral2002evolutionary}}&FiRE.1&\textbf{0.7219}&\textbf{0.6419}&\textbf{0.7431}&\textbf{0.6692}&\textbf{0.9355}\\
            &FiRE&0.0438&0.0814&0.1168&0.1515&0.486\\
            &HBOS&0.0434&0.1199&0.1164&0.1871&0.5303\\
            \hline
            \multirow{3}{*}{RCV1~\cite{rcv1}}&FiRE.1&0.0096&\textbf{0.024}&0.0519&\textbf{0.0657}&0.5792\\
            &FiRE&\textbf{0.0133}&0.0239&\textbf{0.0555}&0.0656&\textbf{0.6068}\\
            &HBOS&0.0108&0.0154&0.0531&0.05758&0.5819\\
            \bottomrule
        \end{tabular}
        \label{more_datasets}
    \end{center}
\end{table}

\section{Conclusions}\label{sec::conclusion}
Despite the existence of several methods for outlier detection, only a few scale well to large datasets in high dimensions. Most reported methods require $O(N^2)$ operations to identify $k$ nearest neighbors for every point, or $O(N^2)$ memory to calculate the distances between data points. Some reported approaches compromise on feature coverage to keep the computational complexity low. \glsxtrshort{fire} and FiRE.1 address both these concerns. Both \glsxtrshort{fire} versions have a run time complexity of $O(N)$. They take into account the entire spectrum of dimensions by screening a large number of feature subsets. \glsxtrshort{hbos} has $O(N)$ complexity as well for fixed bin widths. However, unlike \glsxtrshort{fire}, it assumes independence between features.

\glsxtrshort{fire} was designed to find global outliers in the data generated by a specific technology in molecular biology. Local outliers are generally considered as artifacts or noise in such datasets. Keeping this in mind, \glsxtrshort{fire} was designed using the Sketching process. The granularity of the space in sketching is dependent on the size of the sub-space (or the number of features). Hence, to achieve the required granularity to find local outliers in datasets with fewer features, \glsxtrshort{fire} may need to repeat features to split the space into finer regions. This process will also increase the required computational cost. FiRE.1 proposes a solution to this problem. FiRE.1 works in the sub-space (or original space) and quantizes the projection in a random direction as opposed to the sketching, which binarizes the features by random thresholds. The quantization-width provides explicit control over the size of bins, making it easier to tune the FiRE.1 to detect local outliers. Since FiRE.1 works in the sub-space (or original space), it's computational cost is also optimal.

Experiments on about 1000 datasets show, that the performance of different methods may vary depending on the choice of evaluation measure. We used Friedman ranking to overcome this ambiguity. The overall Friedman ranking of individual methods shows that FiRE.1 outperforms others for all evaluation measures. Our analysis also demonstrates that a method's performance is linked to the abundance of local and global outliers in specific data. While \glsxtrshort{fire} outperforms existing best-practice methods on datasets rich with global outliers, FiRE.1 performs consistently well on all types of datasets and exceptionally well on datasets with abundant local outliers. The performance on large datasets also reinforces this.


\chapter{Enhash: A Fast Streaming Algorithm for Concept Drift Detection\protect\footnote{The work presented in this chapter has been published as a research paper titled ``\textit{Enhash: A fast streaming algorithm for concept drift detection}'' in ESANN proceedings (2021).}}\label{chapter::4}

\section{Introduction}
A data stream environment is often characterized by large volumes of data that flow rapidly and continuously. These are processed in an \textit{online} fashion to accommodate data that cannot reside in main memory. A streaming data environment is commonly used for tasks such as making recommendations for users on streaming platforms~\cite{subbian2016recommendations}, and real-time analysis inside IoT devices~\cite{atzori2010internet}. In such a stream, the underlying data distribution may change, and this phenomenon is referred to as \textit{concept drift}. Formally, the posterior probability of a sample's class changes with time. Consequently, the method must also be able to adapt to the new distribution. To adapt to a new \textit{concept}, the method may require supplemental or replacement learning. Tuning a model with new information is termed as supplemental learning. Replacement learning refers to the case when the model's old information becomes irrelevant, and is replaced by new information. A shift in the likelihood of observing a data point $x$ within a particular class when class boundaries are altered, is called \textit{real concept drift}. \textit{Concept drift} without an overlap of true class boundaries, or an incomplete representation of the actual environment, is referred to as \textit{virtual concept drift}. In \textit{virtual concept drift}, one requires supplemental learning, while \textit{real concept drift} requires replacement learning~\cite{gama2014survey}. The other common way to categorize \textit{concept drift} is determined by the speed with which changes occur~\cite{de2019overview}. Hence, drift may be \textit{incremental}, \textit{abrupt} or \textit{gradual}. A \textit{reoccurring drift} is one that emerges repeatedly. Thus, in order to handle \textit{concept drift}, a model must be adaptive to non-stationary environments.

Several methods have been recently proposed to handle \textit{concept drift} in a streaming environment. The most popular of these are ensemble learners~\cite{learn++,learnNse,onlineSmotebagging,leverageBagging,arf,accuracyWeightedEnsemble,dwm,additiveExpertEnsemble}. As the data stream evolves, an ensemble method selectively retains a few learners to maintain prior knowledge while discarding and adding new learners to learn new knowledge. Thus, an ensemble method is quite flexible, and maintains the \textit{stability-plasticity} balance~\cite{lim2003online} i.e. retaining the previous knowledge (\textit{stability}) and learning new concepts (\textit{plasticity}). 

We propose \textit{Enhash}, an ensemble learner that employs projection hash~\cite{indyk1998approximate} to handle \textit{concept drift}. For incoming samples, it generates a hash code such that similar samples tend to hash into the same bucket. A gradual forgetting factor weights the contents of a bucket. Thus, the contents of a bucket are relevant as long as the incoming stream belongs to the \textit{concept} represented by them. 

\section{The proposed method: Enhash}\label{method}

Several recent methods employ hashing for online learning and outlier detection~\cite{sathe2016subspace,fire}. We propose Enhash, an ensemble learner, that employs hashing for \textit{concept drift} detection. Let $x_{t}\in \mathbb{R}^{d}$ represent a sample from a data stream $S$ at time step $t$ and let $y\in \{1,2,...,C\}$ represent its corresponding \textit{concept}, where $C$ is the total number of \textit{concepts}. Further, let us assume a family of hash functions $H$ such that each $h_{l}\in H$ maps $x_{t}$ to an integer value. The hash code $h_{l}(x_{t})$ is assigned to $x_{t}$ by hash function $h_{l}$. A bucket is a set of samples with the same hash code; both these terms are used interchangeably. The total number of samples in bucket $h_{l}(x_t)$ is denoted by $N_{h_{l}(x_t)}$. Further assume that $N$ samples have been seen and hashed from the stream so far, of which $N_{c}$ samples belong to the \textit{concept class} $c$ such that $\sum_{c=1}^{c=C}N_{c} = N$. Based on the evidence from the data stream seen so far, the probability of bucket $h_{l}(x_{t+1})$ is given by
\begin{align}
\label{eq::probH}
p(h_{l}(x_{t+1})) = \frac{N_{h_{l}(x_{t+1})}}{N}
\end{align}
and prior for class $c$ is given by
$p(c) = N_{c}/N$. Assuming, $(N_{h_{l}(x_{t+1})})_{c}$ represents the samples of \textit{concept} $c$ in bucket $h_{l}(x_{t+1})$. Hence, the likelihood of $x_{t+1}$ belonging to \textit{concept} $c$ in bucket $h_{l}(x_{t+1})$ is given by
\begin{align}
\label{eq::likelihood}
p(h_{l}(x_{t+1}) | c) = \frac{(N_{h_{l}(x_{t+1})})_c}{N_{c}}
\end{align}
The probability of $x_{t+1}$ belonging to class $c$ is given by
\begin{align}
\label{eq::posterior}
p(c|h_{l}(x_{t+1})) = \frac{p(h_{l}(x_{t+1}) | c)p(c)}{p(h_{l}(x_{t+1}))} = \frac{(N_{h_{l}(x_{t+1})})_{c}}{N_{h_{l}(x_{t+1})}}
\end{align}
Equation~(\ref{eq::posterior}) is simply the normalization of counts in bucket $h_{l}(x_{t+1})$.

To predict the \textit{concept class} of $x_{t+1}$, an ensemble of $L$ such hash functions can be employed and the weight for each \textit{concept class} is computed as
\begin{align}
\label{eq::classweight}
\hat{p}_{c} = \sum_{l=1}^{l=L} \log\Big(1 + p(c|h_{l}(x_{t+1}))\Big)
\end{align}
and the \textit{concept class} is predicted as
\begin{align}
\label{eq::class}
\hat{y} = \arg\max_{c\in\{1,..,C\}}\hat{p}_{c}
\end{align}
To accommodate an incoming sample of class $c$, the bucket is updated as
\begin{align}
\label{eq::update}
(N_{h_{i}(x_{t+1})})_{c} = 1 + (N_{h_{i}(x_{t+1})})_{c}\:\forall c\in\{1,..,C\}
\end{align}

Enhash utilizes a simple strategy as described above to build an ensemble learner for \textit{concept drift} detection.

\begin{figure}
    \centering
    \includegraphics[width=\textwidth,keepaspectratio]{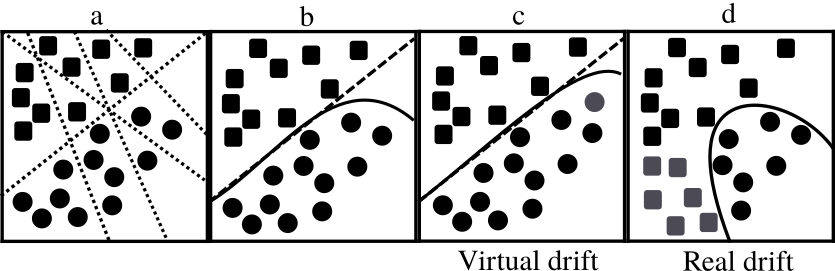}
    \caption{Enhash accommodates both virtual and real drift.}
    \label{fig:drift}
\end{figure}

In Enhash, the projection hash family is selected as a base learner. Here, a hash function involves the dot product of hyperplane $w^{(l)}$ and sample $x_{t}$. It is defined as
\begin{align}
\label{eq::hash}
h_{l}(x_{t}) = \lfloor\frac{1}{\textit{bin-width}}\Bigg(\sum_{j=1}^{j=d}\big(w^{(l)}_{j}*(x_{t})_{j}\big) + \text{bias}^{(l)}\Bigg)\rfloor
\end{align}
where, \textit{bin-width} is quantization width, $w^{(l)}_{j}$ and bias$^{(l)}$ can be sampled from any desired distribution. In our implementation, $w^{(l)}_{j}\sim N(0,1)$ and \\$\text{bias}^{(l)}\sim[-\textit{bin-width}, \textit{bin-width}]$.

Effectively, each $h_{l}$ (\ref{eq::hash}) divides the space into equally spaced unbounded regions of size \textit{bin-width} (earlier referred to as bucket). Equation~(\ref{eq::posterior}) computes the probability of each \textit{concept class} in a region. An ensemble of hash functions makes all regions bounded. The weight of a \textit{concept class} in the bounded region is computed using (\ref{eq::classweight}). An absolute value of \textit{concept class} is assigned to each region in (\ref{eq::class}). Figure~\ref{fig:drift}a shows the arrangement of randomly generated hyperplanes. The solid line in Figure~\ref{fig:drift}b shows the inferred decision boundary (learned distribution) after an absolute assignment of \textit{concept class} to every bounded region. The dashed line in Figure~\ref{fig:drift}b depicts the true decision boundary (true distribution).

Assume that Figure~\ref{fig:drift}b shows the current stage of learner at time $t$. At time $t+1$, a new sample $x_{t+1}$ arrives (gray sample in Figure~\ref{fig:drift}c). After updating the bucket (\ref{eq::update}), the learned distribution shifts and moves towards the true distribution. Hence, Enhash accommodates virtual drift present in the data stream.

Figure~\ref{fig:drift}d depicts the real drift when the true distribution evolves. This requires forgetting some of the previously learned information. Suppose that sample $x_{t+\Delta t}$ hashes to bucket $h_{l}(x_{t+\Delta t})$ at time say, $t + \Delta t$. Assume that previously, $x_t$ from a different \textit{concept}, was hashed into this bucket. In order to accommodate forgetting, Enhash employs a decay factor multiplier to $p(c|h_{l}(x_{t+\Delta t}))$ (\ref{eq::classweight}) while weighting a bounded region.
\begin{align}
\label{eq::updatePosterior}
\hat{p}_c = \sum_{l=1}^{l=L}\log\Big(1 + 2^{-\lambda \Delta t} p(c|h_{l}(x_{t+\Delta t}))\Big)
\end{align}
where $\lambda$ is the decay rate.
The update rule for the bucket (\ref{eq::class}) is also changed to reflect the new \textit{concept class} as follows
\begin{align}
\label{eq::updateBucketNew}
(N_{h_{l}(x_{t+1})})_c = 1+ 2^{-\lambda \Delta t} (N_{h_{l}(x_{t+1})})_c
\end{align}
Setting $\lambda=0$ will reduce these equations to the base case.

In order to break ties in $p(c|h_{l}(x_{t}))$, the class weight in the region is also weighted by the distance of the sample $x_{t}$ from the mean of class samples in bucket $h_{l}(x_{t})$, i.e. $\text{mean}_{h_{l}(x_t)}^{c}$
\begin{align}
\label{eq::distance}
\text{dist}_{c}(x_{t}) = \sqrt{\sum_{i=1}^{i=d}\Big((x_{t})_{i} - (\text{mean}_{h_{l}(x_t)}^{c})_{i}\Big)^{2}}
\end{align}
\begin{align}
\label{eq::finalUpdate}
\hat{p}_c = \sum_{l=1}^{l=L}\log\Big(1 + \frac{2^{-\lambda \Delta t} p(c|h_{l}(x_{t}))}{dist_{c}(x_{t})})\Big)
\end{align}

All remaining ties are broken arbitrarily.

\subsection{Implementation details}
Algorithm~\ref{algo::methodEdit} describes Enhash's pseudo-code. We refer to an instance of the ensemble as an estimator. The parameter \textit{bin-width} determines the granularity of the buckets, and $\lambda$ represents the rate of decay that accounts for gradual forgetting~\cite{klinkenberg2004learning}. 
Every estimator $l$ generates a hash code for incoming stream samples and hashes them into bucket $b^{(l)}$. The hash code for a sample $x$ is compute using (\ref{eq::hash}).
The timestamp when a sample of class $y$ was last hashed into a bucket $b^{(l)}$ is stored in $tstamp^{(l)}[b^{(l)}][y]$. $tstamp^{(l)}$ is an infinitely indexable 2D-array. An infinitely indexable ND-array is a data structure that has N dimensions and can store values at any arbitrary combinations of indexes in those dimensions. In practice, an infinitely indexable ND-array can be created using maps or dictionaries. The recent access time of the hash code $b^{(l)}$ can be retrieved by selecting the maximum value in $tstamp^{(l)}[b^{(l)}]$.
The count of samples in bucket $b^{(l)}$ is stored in another infinitely indexable 2D-array $counts^{(l)}$ at an index $counts^{(l)}[b^{(l)}]$. The information required to break ties during prediction is stored in $sCounts^{(l)}$ and $sAcc^{(l)}$. Variable $sCounts^{(l)}$, an infinitely indexable 2D-array, keeps track of the number of samples from every class $y$ falling into the bucket $b^{(l)}$. $sAcc^{(l)}$, an infinitely indexable 3D-array, stores the class-wise vector sum of all the samples falling into the bucket $b^{(l)}$. The last dimension in this array belongs to the features in the dataset.

Enhash has two phases. In the first one, Enhash predicts the class of a new sample. Assuming that sample $x$ falls into bucket $b^{(l)}$, its distance from all the cluster centers in the bucket is computed (steps~\ref{step::mean} and \ref{step::dist}). The variable $cweights$ accumulates the prediction for a sample $x$ via each estimator. This variable is also an infinitely indexable array so that it can accommodate unseen class labels. To get the prediction for sample $x$ from an estimator $l$, a decayed value of count (weighted by distance) is computed (step~\ref{step::pred}). The decay factor $2^{-\lambda \Delta t_{1}}$ , depends upon the decay parameter $\lambda$ and the difference of the current time and the last access time of bucket $b^{(l)}$. The decay factor determines whether the previous value in $counts^{(l)}[b^{(l)}]$ is relevant or not, and hence, introducing the forgetting effect in the algorithm. In effect, if the difference in the time is large, then decay is almost zero, and this emulates local replacement in the bucket $b^{(l)}$~\cite{carmona2011gnusmail}. On the other hand, $\lambda$ reduces the effect of samples (possibly, of the same \textit{concept}) hashing into a bucket. The higher is the value of $\lambda$, then more is the rate of decay. Collectively, $\lambda$, and time difference play an important role in drift detection. The log-transformed value of the prediction is added to $cweights$. A pseudo-count of $1$ is added during log-transformation to handle $0$ or near-zero values in prediction. After accumulating predictions from every estimator, a class with maximum weight is designated as the class of sample $x$ (step~\ref{step::classassignment}).

In the second phase, the variables are updated to accommodate recent changes in the \textit{concept}. Assume that a recent sample belongs to class $y$. Hence, the value in $tstamp[b^{(l)}][y]$ is set to the current timestamp. Further, to update the effective count in bucket for class $y$ in $counts^{(l)}[b^{(l)}][y]$, the present value is decayed by the difference of the current time and the last seen time of sample from class $y$ and then incremented by $1$ (steps~\ref{step::cUpdate}- \ref{step::countupdate}). This introduces the forgetting phenomenon during updates and also handles spurious changes. For example, if a bucket was accustomed to seeing samples from a particular class and the sudden arrival of a sample from another class that had not been seen by the bucket for a long time, it would not alter the bucket's prediction abruptly. However, after seeing a few samples from the new class, the bucket's confidence will grow gradually towards the recent trends. The value in $counts^{(l)}[b^{(l)}]$ is normalized for numerical stability. Finally, $sCounts^{(l)}[b^{(l)}[y]$ is incremented by 1 and a new sample $x$ is added in $sAcc^{(l)}[b^{(l)}[y]$.

In essence, samples with a similar \textit{concept} are most likely to have the same hash code and hence, share the bucket. For an evolving stream, thus, different buckets are populated. The weight associated with the bucket is gradually incremented when samples of the same \textit{concept} arrive. The contents of a bucket are more relevant when the \textit{concept} reoccurs in the near future than in a faraway future.

Formally, the temporal nature of the posterior distribution of a sample $x$ belonging to a class $y$ is modeled as the Bayes posterior probability $P(y|x) = P(y) P(x|y)/P(x)$. Let $n$ and $n_{y}$ denote the count of total samples and samples of a class $y$, respectively. In the proposed method, for a given estimator $l$, $P(y) = n_{y}/n$, $P(x|y) = counts^{(l)}[b^{(l)}][y]/n_{y}$, and $P(x) = \sum_{j}(counts^{(l)}[b^{(l)}][j])/n$. \\Thus, $P(y|x) = counts^{(l)}[b^{(l)}][y]/\sum_{j}(counts^{(l)}[b^{(l)}][j])$ and hence, information in\\ $counts^{(l)}[b^{(l)}][y]$ accounts for \textit{concept drift}. For virtual drift, the contents of $b^{(l)}$ may only be supplemented with the additional information from the distribution. For real drift, however, the previous contents of $b^{(l)}$ may be discarded via the decay factor.

\begin{algorithm}
    \setstretch{1.35}
    \caption{Enhash}\label{algo::methodEdit}
    \begin{algorithmic}[1]
        \State \textbf{Input:} Data stream $S$
        \State \hspace{10mm} $L \gets \text{Number of estimators}$
        \State \hspace{10mm} $\textit{bin-width} \gets \text{Width of bucket}$
        \State \hspace{10mm} $\lambda \gets \text{Rate of decay}$
        \State \textbf{Initialize:} $t \gets 0$
        \State \hspace{15mm} For every estimator $l \in \{1,...,L\}$
        \State \hspace{20mm} $counts^{(l)} \gets \text{infinitely indexable 2D-array}$
        \State \hspace{20mm} $tstamp^{(l)} \gets \text{infinitely indexable 2D-array}$
        \State \hspace{20mm} $sCounts^{(l)} \gets \text{infinitely indexable 2D-array}$
        \State \hspace{20mm} $sAcc^{(l)} \gets \text{infinitely indexable 3D-array}$
        \State \textbf{Run:}
        \While{$HasNext(S)$}
        \State $(x,y) \gets next(S)$
        \State $t \gets t + 1$
        \State $cweights \gets \text{array initialized with 0s}$
        \For{$l \in \{1,...,L\}$}
        \State $b^{(l)}$ = generate hash code using (\ref{eq::hash}) for estimator $l$\label{step::hashing}
        \State $sMean = \frac{sAcc^{(l)}[b^{(l)}]}{sCounts^{(l)}[b^{(l)}]}$\label{step::mean}
        \State $dist = \sqrt{\sum((sMean - x)^{2})}$\label{step::dist}
        \State $\Delta t_{1} = t - \max_{j}(tstamp^{(l)}[b^{(l)}][j])$
        \State $v = 2^{-\lambda \times \Delta t_{1}} \times \frac{counts^{(l)}[b^{(l)}]}{dist}$\label{step::pred}
        \State $cweights = cweights + \log(1+v)$ \label{step::predend}
        \State $\Delta t_{2} = t - tstamp^{(l)}[b^{(l)}][y]$\label{step::updatestart}

        \State $counts^{(l)}[b^{(l)}][y] = 1 +  (2^{-\lambda \times \Delta t_{2}} \times counts^{(l)}[b^{l)}][y])$\label{step::cUpdate}
        \State $counts^{(l)}[b^{(l)]} = \frac{counts^{(l)}[b^{(l)}]}{\sum_{j}(counts^{(l)}[b^{(l)}][j])}$\label{step::countupdate}
        \State $tstamp^{(l)}[b^{(l)}][y] = t$
        \State $sCounts^{(l)}[b^{(l)}][y] = 1 +  sCounts^{(l)}[b^{(l)}][y]$
        \State $sAcc^{(l)}[b^{(l)}][y] = x + sAcc^{(l)}[b^{(l)}][y]$ \label{step::updateend}
        \EndFor
        \State $\hat{y} \gets \arg\max(cweights)$\label{step::classassignment}
        \EndWhile
    \end{algorithmic}
\end{algorithm}

\subsection{Time complexity analysis}

There are three important steps in Algorithm~\ref{algo::methodEdit}, namely, the computation of hash function (step~\ref{step::hashing}), prediction of class (steps~\ref{step::mean}-\ref{step::predend}), and update of model parameters (steps~\ref{step::updatestart}-\ref{step::updateend}). In Enhash, the hash function computation is the dot product of a sample with weight parameters (\ref{eq::hash}). Let the time required to compute the hash function be given by $\psi(d)=\mathcal{O}(d)$. The prediction of a sample's class involves the computation of distances from cluster centers in the bucket (\ref{eq::distance}). Assuming that, Enhash has observed $C$ concept classes so far; then, the time required for prediction is given by $\phi(d, C)=\mathcal{O}(d*C)$. Model update involves updating the bucket's timestamp, the effective count of the \textit{concept class} in the bucket, the total samples hashed into the bucket, and accumulation of samples. Effective counts are also normalized in the update step which depends upon $C$. Let the time required to update model parameter be denoted by $\zeta(d, C)=\mathcal{O}(d+C)$. An estimator takes time $\mathcal{O}(\psi(d) + \phi(d, C) + \zeta(d, C)) = \mathcal{O}(d+d*C+d+C)=\mathcal{O}(d*C)$. These steps are repeated by all $L$ estimators, and hence, the total time complexity of Enhash, for an arbitrary dataset and hyper-parameters settings, is given by $\mathcal{O}(L*d*C)$.

\begin{table}[th!]
\caption{Time complexities of algorithms is compared. This table presents the complexity to process the $N$ samples from a stream. The base estimators are as per the default parameters of the corresponding classes in \texttt{scikit-multiflow} package. In the table $N$ represents number of samples, $d$ represents number of dimensions, $L$ represents number of estimators, $C$ represents the number of classes, $w$ is the window size, $k$ is the number of trials coming from Poisson distribution, and $s$ is the oversampling rate. To be noted, these are the simplified estimates of the time complexity.}
\label{table:enhash:time}
\centering
\makebox[\textwidth][c]{
\resizebox{1.3 \linewidth}{!}{ %
\begin{tabular}{| c | c | c |}
        \toprule
        Algorithm & Complexity & Remarks \\
        \hline
        & & \\
        Enhash~\cite{enhash} & O$(L*d*N*C)$ & \\
        & & \\
        \hline
        & & \\
        DWM~\cite{dwm} & O$(L*d*(N+C))$ & The base classifier is Naive Bayes' Classifier.\\
        & & \\
        \hline
        & & \\
        Learn$^{++}$.NSE~\cite{learnNse} & O$(L*(\log N + N*d*\log N))$ & The base classifier is decision tree. The best case complexity for training and testing is assumed in this estimate.\\
        & & \\
        \hline
        & & \\
        Learn$^{++}$~\cite{learn++} & O$(L*(\log N + N*d*\log N))$ & The base classifier is decision tree. The best case complexity for training and testing is assumed in this estimate.\\
        & & \\
        \hline
        & & \\
        LB~\cite{leverageBagging} & O$(L * N * k * \log N)$ &  The base classifier is \texttt{KNNClassifier} from \texttt{scikit-multiflow} package.\\
        & & \\
        \hline
        & & \\
        OB~\cite{adwin,oza2005online} & O$(L * N * k * (\log N + \log w))$ &  The base classifier is \texttt{KNNAdwinClassifier} from \texttt{scikit-multiflow} package.\\
        & & \\
        \hline
        && \\
        OSMOTEB~\cite{onlineSmotebagging}  & O$(L * N * k * s * (\log N + \log w))$ & The base classifier is \texttt{KNNAdwinClassifier} from \texttt{scikit-multiflow} package. \\
        && \\
        \hline
        & & \\
        AWE~\cite{accuracyWeightedEnsemble} & O$(L*d*(N+C))$ & The base classifier is Naive Bayes' classifier. \\
        & & \\
        \hline
        & & \\
        AEE~\cite{additiveExpertEnsemble}  & O$(L*d*(N+C))$ & The base classifier is Naive Bayes' classifier.\\
        & & \\
        \hline
        & & \\
        ARF~\cite{arf} & O$(L(\log N + N*d*\log N))$ & The best case complexity for training and testing is assumed in this estimate. \\
        & & \\
        \bottomrule
        \end{tabular}
    }
}
\end{table}

Thus, for a fixed setting of hyper-parameter $L$, the time complexity of Enhash to process a newly arrived sample for an arbitrary dataset is given by $\mathcal{O}{(d*C)}$. However, it can be argued that for a fixed dataset, the dimension of data $d$ and number of concept classes $C$ are fixed. Hence, Enhash will effectively take only a constant time, $\mathcal{O}(1)$, in the processing of a new sample. The comparison, in terms of time taken in processing $N$ samples from stream, with other algorithms is presented in Table~\ref{table:enhash:time}.  

\section{Experimental Setup}\label{setup}
Enhash's performance on \textit{concept drift} detection was compared with some widely used ensemble learners. These include Learn++~\cite{learn++}, $\text{Learn}^{++}\text{.NSE}$~\cite{learnNse}, Accuracy-Weighted Ensemble (\glsxtrshort{awe})~\cite{accuracyWeightedEnsemble}, Additive Expert Ensemble (\glsxtrshort{aee})~\cite{additiveExpertEnsemble}, DWM~\cite{dwm}, Online Bagging-ADWIN (\glsxtrshort{ob})~\cite{adwin,oza2005online}, Leveraging Bagging~\cite{leverageBagging}, Online SMOTE Bagging (\glsxtrshort{osmoteb})~\cite{onlineSmotebagging}, and \glsxtrshort{arf}~\cite{arf}. The implementation of these methods is available in \texttt{scikit-multiflow} \texttt{python} package~\cite{skmultiflow}. A fixed value of \textit{estimators} was used for all the methods. For methods such as Accuracy-Weighted Ensemble, $\text{Learn}^{++}\text{.NSE}$, Learn++, Leveraging Bagging, and Online Bagging-ADWIN, the maximum size of window was set $\min(5000, 0.1*n)$, where $n$ is the total number of samples. For the rest of the parameters, the default value was used for all the methods.

\subsection{Evaluation metrics for performance comparison}
We evaluated all the experiments in terms of time, memory consumption, and classifiers' performance. The memory consumption is measured in terms of RAM-hours~\cite{bifet2010fast}. Every GB of RAM employed for an hour defines one RAM-hour. The performance of a classifier is measured in terms of accuracy/error, Kappa M, and Kappa Temporal~\cite{bifet2015efficient}. Kappa M, and Kappa Temporal handle imbalanced data streams, and data streams that have temporal dependencies, respectively. We evaluated the classifiers' performance using the Interleaved Test-Train strategy~\cite{losing2016knn}. This strategy is commonly employed for incremental learning since every sample is used as a test and a training point as it arrives.

\subsection{Description of datasets}
We used 6 artificial/synthetic (Samples x Features)- transientChessboard (200000 x 2), rotatingHyperplane (200000 x 10), mixedDrift (600000 x 2), movingSquares (200000 x 2), interchangingRBF (200000 x 2), interRBF20D (201000 x 20), and 4 real datasets- airlines (539383 x 7), elec2 (45312 x 8), NEweather (18159 x 8), outdoorStream (4000 x 21) for all experiments. The datasets are available on \url{https://github.com/vlosing/driftDatasets}. The synthetic datasets simulate drifts such as abrupt, incremental, and virtual. The real datasets have been used in the literature to benchmark \textit{concept drift} classifiers. The count of samples varies from 4000 (in outdoorStream) to 600,000 (in mixedDrift). Also, outdoorStream has the maximum number of classes, i.e., 40. The summary of the description of datasets is available in Table~\ref{description}.

\subsection{System details}
All experiments were performed on a workstation with 40 cores using Intel Xeon E7-4800 (Haswell-EX/Brickland Platform) CPUs with a clock speed of 1.9 GHz, 1024 GB DDR4-1866/2133 ECC RAM and Ubuntu 14.04.5 LTS operating system with the 4.4.0-38-generic kernel.

\begin{table}
    \caption{Description of datasets.}
    \centering
        \resizebox{1\textwidth}{!}{
    \begin{tabular}{c c c c}
        \toprule
        Synthetic datasets & Samples x Features & Classes & Drift\\
        \midrule
        transientChessboard & 200,000 x 2 & 8 & Virtual\\
        rotatingHyperplane & 200,000 x 10 & 2 & Abrupt\\
        mixedDrift & 600,000 x 2 & 15 & Incremental, Abrupt, and Virtual\\
        movingSquares & 200,000 x 2 & 4 & Incremental\\
        interchangingRBF & 200,000 x 2 & 15 & Abrupt\\
        interRBF20D & 201,000 x 20 & 15 & Abrupt\\
    \bottomrule
    \end{tabular}
    \hspace{1cm}
    \begin{tabular}{c c c c}
        \toprule
        Real datasets & Samples x Features & Classes\\
        \midrule
        airlines & 539,383 x 7 & 2\\
        elec2 & 45,312 x 8 & 2\\
        NEweather & 18,159 x 8 & 2\\
        outdoorStream & 4,000 x 21 & 40\\
    \bottomrule
    \end{tabular}}
    \label{description}
\end{table}

\section{Tuning of parameters for Enhash}\label{sec::tuningEnhash}
The hyperparameters that govern the performance of Enhash constitute $L$ (number of estimators), and $\textit{bin-width}$ (quantization parameter). Even though the decay rate $\lambda$ is also one of the hyperparameters, its value does not require much tuning, and usually, $\lambda=0.015$ is preferred~\cite{sathe2016subspace}. However, making $\lambda=0$ is equivalent to removing the forgetting phenomenon and hence, worsens the performance (discussed further in Section~\ref{sec::ablation}).

\subsection{Constraints to tune L}
\begin{figure}
    \makebox[1 \textwidth][c]{
    \resizebox{1.1 \linewidth}{!}{ %
    \includegraphics[width=\textwidth,keepaspectratio]{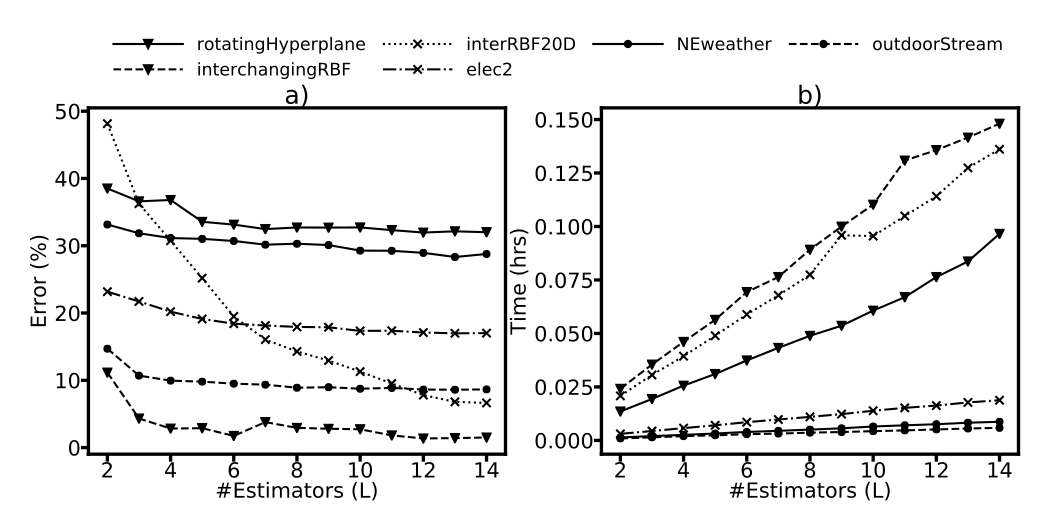}
    }}
    \caption{Tuning of $L$. For both synthetic and real datasets, we show a trend in performance metrics, Error (\%), and Time(hrs) for an increase in values of $L$. The value of $L$ varies from [2, 14]. For a given value of $L$, the running time is measured across all the samples for all the estimators in a configuration. The value of error is evaluated across all the samples for a given value of $L$.}
    \label{fig:tuneL}
\end{figure}

In Figure~\ref{fig:tuneL}, it is shown empirically that with an increase in the value of $L$, the performance of Enhash eventually saturates. The time taken by Enhash also increases with $L$. This may be attributed to the fact that for every sample arriving at time $t$, the insertion involves calculating the hash code of the sample for every estimator $l$. It should be emphasized that, as shown empirically in Figure~\ref{fig:tuneL}, only a few estimators are needed to achieve optimum performance. Further, the increase in time due to an increase in $L$ can be reduced through a parallel implementation of Enhash. In that case, evaluation of a hash code for a sample for each $l$ can be done independently of others. Consequently, a moderate value of $L = 10$ is used to perform all the experiments.

\subsection{Constraints to tune \textit{bin-width}}
\begin{figure}
    \makebox[1 \textwidth][c]{
    \resizebox{1.1 \linewidth}{!}{ %
    \includegraphics[width=\textwidth,keepaspectratio]{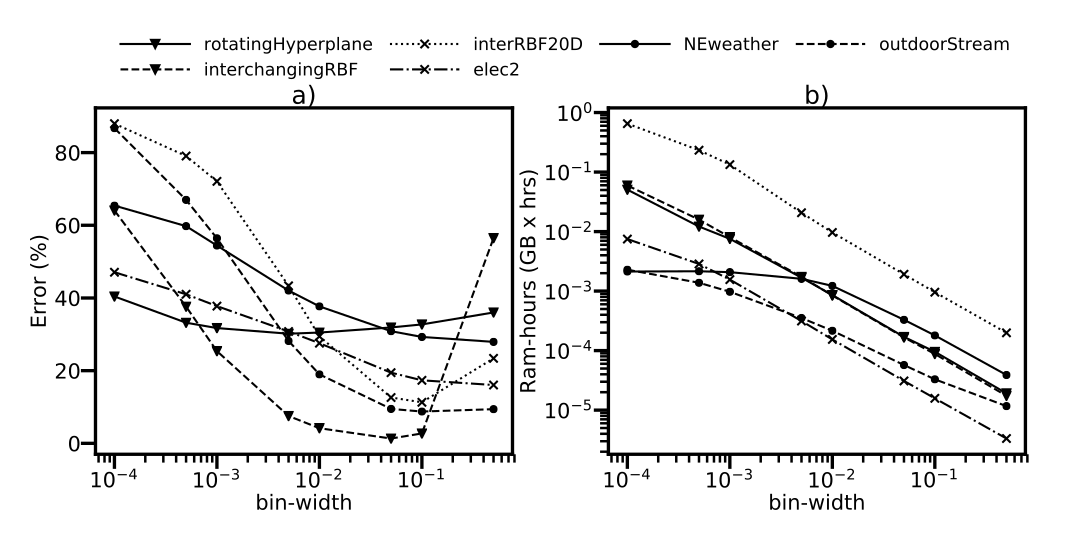}
    }}
    \caption{Tuning of $\textit{bin-width}$. For both synthetic and real datasets, we show a trend in performance metrics, Error (\%), and Ram-hours for different values of $\textit{bin-width}$. The value of $\textit{bin-width}$ varies in [0.0001, 0.0005, 0.001, 0.005, 0.01, 0.05, 0.1, 0.5]. a) The value of error is evaluated across all the samples for a given value of $\textit{bin-wdith}$. The values of $\textit{bin-width}$ on x-axis are on logarithmic scale. b) Ram-hours are calculated across all the samples for all the estimators for a given configuration. The values of $\textit{bin-width}$ on x-axis and Ram-hours on y-axis, both are on log scale.}
    \label{fig:tunebinwidth}
\end{figure}

The parameter $\textit{bin-width}$ divides the space into equally spaced unbounded regions of size $\textit{bin-width}$. The smaller is the value of $\textit{bin-width}$, the more granular is the division of space. In other words, for smaller values of $\textit{bin-width}$, the possible values of different hash codes (or buckets) increase rapidly. In the worst case, every sample may fall into a different bucket. Thus, the overall cost to store the contents of all buckets for every estimator grows exponentially. Figure~\ref{fig:tunebinwidth} shows empirically the effect of $\textit{bin-width}$ on memory consumption. The figure highlights that for small values of $\textit{bin-width}$, the overall memory requirement is extremely high. Also, the prediction of the \textit{concept} for the sample may be arbitrary for extremely small values of $\textit{bin-width}$ since there is no neighborhood information in the bucket. As a result, for every new sample, a new \textit{concept} may be falsely predicted. On the other extreme, for large values of $\textit{bin-width}$, samples from different classes may lead to frequent collision. As a result of this, the \textit{concept} of an arriving sample may not be predicted correctly due to confusion in the bucket. Thus, in general, an intermediate range of values for $\textit{bin-width} \in \{0.01, 0.1\}$ is more suitable for all datasets.

\begin{table}
\caption{Error (in \%) is reported to compare the performance of Enhash~\cite{enhash} with other methods. For a given dataset, the method with the least error is in boldface. Due to implementation constraint, $\text{Learn}^{++}\text{.NSE}$ could not run for the outdoorStream dataset.}
\makebox[1 \textwidth][c]{
\resizebox{1.1 \linewidth}{!}{ %
    \begin{tabular}{l@{\hspace{1em}} r@{\hspace{1em}} r@{\hspace{1em}} r@{\hspace{1em}} r@{\hspace{1em}} r@{\hspace{1em}} r@{\hspace{1em}} r@{\hspace{1em}} r@{\hspace{1em}} r@{\hspace{1em}} r}
        \toprule
        Dataset & DWM & $\text{Learn}^{++}$ & Learn++ & LB & OB & OSMO- & AWE & AEE & ARF & Enhash \\
        & \cite{dwm} & .NSE~\cite{learnNse} & \cite{learn++} & \cite{leverageBagging} & \cite{adwin,oza2005online} & -TEB~\cite{onlineSmotebagging} & \cite{accuracyWeightedEnsemble} & \cite{additiveExpertEnsemble} & \cite{arf} & \cite{enhash} \\
        \midrule
        transientChessboard&49.89&{\textcolor{darkgray}{3.67}}&\textbf{2.75}&13.13&29.62&24.28&89.78&85.37&27.56&18.84\\

        rotatingHyperplane&\textbf{10.11}&24.13&19.36&24.65&{\textcolor{darkgray}{15.08}}&19.75&16.27&18.10&16.40&32.72\\

        mixedDrift&65.77&39.46&49.67&{\textcolor{darkgray}{16.71}}&23.63&20.30&77.79&81.17&19.79&\textbf{12.88}\\

        movingSquares&{\textcolor{darkgray}{29.03}}&68.74&67.84&55.69&65.42&61.12&67.51&67.23&58.54&\textbf{13.29}\\

        interchangingRBF&7.02&75.43&82.79&4.61&37.67&22.05&83.49&82.58&{\textcolor{darkgray}{3.22}}&\textbf{2.72}\\

        interRBF20D&7.62&76.36&81.59&{\textcolor{darkgray}{5.62}}&37.02&21.00&84.90&82.35&\textbf{2.39}&11.30\\

        airlines&{\textcolor{darkgray}{37.42}}&42.92&37.53&43.36&39.23&40.66&38.49&39.17&\textbf{33.09}&41.66\\

        elec2&20.62&34.63&31.40&18.64&23.26&24.60&39.64&26.71&\textbf{11.59}&{\textcolor{darkgray}{17.34}}\\

        NEweather&29.52&29.20&24.63&27.20&\textbf{20.84}&22.74&30.72&30.78&{\textcolor{darkgray}{21.43}}&29.27\\

        outdoorStream&57.62&-&60.88&{\textcolor{darkgray}{9.33}}&34.15&21.25&78.55&42.45&26.12&\textbf{8.75}\\

        \bottomrule
    \end{tabular}
    }
    }
    \label{errort}
\end{table}

\section{Experimental Results}\label{results}
For all methods, the number of $estimators$ is considered as 10. In addition for Enhash, \textit{bin-width} was set to \{$0.1$, $0.01$\} and $\lambda$ was set to 0.015. Tables~\ref{errort},~\ref{kappamt}, and~\ref{kappatt} compare the performance of the methods in terms of error, KappaM, and KappaT respectively using Interleaved Test-Train strategy. For these measures, the performance of the proposed method was superior to Accuracy-Weighted Ensemble (\glsxtrshort{awe}), and Additive Expert Ensemble (\glsxtrshort{aee}) on 8 datasets, DWM, $\text{Learn}^{++}\text{.NSE}$, Online SMOTE Bagging (\glsxtrshort{osmoteb}), and Online Bagging-ADWIN (\glsxtrshort{ob}) on 7 datasets, Learn++, and Leveraging Bagging (\glsxtrshort{lb}) on 6 datasets, and \glsxtrshort{arf} on 5 datasets. The performance of Enhash supersedes all other methods for 4 datasets - mixedDrift, movingSquares, interchangingRBF, and outdoorStream.

\begin{table}
\caption{KappaM is tabulated to compare the performances of the methods. For a given dataset, the method with the highest value of KappaM is in boldface.}
\makebox[1 \textwidth][c]{
\resizebox{1.1 \linewidth}{!}{ %
    \begin{tabular}{l@{\hspace{1em}} r@{\hspace{1em}} r@{\hspace{1em}} r@{\hspace{1em}} r@{\hspace{1em}} r@{\hspace{1em}} r@{\hspace{1em}} r@{\hspace{1em}} r@{\hspace{1em}} r@{\hspace{1em}} r}
        \toprule
        Dataset & DWM & $\text{Learn}^{++}$ & Learn++ & LB & OB & OSMO- & AWE & AEE & ARF & Enhash \\
        & \cite{dwm} & .NSE~\cite{learnNse} & \cite{learn++} & \cite{leverageBagging} & \cite{adwin,oza2005online} & -TEB~\cite{onlineSmotebagging} & \cite{accuracyWeightedEnsemble} & \cite{additiveExpertEnsemble} & \cite{arf} & \cite{enhash} \\
        \midrule
        transientChessboard&0.43&\textcolor{darkgray}{0.96}&\textbf{0.97}&0.85&0.66&0.72&-0.03&0.02&0.68&0.78\\

        rotatingHyperplane&\textbf{0.80}&0.52&0.61&0.51&\textcolor{darkgray}{0.70}&0.60&0.67&0.64&0.67&0.35\\

        mixedDrift&0.23&0.54&0.42&\textcolor{darkgray}{0.80}&0.72&0.76&0.09&0.05&0.77&\textbf{0.85}\\

        movingSquares&\textcolor{darkgray}{0.61}&0.08&0.10&0.26&0.13&0.19&0.10&0.10&0.22&\textbf{0.84}\\

        interchangingRBF&0.92&0.18&0.10&0.95&0.59&0.76&0.09&0.10&\textcolor{darkgray}{0.96}&\textbf{0.97}\\

        interRBF20D&0.92&0.17&0.11&\textcolor{darkgray}{0.94}&0.60&0.77&0.08&0.10&\textbf{0.97}&0.88\\

        airlines&\textcolor{darkgray}{0.16}&0.04&\textcolor{darkgray}{0.16}&0.03&0.12&0.09&0.14&0.12&\textbf{0.26}&0.06\\

        elec2&0.51&0.18&0.26&0.56&0.45&0.42&0.07&0.37&\textbf{0.73}&\textcolor{darkgray}{0.59}\\

        NEweather&0.06&0.07&0.22&0.13&\textbf{0.34}&0.28&0.02&0.02&\textcolor{darkgray}{0.32}&0.07\\

        outdoorStream&0.41&-&0.37&\textcolor{darkgray}{0.90}&0.65&0.78&0.19&0.56&0.73&\textbf{0.91}\\

        \bottomrule
    \end{tabular}
    }
    }
    \label{kappamt}
\end{table}

\begin{table}
\caption{KappaT is reported to compare the performances of the methods. The highest value of KappaT in each row is highlighted.}
\makebox[1 \textwidth][c]{
\resizebox{1.1 \linewidth}{!}{ %
    \begin{tabular}{l@{\hspace{1em}} r@{\hspace{1em}} r@{\hspace{1em}} r@{\hspace{1em}} r@{\hspace{1em}} r@{\hspace{1em}} r@{\hspace{1em}} r@{\hspace{1em}} r@{\hspace{1em}} r@{\hspace{1em}} r}
        \toprule
        Dataset & DWM & $\text{Learn}^{++}$ & Learn++ & LB & OB & OSMO- & AWE & AEE & ARF & Enhash \\
        & \cite{dwm} & .NSE~\cite{learnNse} & \cite{learn++} & \cite{leverageBagging} & \cite{adwin,oza2005online} & -TEB~\cite{onlineSmotebagging} & \cite{accuracyWeightedEnsemble} & \cite{additiveExpertEnsemble} & \cite{arf} & \cite{enhash} \\
        \midrule
        transientChessboard&-0.17&\textcolor{darkgray}{0.91}&\textbf{0.94}&0.69&0.30&0.43&-1.11&-1.00&0.35&0.56\\

        rotatingHyperplane&\textbf{0.80}&0.52&0.61&0.50&\textcolor{darkgray}{0.70}&0.60&0.67&0.64&0.67&0.34\\

        mixedDrift&0.28&0.57&0.46&\textcolor{darkgray}{0.82}&0.74&0.78&0.15&0.11&0.78&\textbf{0.86}\\

        movingSquares&\textcolor{darkgray}{0.71}&0.31&0.32&0.44&0.35&0.39&0.32&0.33&0.41&\textbf{0.88}\\

        interchangingRBF&0.92&0.19&0.11&0.95&0.60&0.76&0.11&0.12&\textcolor{darkgray}{0.97}&\textbf{0.97}\\

        interRBF20D&0.92&0.18&0.13&\textcolor{darkgray}{0.94}&0.60&0.78&0.09&0.12&\textbf{0.97}&0.88\\

        airlines&\textcolor{darkgray}{0.11}&-0.02&\textcolor{darkgray}{0.11}&-0.03&0.06&0.03&0.08&0.07&\textbf{0.21}&0.01\\

        elec2&-0.41&-1.36&-1.14&-0.27&-0.59&-0.68&-1.70&-0.82&\textbf{0.21}&\textcolor{darkgray}{-0.18}\\

        NEweather&0.08&0.09&0.23&0.15&\textbf{0.35}&0.29&0.04&0.04&\textcolor{darkgray}{0.33}&0.08\\

        outdoorStream&-4.90&-&-5.23&\textcolor{darkgray}{0.05}&-2.49&-1.17&-7.04&-3.34&-1.67&\textbf{0.10}\\

        \bottomrule
    \end{tabular}
    }
    }
    \label{kappatt}
\end{table}

\begin{table}
\caption{The running time of different methods is compared using Time (in hrs). The method with the fastest speed is highlighted for every dataset.}
\makebox[1 \textwidth][c]{
\resizebox{1.1 \linewidth}{!}{ %
    \begin{tabular}{l@{\hspace{1em}} r@{\hspace{1em}} r@{\hspace{1em}} r@{\hspace{1em}} r@{\hspace{1em}} r@{\hspace{1em}} r@{\hspace{1em}} r@{\hspace{1em}} r@{\hspace{1em}} r@{\hspace{1em}} r}
        \toprule
        Dataset & DWM & $\text{Learn}^{++}$ & Learn++ & LB & OB & OSMO- & AWE & AEE & ARF & Enhash \\
        & \cite{dwm} & .NSE~\cite{learnNse} & \cite{learn++} & \cite{leverageBagging} & \cite{adwin,oza2005online} & -TEB~\cite{onlineSmotebagging} & \cite{accuracyWeightedEnsemble} & \cite{additiveExpertEnsemble} & \cite{arf} & \cite{enhash} \\
        \midrule
        transientChessboard&\textcolor{darkgray}{0.169}&0.287&0.477&8.374&0.983&13.623&0.181&0.570&0.481&\textbf{0.099}\\

        rotatingHyperplane&0.205&0.199&0.851&14.022&6.960&114.615&\textcolor{darkgray}{0.198}&0.660&1.318&\textbf{0.067}\\

        mixedDrift&0.757&\textcolor{darkgray}{0.627}&2.726&26.686&7.916&185.500&1.673&17.149&1.875&\textbf{0.339}\\

        movingSquares&\textbf{0.055}&0.179&0.790&9.368&5.936&28.028&0.142&0.567&7.308&{\textcolor{darkgray}{0.068}}\\

        interchangingRBF&\textbf{0.038}&0.185&0.972&8.298&1.444&7.424&0.376&1.048&0.499&{\textcolor{darkgray}{0.132}}\\

        interRBF20D&0.276&\textcolor{darkgray}{0.166}&0.801&19.167&1.507&7.604&2.091&3.200&3.080&\textbf{0.106}\\

        airlines&\textcolor{darkgray}{0.469}&0.571&2.180&37.910&15.151&403.378&0.744&21.669&3.796&\textbf{0.182}\\

        elec2&0.037&0.020&0.160&7.575&1.777&14.511&\textcolor{darkgray}{0.016}&0.059&0.181&\textbf{0.015}\\

        NEweather&0.021&\textcolor{darkgray}{0.008}&0.068&0.590&0.183&1.326&0.013&0.020&0.083&\textbf{0.006}\\

        outdoorStream&0.055&-&\textcolor{darkgray}{0.014}&0.069&0.127&0.162&0.047&0.129&0.071&\textbf{0.004}\\

        \bottomrule
    \end{tabular}
    }
    }
    \label{timet}
\end{table}

\begin{table}
\caption{The memory consumption is measured in terms of RAM-hours. The method with the least value of RAM-hours is highlighted for every dataset.}
\makebox[1 \textwidth][c]{
\resizebox{1.1 \linewidth}{!}{ %
    \begin{tabular}{l@{\hspace{1em}} r@{\hspace{1em}} r@{\hspace{1em}} r@{\hspace{1em}} r@{\hspace{1em}} r@{\hspace{1em}} r@{\hspace{1em}} r@{\hspace{1em}} r@{\hspace{1em}} r@{\hspace{1em}} r}
        \toprule
        Dataset & DWM & $\text{Learn}^{++}$ & Learn++ & LB & OB & OSMO- & AWE & AEE & ARF & Enhash \\
        & \cite{dwm} & .NSE~\cite{learnNse} & \cite{learn++} & \cite{leverageBagging} & \cite{adwin,oza2005online} & -TEB~\cite{onlineSmotebagging} & \cite{accuracyWeightedEnsemble} & \cite{additiveExpertEnsemble} & \cite{arf} & \cite{enhash} \\
        \midrule
        transientChessboard&\textbf{1.5e-5}&8.1e-5&\textcolor{darkgray}{3.3e-5}&8.0e-2&3.3e-4&5.5e-1&2.3e-4&5.5e-5&1.2e-3&4.4e-4\\

        rotatingHyperplane&\textbf{3.3e-5}&\textcolor{darkgray}{5.6e-5}&5.7e-5&3.5e-1&1.7e-1&4.6e+1&5.8e-4&1.1e-4&3.4e-2&7.8e-5\\

        mixedDrift&\textbf{1.2e-4}&4.9e-4&\textcolor{darkgray}{3.1e-4}&2.6e-1&2.9e-2&3.9e+1&2.5e-3&2.9e-3&1.3e-2&5.3e-3\\

        movingSquares&\textbf{1.2e-6}&4.9e-5&5.8e-5&9.0e-2&5.5e-2&2.3e+0&1.6e-4&3.3e-5&3.1e+0&\textcolor{darkgray}{1.2e-5}\\

        interchangingRBF&\textbf{7.1e-7}&\textcolor{darkgray}{5.2e-5}&1.1e-4&8.2e-2&7.9e-4&2.4e-1&5.6e-4&1.8e-4&9.3e-4&9.9e-5\\

        interRBF20D&\textbf{3.9e-5}&\textcolor{darkgray}{1.7e-4}&6.4e-4&7.8e-1&2.7e-3&1.1e+0&1.8e-2&4.5e-3&2.6e-3&1.0e-3\\

        airlines&\textbf{5.5e-5}&\textcolor{darkgray}{1.3e-3}&3.4e-3&7.3e-1&2.7e-1&3.0e+2&1.7e-3&2.6e-3&1.5e-1&1.6e-1\\

        elec2&\textcolor{darkgray}{4.9e-6}&\textbf{3.7e-6}&1.1e-5&1.4e-1&3.3e-2&1.3e+0&3.2e-5&8.0e-6&1.5e-3&1.8e-5\\

        NEweather&\textcolor{darkgray}{2.7e-6}&\textbf{7.0e-7}&4.6e-6&4.6e-3&1.4e-3&3.1e-2&1.2e-5&2.6e-6&1.2e-3&1.7e-4\\

        outdoorStream&1.5e-4&-&\textbf{2.0e-6}&2.6e-4&1.9e-3&6.0e-3&1.4e-4&4.5e-4&1.0e-4&\textcolor{darkgray}{3.4e-5}\\

        \bottomrule
    \end{tabular}
    }
    }
    \label{ramt}
\end{table}

Other evaluation criteria are speed (Table~\ref{timet}) and RAM-hours (Table~\ref{ramt}). Table~\ref{timet} reports the overall time (in hrs) taken by each method for a given dataset. For a majority of the datasets, Enhash takes the least time. For instance, it took only 0.339 hrs for the mixedDrift dataset, followed by $\text{Learn}^{++}\text{.NSE}$, which took 0.627 hrs. Notably, \glsxtrshort{osmoteb} took more than 185 hrs.

In terms of speed, DWM, $\text{Learn}^{++}\text{.NSE}$, and Learn++ are comparable to Enhash. In terms of accuracy, however, Enhash supersedes the individual methods on the majority of the datasets. Our findings were consistent across all commonly used evaluation metrics namely error, KappaM, and KappaT. For instance, DWM requires 0.038 hrs on the interchangingRBF dataset as compared to 0.132 hrs needed by Enhash. However, the error of Enhash is 2.72\%, while DWM has an error of 7.02\%.

The overall closest competitors of Enhash in terms of evaluation measures error, KappaM, and KappaT are \glsxtrshort{arf}, \glsxtrshort{lb}, and \glsxtrshort{ob}. Enhash's speed and RAM-hours' requirement are almost insignificant when compared with other methods. For instance, on the movingSquares dataset, Enhash needed only 1.2e$-$5 RAM-hours, while \glsxtrshort{arf}, \glsxtrshort{lb}, and \glsxtrshort{ob} required 3.1, 9.0e$-$2, and 5.5e$-$2 RAM-hours, respectively. Similarly, on the movingSquares dataset, Enhash completed the overall processing in 0.068 hrs, while \glsxtrshort{arf}, \glsxtrshort{lb}, and \glsxtrshort{ob} took 7.308, 9.368, and 5.936 hrs, respectively.

OSMOTE is the slowest when compared with all other methods. For the airlines dataset, OSMOTE took more than 403 hrs, while Enhash required only the minimum amount of time of 0.182 hrs.

The remaining methods, \glsxtrshort{awe}, and \glsxtrshort{aee} have inadequate performances as compared to Enhash. On the transientChessboard dataset, the error values are as high as 89.78\% and 85.37\% for \glsxtrshort{awe} and \glsxtrshort{aee}, respectively.

In summary, DWM has relatively poor performance in detecting virtual drifts but fairs well in abrupt and incremental drifts. Learn++ is exceptionally well in detecting virtual drift but severely under-performs in abrupt and incremental drifts. \glsxtrshort{lb} and \glsxtrshort{arf} suffer in detecting incremental drifts. However, \glsxtrshort{lb} and \glsxtrshort{arf} have an overall satisfactory performance. Enhash performs relatively well on all datasets. Enhash has a superior performance on a dataset consisting of three different kinds of drifts, namely incremental, virtual, and abrupt drifts. Although Enhash falls short in detecting abrupt drifts, but the performance gap is not very significant.

\section{Ablation study}\label{sec::ablation}

Enhash is built by modifying (\ref{eq::class}) and (\ref{eq::update}). The two major changes in Enhash from these are the inclusion of a forgetting factor and a heuristic for tie braking mechanism to reduce the false positives. In this section, we assess the impact of these changes, which make Enhash suitable for the \textit{concept drift} detection.

Table~\ref{ablation} compares the performances of Enhash with its two variants - Enhash-lambda0 and Enhash-noWeights. Enhash-lambda0 refers to the variant when $\lambda=0$ or equivalently, the forgetting phenomenon is not accounted. Enhash-noWeights refers to the variant of Enhash when ties in the assignment of \textit{concept class} are broken randomly. In other words, the distance (\ref{eq::distance}) of an incoming sample from the mean of the classes in the bucket is not used to determine the class in case of ties. The values of the other hyperparameters are the same as that in Section~\ref{results}. Table~\ref{ablation} shows that the performance of Enhash is much superior to its both variants for the majority of the datasets. The sub-optimal results of Enhash-noWeights may be attributed to the fact that when the same count of samples from different classes is present in the same bucket, the class for an incoming sample gets assigned randomly.

\begin{table}
\caption{Ablation study of Enhash. The performance of Enhash is compared with its two different variants- 1. Enhash with $\lambda=0$ (referred to as Enhash-lambda0), and 2. Enhash when ties in \textit{concept class} assignments are not broken by considering the distance of an incoming sample from the mean of classes in the bucket (referred to as Enhash-noWeights).}
\centering
\begin{tabular}{lrrrr}
\toprule
Dataset & Enhash & Enhash-lambda0 & Enhash-noWeights \\
\midrule
transientChessboard&\textbf{18.84}&19.57&30.58\\

rotatingHyperplane&\textbf{30.49}&32.26&36.69\\

mixedDrift&\textbf{12.88}&13.13&16.45\\

movingSquares&11.76&11.68&\textbf{11.66}\\

interchangingRBF&\textbf{2.72}&2.82&3.58\\

interRBF20D&\textbf{11.30}&11.75&12.98\\

airlines&41.66&\textbf{41.36}&43.03\\

elec2&17.34&\textbf{17.28}&17.65\\

NEweather&\textbf{29.27}&30.17&32.19\\

outdoorStream&\textbf{8.75}&\textbf{8.75}&9.55\\

\bottomrule
\end{tabular}
\label{ablation}
\end{table}

\section{Conclusions}\label{conclusion}
We conclude that Enhash supersedes other methods in terms of speed since the algorithm effectively requires only $\mathcal{O}(1)$ running time for each sample on a given estimator. In addition, the performance of Enhash in terms of error, KappaM, and KappaT is better or comparable to others for majority datasets. These datasets constitute abrupt, gradual, virtual, and reoccurring drift phenomena. The closest competitor of Enhash in terms of performance is the Adaptive Random Forest. Notably, Enhash requires, on an average 10 times lesser RAM-hours than that of Adaptive Random Forest.

\chapter{Conclusions and Future Work}

\section{Conclusions}

This thesis is motivated by the observation that rare events are not always noise but often represent significant events. The detection of rare events poses several challenges in terms of speed and performance. We designed multiple hashing-based rare events and outlier detection algorithms to overcome these challenges. In this regard, Chapter 1 contains introductory remarks and a survey of some background literature. Chapter 2 proposed FiRE, an algorithm for identifying rare events. The application of FiRE to rare cell type identification in single-cell RNA data was presented. The FiRE algorithm is an ultra-fast linear time rare event detection algorithm that can scale up to large databases. The simulation study with a minor cell cluster highlighted the efficiency of FiRE in the detection of rare cells. In another study, FiRE could identify dendritic cell subtypes amongst thousands of single cells. However, there are a few limitations of FiRE. Firstly, FiRE does not discriminate between outliers and cells representing minor cell types. The outliers, if any, are submerged into the minor cell clusters. To flag outlier cells, one may use hierarchical or density-based clustering techniques along with FiRE. Secondly, the number of possible bins that can be created in FiRE is limited by the number of dimensions used to create hashes. This limitation arises from the use of sketching as the base hash function in FiRE. Although sketching makes the FiRE robust to noises, it impedes its adoption for various cases where dimensions are less. As a solution, one may tune hyperparameters by creating more estimators or use super-spacing to increase the dimensionality of data. Note that the biological datasets do not have such issues as they tend to have large dimensions and may work well with the proposed values of hyperparameters. However, to address this issue, we proposed FiRE.1 in Chapter 3, which solves the dimension issue by replacing sketching with projection hash and introducing a new hyperparameter for quantization. This hyper-parameter reduces the hashing dependency from the number of dimensions and also extends the functionality of FiRE.1 to identify local outliers as well; thus, FiRE.1 can efficiently identify both local and global outliers. FiRE.1 was extensively compared with relevant state-of-the-art algorithms. In this chapter, a supervised scoring mechanism was also proposed that can be used to characterize datasets based on their outlier composition. About $\sim$1000 datasets were characterized and categorized to compare the performance of FiRE.1 with other relevant linear-time outlier detection algorithms. Although the quantization and projection hashing make FiRE.1 more suitable for outlier detection, they make FiRE.1 prone to noise. A too-low quantization value tends to mark every sample as an outlier, and a large quantization value converts the algorithm to a global outlier detection algorithm. Thus a proper selection of quantization values is needed. These issues may be solved using more sophisticated hashing algorithms such as spectral hashing~\cite{sh}, spherical hashing~\cite{spherical}, semi-supervised hashing~\cite{ssh1, ssh2}, etc., or building new heuristics to estimate such parameters.

The algorithms discussed so far assumed that the entire dataset is available all the time, and the distribution of the dataset remains unchanged. However, this is not always the case. The learned distribution may change with time, and in resource-constrained environments such as IoT devices, there is not enough space to accommodate the entire dataset. To overcome these challenges, Chapter 4 extended the FiRE family to non-stationary environments and proposed Enhash, an algorithm that adapts to a changing environment and identifies concept drift. Enhash incorporates a gradual forgetting of past phenomena and incrementally updates the model in response to new events to identify and adapt to concept drift. All of the above algorithms exploit the concept of hashing to their advantage, resulting in fast algorithms.

\section{Future Work}
In summary, the thesis proposes a family of FiRE algorithms that work well in tabular and streaming data. However, temporal continuity is also important, where time forms the contextual variable. Some of the prevalent examples of anomaly detection in time-series domain are electrocardiogram (ECG) signals, intrusion detection, patient-health monitoring, sudden drift in customer behavior, vibration monitoring system for machines, etc. We briefly discuss the performance of FiRE when adapted for time-series.

\subsection{Identification of anomalies in time-series data}
Outlier detection plays an important role in time series data, where temporal continuity is paramount~\cite{fox1972outliers, burman1988census}. Unusual changes in temporal patterns in data are used to model outliers. In effect, time forms the contextual variable on the basis of which analysis is performed. Data arrives sequentially, and must be processed in an online fashion. The statistics of a system may also change with time, referred to as concept drift, which was discussed in the previous chapter. Anomalous behaviour needs to be detected as soon as possible~\cite{cui2016continuous, gupta2013outlier}, and latency may be unacceptable. In some applications, computational and storage resources may be limited, and anomaly detection in such a scenario is doubly challenging.

A significant amount of work has been done to identify time series outliers. They consist of statistical techniques~\cite{barnett1984outliers, enderlein1987hawkins, rousseeuw2005robust} such as Hidden Markov Models~\cite{fine1998hierarchical}, Auto-Regressive Moving Average (ARMA), Auto-Regressive Integrated Moving Average (ARIMA) \cite{durbin2012time}, Vector Auto-Regression Moving Average (VARMA)~\cite{reinsel1993vector}, etc. Recurrent Neural Networks (RNNs)~\cite{mandic2001recurrent}, Long Short-Term Memory (LSTM)~\cite{hochreiter1997long} also have the ability to process data sequentially and model dependencies through time. Some other commonly used algorithms for identifying outliers in time-series are Bayesian Online Changepoint detection~\cite{adams2007bayesian}, EXPoSE~\cite{schneider2016expected}, and Multinomial Relative Entropy~\cite{wang2011statistical}.

\begin{figure}
    \centering
    \includegraphics[width=\linewidth, keepaspectratio]{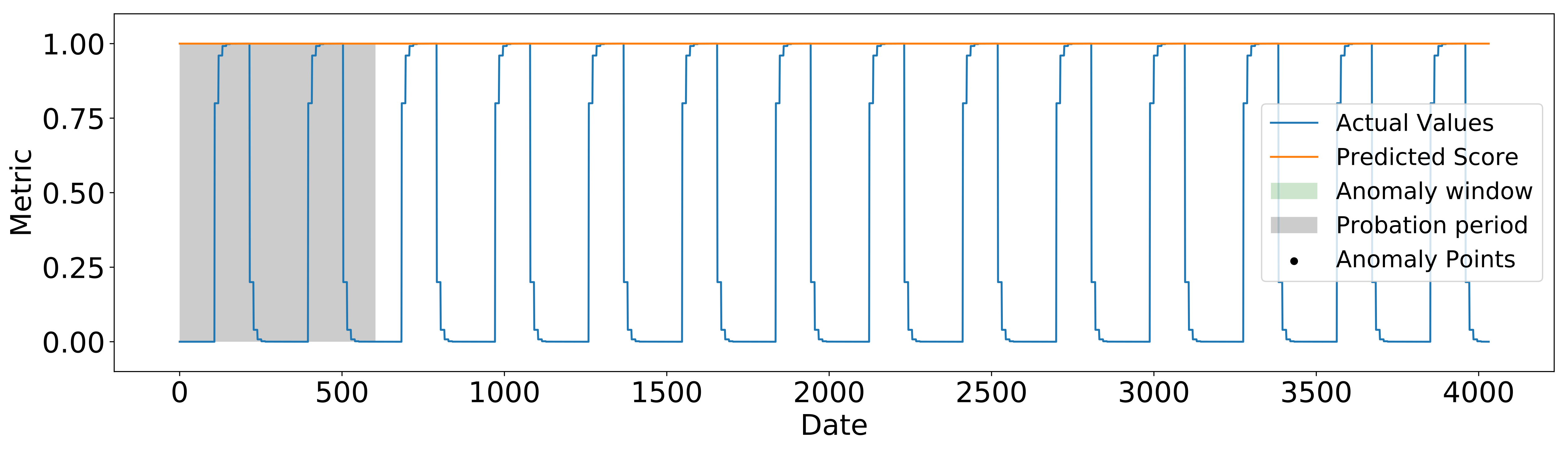}
    \caption{The data stream is periodic, has no fluctuation in observed values across different time periods and there is no anomaly. }
    \label{fig:timeSeries1}
\end{figure}

\begin{figure}
    \centering
    \includegraphics[width=\linewidth, keepaspectratio]{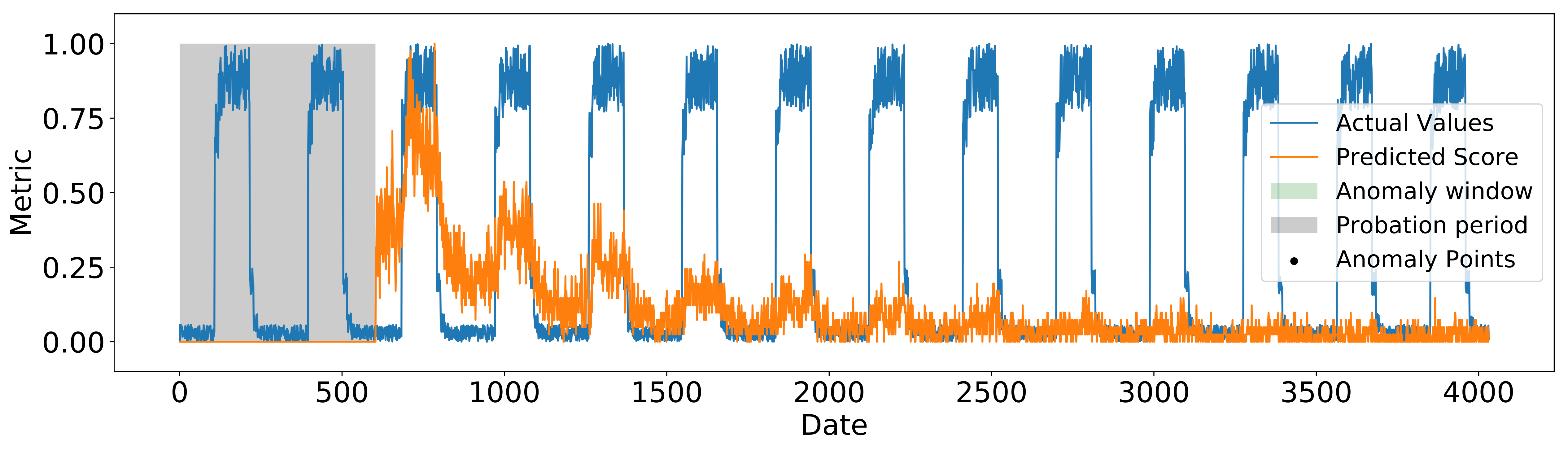}
    \caption{A periodic data stream with minor fluctuations and no anomaly. }
    \label{fig:timeSeries2}
\end{figure}

\begin{figure}
    \centering
    \includegraphics[width=\linewidth, keepaspectratio]{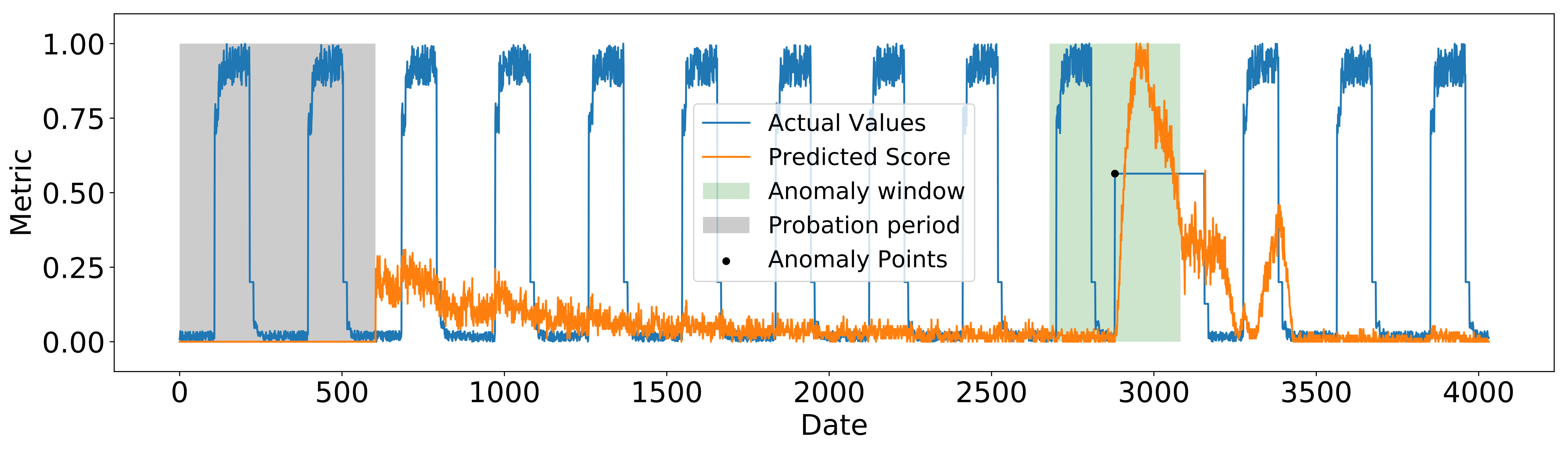}
    \caption{A periodic time-series with an anomaly. }
    \label{fig:timeSeries3}
\end{figure}

\begin{figure}
    \centering
    \includegraphics[width=\linewidth, keepaspectratio]{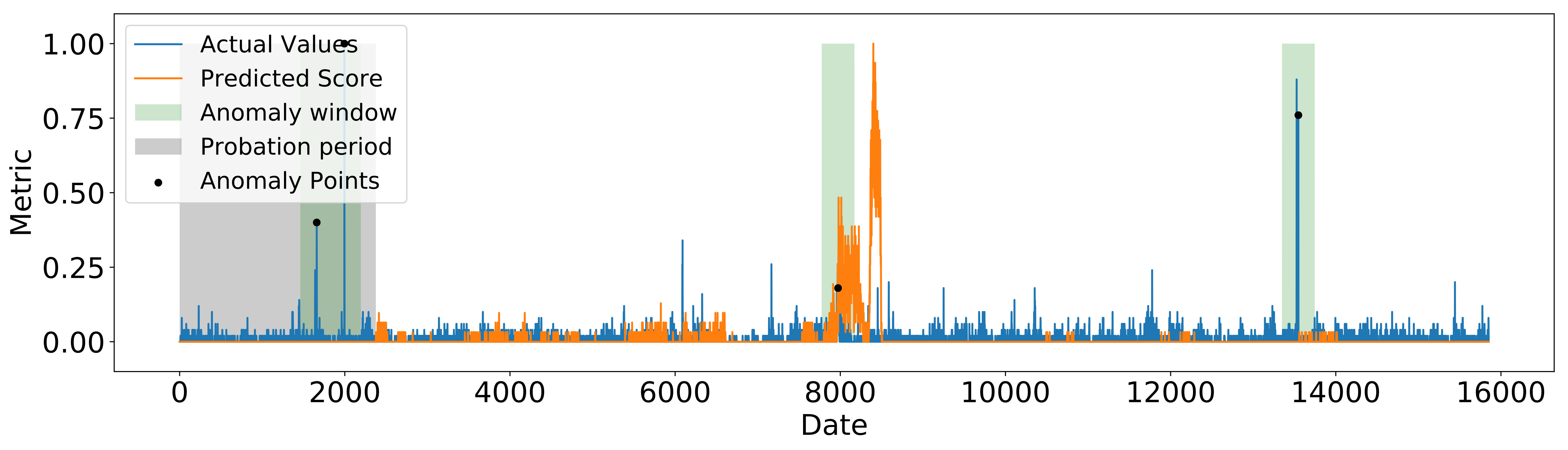}
    \caption{The presence of an anomaly in an aperiodic time-series. }
    \label{fig:timeSeries4}
\end{figure}

\begin{figure}
    \centering
    \includegraphics[width=\linewidth, keepaspectratio]{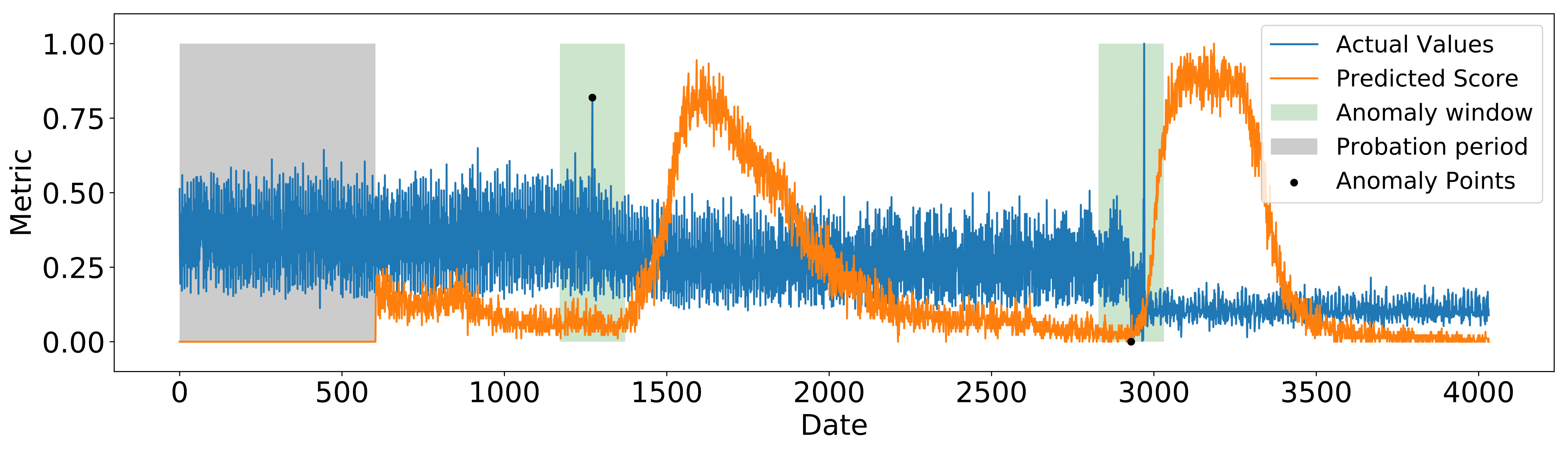}
    \caption{A data stream with a concept drift. }
    \label{fig:timeSeries5}
\end{figure}

The NAB corpus~\cite{ahmad2017unsupervised} consists of about 58 labelled data streams from various sources. The sources range from a variety of domains including social media, industrial production, cpu utilization, etc. The characteristics of data streams vary in terms of periodicity, number of outliers, concept drift, etc. We show empirically the performance of FiRE adapted for time-series with different data streams. For every stream, we define a probation period to fit the model. The initial 15\% of the entire records defines the probation period. Figure~\ref{fig:timeSeries1} shows the performance of FiRE when the stream is periodic, has no fluctuation in observed values for different periods and there is no anomaly. When the model is used to predict for the remaining stream, the FiRE-score is a constant and in correspondence with the fact that there is no anomaly. Figure~\ref{fig:timeSeries2} represents a periodic data stream with minor fluctuations in observed values but there is no anomaly. We observe that FiRE is particularly sensitive to minor fluctuations, and hence, after the probation period completes, minor fluctuations in the observed values lead to high values of FiRE-score. Once FiRE adapts, after some iterations, FiRE-scores eventually reduce. Figure~\ref{fig:timeSeries3} shows FiRE's behaviour, when the periodic time-series has an anomaly. A high peak in FiRE-score is observed with a slight delay w.r.t. the point of anomaly. FiRE uses a window-based technique to hash the incoming records. As a result, it takes some time for the effect to show up and the effect remains for the size of the window. Figure~\ref{fig:timeSeries4} represents an aperiodic time-series. FiRE is not able to identify all outliers. Figure~\ref{fig:timeSeries5} shows a data stream with a concept drift. At the point of drift, FiRE-score has a high peak, and adapts eventually.

In effect, FiRE can provide a fast and scalable solution to detect outliers where temporal continuity is important. However, there are certain limitations in the current approach and it needs some update to incorporate temporal continuity more effectively. These are some challenges that researchers may endeavour to address in the future.


\color{black}

\bibliographystyle{plain}

\begingroup
    \setlength{\bibsep}{10pt}
    \setstretch{1}
    \bibliography{main}
\endgroup

\includepdf{blank.pdf}

\newpage
\chapter*{List of Publications}
\addcontentsline{toc}{chapter}{\numberline{}List of Publications}

\section*{Publications related to thesis chapters}

\begin{enumerate}
\item Jindal A, Gupta P, Jayadeva, Sengupta D. Discovery of rare cells from voluminous single cell expression data. Nature communications. 2018 Nov 9;9(1):1-9.
\item Gupta P, Jindal A, Jayadeva, Sengupta D. Linear time identification of local and global outliers. Neurocomputing. 2021 Mar 14;429:141-50.
\item Jindal A, Gupta P, Sengupta D., Jayadeva Enhash: A Fast Streaming Algorithm for Concept Drift Detection. ESANN 2021 proceedings. Online event, 6-8 October 2021, i6doc.com publ., ISBN 978287587082-7.
\end{enumerate}

\section*{Co-Authored Publications} 

\begin{enumerate}
\item Gupta P, Jindal A, Jayadeva, Sengupta D. ComBI: Compressed Binary Search Tree for Approximate k-NN Searches in Hamming Space. Big Data Research. 2021 Jul 15;25:100223.
\item Gupta P, Jindal A, Ahuja G, Jayadeva, Sengupta D. A new deep learning technique reveals the exclusive functional contributions of individual cancer mutations. Journal of Biological Chemistry. 2022 Aug 1;298(8).
\end{enumerate}

\section*{Preprints}

\begin{enumerate}
\item Gupta P, Jindal A, Jayadeva, Sengupta D. Guided Random Forest and its application to data approximation. arXiv preprint arXiv:1909.00659. 2019 Sep 2.
\end{enumerate}


\includepdf{blank.pdf}
\chapter*{Brief Biodata of Author}
\addcontentsline{toc}{chapter}{\numberline{}Brief Biodata of Author}

\section*{Name: Aashi Jindal}

\vspace{8pt} 

\section*{Educational Qualifications}

\textbf{Ph.D.} (completed) \hfill 2023\\
Department of Electrical Engineering\\
Indian Institute of Technology Delhi, Delhi, India.\\

\noindent Bachelor of Technology (B.Tech.) \hfill 2013\\
Department of Information Technology \\
Bharati Vidyapeeth's College Of Engineering, New Delhi, India.\\

\section*{Industrial Experience}

Applied Solar Technologies India Pvt. Ltd. \hfill Dec 2021- till now\\
Data Scientist\\

\noindent NableIT Consultancy Pvt. Ltd. \hfill Nov 2020 - Dec 2021\\
Software Engineer (Machine Learning)\\

\noindent Aricent \hfill Aug 2013 - July 2015\\
Software Enfineer\\

\section*{Areas of Interest}
Machine Learning, Deep Learning, Computational Biology, Computer Vision, Natural Language Processing

\end{document}